 \documentclass[final,5p,times,twocolumn,authoryear]{elsarticle}

\usepackage{amssymb}
\usepackage{lipsum}
\usepackage{lineno}
\usepackage{amsmath}
\usepackage{hyperref}
\usepackage{cleveref}
\usepackage{url}

\usepackage{booktabs}
\usepackage{tabularx} % adjusting column width
\usepackage{multirow}
 \usepackage{doi} 
\usepackage{threeparttable}
\usepackage{makecell}
\usepackage{natbib}

\usepackage[utf8]{inputenc}  % For older engines
\usepackage[T1]{fontenc}

\usepackage{tikz}
\usepackage{pgfplots}
\pgfplotsset{compat=1.18}
\usetikzlibrary{intersections}
\usepackage{subcaption}

\usepackage{acronym}

\makeatletter
\providecommand{\url}[1]{}

\makeatother

\newcommand{\new}[1]{{\color{blue}#1}}
\newcommand{\newreplace}[2]{{{\color{red}\sout{#1}}{\color{blue}#2}}}
\newcommand{\deleted}[1]{{\color{red}[Removed: #1]}}
\newenvironment{newtable}{\color{blue}}{}

\renewcommand{\new}[1]{#1}
\renewcommand{\newreplace}[2]{#2}
\renewcommand{\deleted}[1]{}
\renewenvironment{newtable}{}{}

\usepackage[normalem]{ulem} % for \sout

\usepackage{xcolor}
\colorlet{darkblue}{blue}

\journal{Science of Remote Sensing}

\begin{document}

\begin{frontmatter}



\title{Assessing the Effectiveness of Deep Embeddings for Tree Species Classification in the Dutch Forest Inventory}

\author[first]{Takayuki Ishikawa}
\ead{wildflowers315@gmail.com}
\affiliation[first]{organization={Wageningen University},%Department and Organization
            addressline={Droevendaalsesteeg 3
            % Gaia, buildingnumber 101
            }, 
            city={Wageningen},
            postcode={6708 PB}, 
            % state={Gelderland},
            country={The Netherlands}}

\author[second]{Carmelo Bonannella}
\affiliation[second]{organization={OpenGeoHub Foundation},%Department and Organization
            addressline={Waldeck Pyrmontlaan 14}, 
            city={Doorwerth},
            postcode={6865 HK}, 
            country={The Netherlands}}

\author[third]{Bas J. W. Lerink}
\affiliation[third]{organization={Wageningen Environmental Research},%Department and Organization
            addressline={P.O. Box 47}, 
            city={Wageningen},
            postcode={6700 AA}, 
            country={The Netherlands}}

\affiliation[fourth]{organization={University of Bonn},%Department and Organization
            addressline={Nußallee 21}, 
            city={Bonn},
            postcode={53115}, 
            country={Germany}}
\author[first,fourth]{Marc Rußwurm}

%% Abbreviations
\newacro{AoI}{Area of Interest}
\newacro{API}{Application Programming Interface}
\newacro{BA}{Basal Area}
\newacro{DW}{Dynamic World}
\newacro{ERA5}{European Centre for Medium-Range Weather Forecasts Re Analysis v5}
\newacro{GEE}{Google Earth Engine}
\newacro{FAO}{Food and Agriculture Organization}
\newacro{FRA}{Forest Resource Assessment}
\newacro{GRD}{Ground Range Detected}
\newacro{GRS}{Laboratory of Geo-information Science and Remote Sensing}
\newacro{GPU}{Graphics Processing Unit}
\newacro{MAE}{Masked Auto Encoder}
\newacro{MLP}{Multi Layer Perceptron}
\newacro{MSE}{Mean Squared Error}
\newacro{NDVI}{Normalized Difference Vegetation Index}
\newacro{NFI}{National Forest Inventory}
\newacro{PALSAR}{Phased Array type L-band Synthetic Aperture Radar}
\newacro{RF}{Random Forest}
\newacro{REDD+}{Reducing Emissions from Deforestation and forest Degradation, and the role of conservation, sustainable management of forests, and enhancement of forest carbon stocks in developing countries}
\newacro{RS}{Remote Sensing}
\newacro{SAR}{Synthetic-Aperture Radar}
\newacro{SITS}{Satellite Image Time Series}
\newacro{S1}{Sentinel-1}
\newacro{S2}{Sentinel-2}
\newacro{SRTM}{Shuttle Radar Topography Mission}
\newacro{Presto}{Pretrained Remote Sensing Transformer}
\newacro{OA}{Overall Accuracy}

\begin{abstract}

\acf{NFI} serves as the primary source of forest information, however, maintaining these inventories requires labor-intensive on-site campaigns by forestry experts to identify and document tree species. Embeddings from deep pre-trained remote sensing models offer new opportunities to update \ac{NFI}s more frequently and at larger scales. While training new deep learning models on few data points remains challenging, we show that using \newreplace{pre-computed}{deep} embeddings \new{from pre-trained models}can directly outperform existing hand-designed features from \acf{SITS} that have proven effective for distinguishing tree species through seasonal canopy reflectance patterns in combination with {Random Forest (RF)} classifiers.
This work systematically investigates \emph{how deep embeddings improve tree species classification accuracy in the Netherlands with few annotated data} (1,462 pure species plots). We evaluate this question on three embedding models: Presto, Alpha Earth, and T\newreplace{essera}{ESSERA}, using three tree species datasets of varying difficulty.
Data-wise, we compare the available embeddings from Alpha Earth and T\newreplace{essera}{ESSERA} with dynamically calculated embeddings from a pre-trained Presto model, for which we extracted time series from Sentinel-1 (S-1), Sentinel-2 (S-2), and weather data, along with elevation data downloaded from Google Earth Engine. 
% approximately 7–17 percentage points
Our results demonstrate that \newreplace{fine-tuning a} publicly available remote sensing time series \newreplace{pre-trained model}{deep embeddings} outperform the current state-of-the-art \new{hand-crafted features }in \ac{NFI} \new{species} classification in the Netherlands, yielding performance gains\newreplace{of approximately 2–9 percentage points across datasets and evaluation metrics}{of roughly 7 to 9 percentage points in overall accuracy and up to 17 points in macro-F1 on the NFI datasets}\new{ at the reference plot level}. This indicates that classic hand-defined features are too simple for this task and highlights the potential of using deep embeddings for data-limited applications such as \ac{NFI} classification.
% \new{Two from-scratch deep-learning baselines trained on the same raw input confirm that this advantage comes specifically from self-supervised pre-training rather than from deep learning alone. A gradient-based attribution analysis shows the fine-tuned embedding model draws on several spectral bands rather than a single index, concentrated in phenologically informative months.}
\new{Country-scale maps built from the pre-computed embeddings reach accuracy consistent with the plot-level results and with an independent comparison against the National Forest Inventory.}
By leveraging openly available satellite data and deep embeddings from pre-trained models, this approach \newreplace{significantly}{consistently} improves classification accuracy compared to traditional methods and can effectively complement existing forest inventory processes.

\end{abstract}

%%%%%%%%%%%%%%%%%%%%%%%%%%%%%%%

%     \centering
%     \includegraphics[width=\textwidth]{2025_NFI_graphical_abstract.pdf}
%    \label{fig:Graphical_abstract_image}

% \begin{highlights}
%         \item \textbf{Research Setup:} We investigate the impact of deep time series feature embeddings on tree species classification using national forest inventory data in the Netherlands.
%         \item  \textbf{Takeaway Multi-modal sensor integration: 0.4-1.4\% accuracy improvements} when adding S-1 to S-2 features
%         \item \textbf{Takeaway Deep vs hand-crafted features: 5-10\% accuracy increase in deep embeddings} when using a same Random Forest (RF) classifier
%         \item \textbf{Takeaway MLP vs RF classifier: additional 1-4\% accuracy} increase in MLP
%         \item \textbf{Conclusion:} Computational efficiency of using deep feature embeddings from pre-trained time series foundation models like \acs{Presto} can supplement tree species mapping for large areas
% \end{highlights}

%%%%%%%%%%%%%%%%%%%%%%%%%%%%%%%


\end{frontmatter}

%\tableofcontents

% \linenumbers

\section{Introduction}
\label{introduction}
%1 Importance of Forest
Forests provide critical climate-mitigation and ecosystem services, and tree-species diversity increases productivity and resistance to disturbances \citep{tomppo_NFI_2010, FAO_FRA2020_main_2020, francini_DutchForestSpeciesMap_2024, jactel_TreeDiversityDrivesResistance_2017}. As a result, timely spatially explicit information on tree-species distribution is increasingly needed to support sustainable forest management, biodiversity assessments and carbon accounting \citep{tomppo_NFI_2010, bonannella_SpatiotemporalModelingOfVegitationDynamics_2024, unfccc_paris_agreement_2015, schelhaas_NFI6_2014, hermosilla_CanadaTreeSpeciesMapping_2022, blickensdorfer_GermanTreeSpeciesMap_2024}. 
%
%2 Importance of NFIs + problem of current NFI
\acf{NFI}s are the primary national reference for these applications, but they are fundamentally constrained by sampling-based field measurements typically repeated every 5--10 years, limiting their ability to capture rapid climate- and land-use-driven change at relevant scales \citep{tomppo_NFI_2010, bonannella_SpatiotemporalModelingOfVegitationDynamics_2024}. Spatial tree species distribution information plays an important role in \ac{NFI}s for various applications such as carbon storage estimation, forest management, and biodiversity assessments \citep{hermosilla_CanadaTreeSpeciesMapping_2022, blickensdorfer_GermanTreeSpeciesMap_2024}. Additionally, detailed tree species information is essential for national reports to the \ac{FRA} of the \ac{FAO} and Forest Europe \citep{schelhaas_NFI6_2014}.
%
%4.1 Advantage of RS 
Remote sensing has therefore become a key component of enhanced \ac{NFI}s providing frequent, wall-to-wall observations that can complement plot networks and enable more up-to-date mapping \citep{hermosilla_CanadaTreeSpeciesMapping_2022, francini_DutchForestSpeciesMap_2024, white2025enhanced}. Yet the scale and complexity of satellite time series make manual interpretation infeasible, motivating automated approaches.
%
% 4.2 Advantage of ML and \ac{RF}
Machine learning methods, especially those leveraging multi-temporal imagery, have improved performance in Earth observation applications including tree-species classification \citep{blickensdorfer_GermanTreeSpeciesMap_2024, francini_DutchForestSpeciesMap_2024, hermosilla_CanadaTreeSpeciesMapping_2022}. In operational NFI workflows, this has largely been addressed with \acf{RF} models over engineered time-series predictors, which perform well but hinge on careful feature design and local domain knowledge.
 The Random Forest algorithm is one of the most popular machine learning models for tree species classification due to its robustness, interpretability, and ability to handle high-dimensional data \citep{Breiman_RandomForest_2001}. However, \ac{RF} models require well-designed feature engineering, and the selection of appropriate features is crucial for model performance \citep{Heaton_FeatureEngineering_2016}. These choices depend on domain knowledge and target area characteristics such as climate and tree species variety, and often fail to include all necessary features \citep{ahlswede_TreeSatAIBenchArchive_2023}.

% current SOTA for tree species classification with RF 

In the context of \ac{NFI}s, several state-of-the-art machine learning models utilizing \ac{RF} have emerged \citep{hermosilla_CanadaTreeSpeciesMapping_2022, blickensdorfer_GermanTreeSpeciesMap_2024, francini_DutchForestSpeciesMap_2024}.\newreplace{}{ }The recent models have taken advantage of Satellite Image Time Series (SITS) to capture phenology differences among tree species \citep{Hemmerling_MappingTemperateForest_2021, Grabska_S2TimeSeriesPhenologyDeciduousTreeSpecies_2025}. These current methods for tree species classification rely on country-specific knowledge for selection and calculation of input features. Even though those models have achieved promising results, generalization and improvement of input data for model training remains the question. 
%
% 4.3 Deep learning model feature extraction
% why deep learning is needed
Deep learning models such as transformer architectures \citep{vaswani_attention_2023} have been recently introduced for forest monitoring, including tree species classification \citep{ma_UAV-LiDAR-treespeciesClassification_2024}. This adoption is driven by increasing interest in multimodal and time-series data fusion in \ac{RS}, enabled by the availability of big data and advancements in deep learning models \citep{Li2_ReviewOnDLinMultimodalRSdataFusion_2022}. 
Earth embeddings \citep{klemmer2025earth} from pre-computed features of deep learning models can capture complex patterns in an easy-to-access form that can be directly used for downstream tasks with classifiers and regressors including \ac{RF} \citep{Basu_DeepSat_2015}. It is evident that models such as the temporal transformer have been introduced to capture complex temporal dependencies \citep{Rußwurm_multi-temporal-land-classification_2018, saintefaregarnot_SITS_classification_2020}.
While recent studies have achieved success in regional-scale tree species classification using high-quality labeled data, significant gaps remain in large-scale classifications \citep{fassnacht_ReviewOnTreeSpeciesClassification_2016} for \ac{NFI}s due to limited labeled data availability and high computational cost for training.
%
%5 Advantage of pre-trained models
Freely available pre-trained models and derived embeddings, trained on large unlabeled datasets containing millions of pixels or images, have emerged as powerful tools for various downstream tasks \citep{tseng_Presto_2023, feng2025tessera, brown2025alphaearth}. These models and embeddings can achieve comparable or superior accuracy to traditional state-of-the-art machine learning approaches through fine-tuning without computationally expensive pre-training \citep{Bommasani_OppotunitiesOfFoundationModels_2022}. Self-supervised learning, where models are trained without labels, has gained particular attention in \ac{RS} applications \citep{wang_Self-Supervised-Learning-RS_2022}. Using self-supervised learning as a model backbone with fine-tuning on limited labeled data has demonstrated significant accuracy improvements \citep{yu_SemiSupervisedLearning_2022}, particularly in time-series analysis tasks, while requiring less inductive bias \citep{dosovitskiy_Vit_2021}. The rapid growth of both labeled and unlabeled datasets for \ac{RS} \citep{gorelick_GoogleeEarthEngine_2017, ahlswede_TreeSatAIBenchArchive_2023} has enabled the development of various pre-trained models for tasks including tree classification \citep{lu_SurveyFoundationModelForRS_2024}. 

%%%%%%%%%%%%%%%%%%%%%%%%%%%%%%%%
% Research question part

%

Despite rapid progress in machine learning for remote sensing, a key scientific challenge remains: \emph{under limited labeled data, can deep feature embedding representations capture spatial and temporal detail required by national forest inventories better than existing hand-crafted models}. While existing approaches, such as country-specific \ac{RF} models, depend heavily on expert-designed features and do not easily transfer to other ecological or sensing conditions\new{; }new pre-trained embedding models promise scalable feature extraction but have not yet been systematically assessed for operational forest inventory settings.

In short, this study asks whether representations learned from large, unlabeled remote sensing datasets can outperform or complement traditional hand-crafted features in \ac{NFI} workflows.
Our main contributions are \newreplace{threefold}{the following}:

\begin{enumerate}
    \item We integrate deep embeddings from fine-tuned self-supervised models into a tree species classification pipeline, directly comparable to a state-of-the-art \newreplace{RF}{ }approach \citep{francini_DutchForestSpeciesMap_2024}.
    \item Using Dutch National Forest Inventory data, we conduct a systematic comparison \newreplace{in plot-level}{at the reference-pixel level} between hand-crafted harmonic and seasonal medoid features\newreplace{---phenological descriptors derived from Sentinel time series using seasonal composites and harmonic regression---}{ }and \newreplace{fine-tuned }deep embeddings representations, analyzing accuracy, robustness, and generalization potential.
    % \item \new{We isolate the specific contribution of self-supervised pre-training, as opposed to deep learning in general, by comparing against two from-scratch baselines trained on identical inputs and splits.}
    \item \new{We assess operational readiness of the deep-embedding approach through country-scale wall-to-wall mapping with accuracy assessment against the National Forest Inventory.}
\end{enumerate}

\section{Methods}
\label{sec:methods}

In this study, we \newreplace{fundamentally study}{explore} the impact of time series feature embeddings on \newreplace{national forest inventory (NFI)}{\ac{NFI}} by comparing hand-designed features from \citet{francini_DutchForestSpeciesMap_2024}, which represent the current state-of-the-art for Dutch NFI mapping, with deep learning embeddings from Presto, T\newreplace{essera}{ESSERA}, and AlphaEarth \new{Foundations {AEF}}. 
\Cref{tab:model_comparison} provides a high-level comparison in terms of data and usability criteria and the following\newreplace{-}{ }subsections summarize the respective models. Time series input is relevant to capture phenological sequential patterns, downloadable embeddings make applying deep embeddings very user-friendly and applicable for large-scale mapping tasks, and an open-source available fine-tunable model allow\newreplace{}{s} further modifications and possibly better downstream accuracy.
Hence, we outline the methodological details of both complementary approaches in the next two sections before detailing the experimental setup in \cref{sec:experiments}.

\begin{table*}[ht]
\centering \small
\caption{Comparison of deep learning embedding models Presto \citep{tseng_Presto_2023}, TESSERA \citep{feng2025tessera}, and \newreplace{AlphaEarth}{AEF} \citep{brown2025alphaearth} with traditional features \citep{francini_DutchForestSpeciesMap_2024} across time data and usability criteria relevant to NFI applications. Additional information includes encoder parameter count, pretraining dataset size, and pretraining type.}
% \begin{tabular}{lcccccc}
\begin{tabular}{
        >{\raggedright}p{4.8cm}
        >{\centering}p{2.9cm}
        >{\centering}p{1.5cm}
        >{\centering}p{1.3cm}
        >{\raggedleft}p{1.6cm}
        >{\raggedleft}p{1.8cm}
        >{\centering\arraybackslash}p{1.5cm}}
\hline
\textbf{Model} & 
\textbf{Time Series Input} &
\textbf{Open\newline Embeddings} &
\textbf{Open\newline Model} &
\textbf{Encoder Parameters} &
\textbf{Pretraining Data Size} &
\textbf{Pretraining Type} \\
\hline
Presto \citep{tseng_Presto_2023} & 12 fixed monthly steps & No & Yes & 0.4M & 21.5M pixels & Pixel-level \\
TESSERA \citep{feng2025tessera} & 40 random steps/year & Yes (on request) & Yes (on request) & 40M & 0.8B pixels & Pixel-level \\
\newreplace{AlphaEarth}{AEF} \citep{brown2025alphaearth} & No (implicit temporal summarization) & Yes & No & 480M–1B & 3B images & Image-level \\
\cmidrule(lr){1-1}
Harmonic + Medoid features \citep{francini_DutchForestSpeciesMap_2024} & Harmonic coefficients and seasonal median & No & N/A & N/A & N/A & N/A \\
\hline
\end{tabular}
\label{tab:model_comparison}
\end{table*}

\subsection{Deep \newreplace{E}{e}mbeddings}
\label{sec:methods:deepfeatures}

These models reflect a recent development toward making pre-trained deep embeddings openly available (\textit{open embeddings}) either directly for download, or through open-source, pre-trained, and fine-tunable models (\textit{open model}). 

\subsubsection{Presto ``Pretrained Remote Sensing Transformer'' \citep{tseng_Presto_2023}} 
\textbf{Architecture and Training:} 
Presto is a lightweight pixel-based Transformer trained on explicit multi-sensor time series. Each pixel is represented by a 12-month sequence of Sentinel-1 (S-1), Sentinel-2 (S-2), NDVI, ERA5 climate, and Dynamic World land-cover data, augmented with static contextual variables such as SRTM elevation and geographic coordinates. Presto learns through self-supervised masked autoencoding, reconstructing missing timesteps and sensor channels to capture temporal dependencies and handle missing data.  

\textbf{Suitability for NFI:} 
Because Presto directly models monthly vegetation trajectories and integrates environmental context, it is highly suited for \newreplace{national forest inventory (NFI)}{\ac{NFI}} tasks where phenological signals (e.g., seasonal greenness, moisture, canopy change) dominate. It can characterize both intra-annual and inter-annual forest dynamics. The trade-off is that Presto must be run locally, requiring access to raw satellite data, but it offers the strongest temporal interpretability for NFI-type monitoring.

\textbf{Accessibility of model and embeddings:}
The Presto model is openly available (e.g., via GitHub) for download and local execution. However, pre-computed embeddings are not published for direct download — users must compute embeddings themselves from raw time-series inputs. This implies a heavier setup but full control over fine-tuning and feature extraction. We detail our setup in the \cref{sec:presto_implmenentation}.

\subsubsection{T\newreplace{essera}{ESSERA} ``Temporal Embeddings of Surface Spectra'' \citep{feng2025tessera}} 
\textbf{Architecture and Training:} 
T\newreplace{essera}{ESSERA} uses a dual-encoder Transformer architecture—one branch for S-1 SAR and another for S-2 optical data—processing entire annual time series per pixel (timestep~$\times$~channels). Its self-supervised pretraining applies a modified Barlow Twins loss that enforces invariance between partial temporal views of the same pixel, encouraging the model to learn phenologically stable annual representations. The output is a 128-dimensional annual embedding per 10~m pixel, globally covering 2017–2024, available as pre-computed data on request.  

\textbf{Suitability for NFI:} 
T\newreplace{essera}{ESSERA} is well suited for NFI applications because it directly encodes annual spectral and backscatter evolution—the key seasonal information driving forest structure and productivity estimates—while being readily accessible as pre-computed embeddings. However, since it outputs annual summaries rather than continuous time series, it provides a coarser representation of intra-annual phenology compared to Presto.

\textbf{Accessibility of model and embeddings:}
The T\newreplace{essera}{ESSERA} repository is publicly accessible (MIT license) and supports inference and embedding extraction. Pre-computed embeddings are provided via the accompanying GeoTessera library for many years and regions, subject to request. Fine-tuning the model is possible since the code and architecture are open, but the underlying model weights need to be requested.

\subsubsection{AlphaEarth Foundations (AEF) ``Embedding Field Model'' \citep{brown2025alphaearth}} 
\textbf{Architecture and Training:} 
\newreplace{AlphaEarth}{AEF} integrates diverse Earth observation sources (S-1/2, Landsat, GEDI, ERA5-Land, GRACE, and others) through a Space-Time Precision (STP) encoder that combines spatial and temporal attention with convolutional operations. It is trained as a multi-modal autoencoder with implicit decoders that reconstruct missing observations across time and modalities, regularized by a teacher–student consistency and contrastive alignment loss. The model learns continuous-time latent fields, producing annual ``embedding layers'' distributed globally via Google Earth Engine \citep{gorelick_GoogleeEarthEngine_2017}.  

\textbf{Suitability for NFI:} 
While \newreplace{AlphaEarth}{AEF} embeddings reflect long-term temporal consistency and cross-sensor coherence, they do not directly input dense time series per pixel. Instead, they summarize information over broader ``valid periods,'' emphasizing spatial and modal fusion rather than explicit phenological sequencing. This makes \newreplace{AlphaEarth}{AEF} moderately suited for NFI applications—excellent for stable, multi-sensor mapping of forest attributes, but less precise for capturing fine-scale seasonal forest dynamics critical to phenology-based inventories.

\textbf{Accessibility of model and embeddings:}
The embedding dataset of \newreplace{AlphaEarth}{AEF} (annual 10 m resolution embeddings) is openly accessible through the Earth Engine Data Catalog. However, the full trained model weights (for fine-tuning or re-training) are not officially released in full form by the authors. Therefore, end users can readily apply the embeddings, but cannot fine-tune the original model out of the box.

\subsection{Harmonic and \newreplace{S}{s}easonal \newreplace{F}{f}eatures}
\label{sec:methods:harmfeatures}

We reproduced closely the feature engineering approach of \citet{francini_DutchForestSpeciesMap_2024}, who published a new state-of-the-art classification approach for tree species mapping in the Netherlands.
Their approach generates 209 features containing \emph{Seasonal Medoid Statistics} and \emph{Harmonic Features} that we describe in the next two paragraphs.
While \citet{francini_DutchForestSpeciesMap_2024} used only features extracted from S-2 time series, we found experimentally (detailed later in \cref{sec:results:harmfeatures} and \cref{tab:combination_S1_S2}) that including features from S-1 further slightly improves the results.
Since we will also use S-1 to extract deep embeddings (\cref{sec:methods:deepfeatures}), we include S-1 features to ensure a fair comparison.

\paragraph{76 Seasonal Medoid Statistics}In total, 12 bands from S-1 and S-2 are selected and additional 7 indices are calculated from S-2 including \ac{NDVI}, Normalized Burn Ratio (NBR), Enhanced Vegetation Index (EVI), and Tasseled Cap transformations: Brightness (TCB), Wetness (TCW), Greenness (TCG), Angle (TCA). 
This results in 19 bands, where for each band the median band value in each of the 4 seasons, winter (January, February, December), spring (March-May), summer (June-August), and autumn (September-November), the medoids values are extracted. 
These features describe the data through season-wise statistics, but do not capture the dynamic change in seasonality itself.

\paragraph{133 Harmonic Features}To capture seasonal changes, sine and cosine harmonics 
\begin{equation}
    P_t = \beta_0 + \beta_1t + \beta_2\cos(2\pi\omega t) + \beta_3\sin(2\pi\omega t) 
    \label{eq:harmonic}
\end{equation}
are fitted to the underlying monthly signal of each band\newreplace{ according}{}. The \newreplace{seven}{six} harmonic parameters $\beta_0$, $\beta_1$, $\beta_2$, $\beta_3$, $A$, $\phi$ for each of the 19 band\newreplace{}{s} are used as classification features alongside the residual $\text{RMSE}$:
$\beta_0$ is the constant, $\beta_1$ is the time coefficient, and $\beta_2$ and $\beta_3$ are the frequency sine and cosine coefficients, respectively. To fit these four coefficients and to select $P_t$, the pixel $p$ harmonic values at time $t$, we used a least squares regression to fit Eq. \ref{eq:harmonic}. $A = \sqrt{\beta_2^2 + \beta_3^2}$ is the amplitude of the harmonic curve on the y-axis, $\phi = \arctan(\beta_3/\beta_2)$ is the phase of the curve on the x-axis to the origin and $\text{RMSE} = \sqrt{\frac{1}{n}\sum_{i=1}^{n}(P_t - X_t)^2}$ is root mean square error between $P_t$ and the actual pixel values $X_t$. The frequency $\omega$ = 1 (in years) and indicates one cycle per unit of time\newreplace{,}{.} \newreplace{i}{I}nitial guess values before fitting \newreplace{}{the }four coefficients were set as $\beta_0=0.1$, $\beta_1=0.1$, $\beta_2=0.4$, $\beta_3=0.4$.

This approach was proposed by \citet{francini_DutchForestSpeciesMap_2024} based on S-2 features. We added of S-1 features, as we found them to be slightly beneficial in accuracy (see \cref{sec:results:harmfeatures}).

%%%%%%%%%%%%%%%%%%%%%%%%%%
% Ground data
%%%%%%%%%%%%%%%%%%%%%%%%%%
\section{Data}
\label{sec:data}

\begin{table}[!ht]
    \centering
    \caption{Species count and grouping in NFI and Francini datasets. COMB: Complex \& imbalanced dataset with 13 tree species classes; SIMB: Simplified \& imbalanced dataset with 7 classes; SIBA: Simplified \& balanced dataset with 7 classes (see Section \ref{sec:data} for details).}
    \label{tab:Species_count}
    \footnotesize
    \begin{tabular}{lllll}
        \toprule
        \textbf{Dominant Species} & \textbf{COMB} & \textbf{Aggregated} & \textbf{SIMB} & \textbf{SIBA} \\
        \midrule
        \multirow[t]{1}{*}{\textit{Pinus sylvestris}} 
            & 513 & \multirow{2}{*}{Pinus} & \multirow{2}{*}{603} & \multirow{2}{*}{1,970} \\
        % \textit{Pinus pinaster} removed
        \multirow[t]{1}{*}{\textit{Other Pinus}}&89  &                          &                       & \\
        \cmidrule(l){1-5}
        \textit{Larix} spp              & 56  & Larix            & 56  & 1,970 \\
        \cmidrule(l){1-5}
        \multirow[t]{1}{*}{\textit{Quercus robur petraea}} 
            & 255 & \multirow{2}{*}{Quercus} & \multirow{2}{*}{288} & \multirow{2}{*}{1,970} \\
        \multirow[t]{1}{*}{\textit{Other Quercus}}&33  &                             &                       & \\
        \cmidrule(l){1-5}
        \textit{Fagus} spp               & 58  & Beech            & 58  & 1,970 \\
        \cmidrule(l){1-5}
        \textit{Populus} spp             & 72  & Populus          & 72  & 1,970 \\
        \cmidrule(l){1-5}
        \multirow[t]{4}{*}{\textit{Alnus} spp} 
            & 30  & \multirow{4}{*}{Other Broadleaves} & \multirow{4}{*}{242} & \multirow{4}{*}{1,970} \\
        \textit{Betula} spp              & 58  &                  &                       & \\
        \textit{Fraxinus} spp            & 40  &                  &                       & \\
        Other broadleaved                & 102 &                  &                       & \\
        \cmidrule(l){1-5}
        \multirow[t]{2}{*}{\textit{Pseudotsuga menziesii}} 
            & 90  & \multirow{2}{*}{DarkConifer} & \multirow{2}{*}{160} & \multirow{2}{*}{1,970} \\
        \textit{Picea} spp               & 66  &                       &                       & \\
        \midrule
        total & 1,462 &                        & 1,479 & 13,790 \\
        \bottomrule
    \end{tabular}
\end{table}

\subsection{Study area}
\label{sec:study_area}
\begin{figure*}[!ht]
    % \centering
    \begin{subfigure}{.5\textwidth}
        \centering
        \includegraphics[width=\textwidth, height=7cm, keepaspectratio]{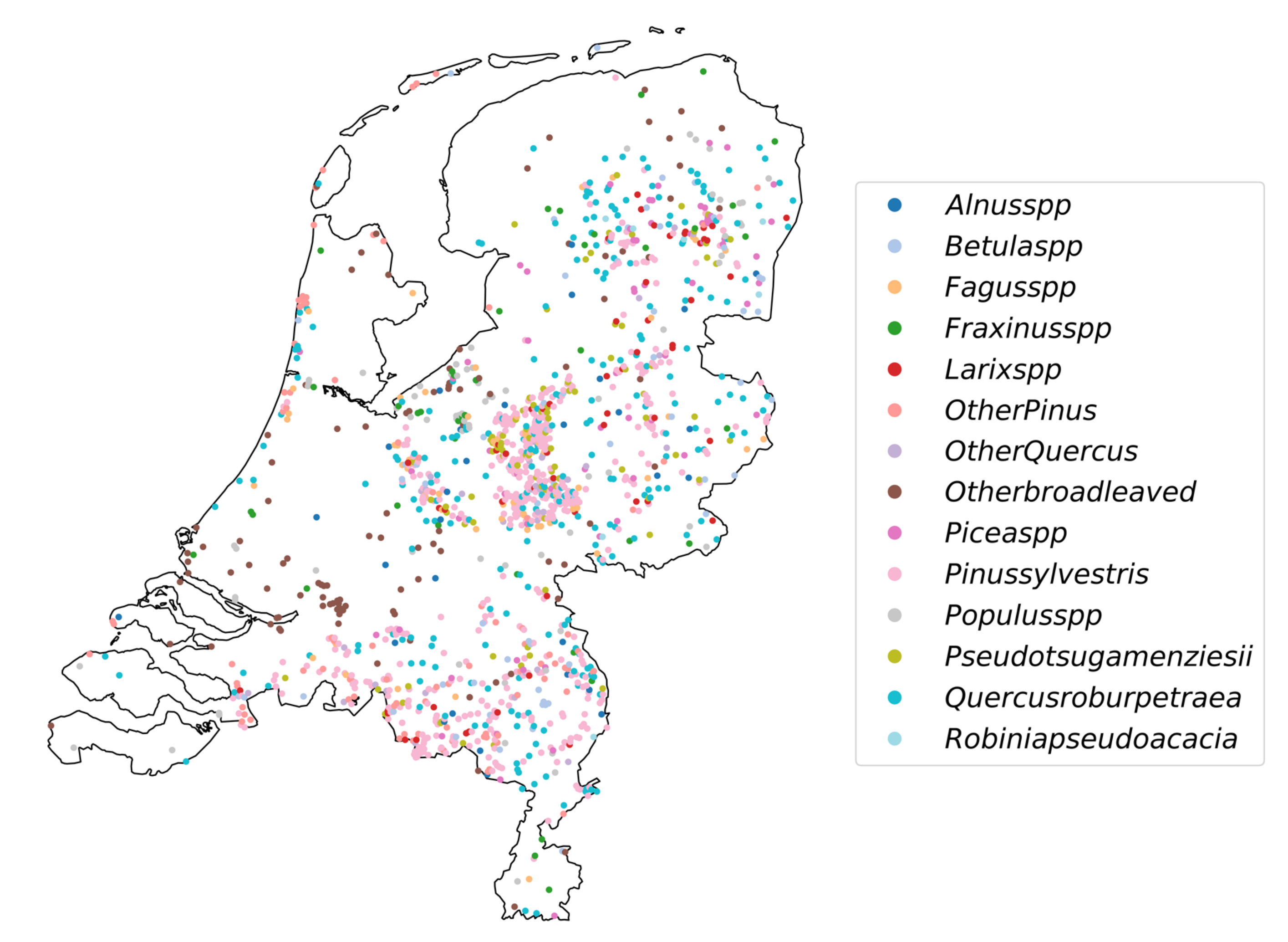}
        \caption{COMB (13 classes \& imbalanced)}
        \label{fig:study:comb}
    \end{subfigure}
    \hfill
    \begin{subfigure}{.5\textwidth}
        \centering
        \includegraphics[width=\textwidth, height=7cm, keepaspectratio]{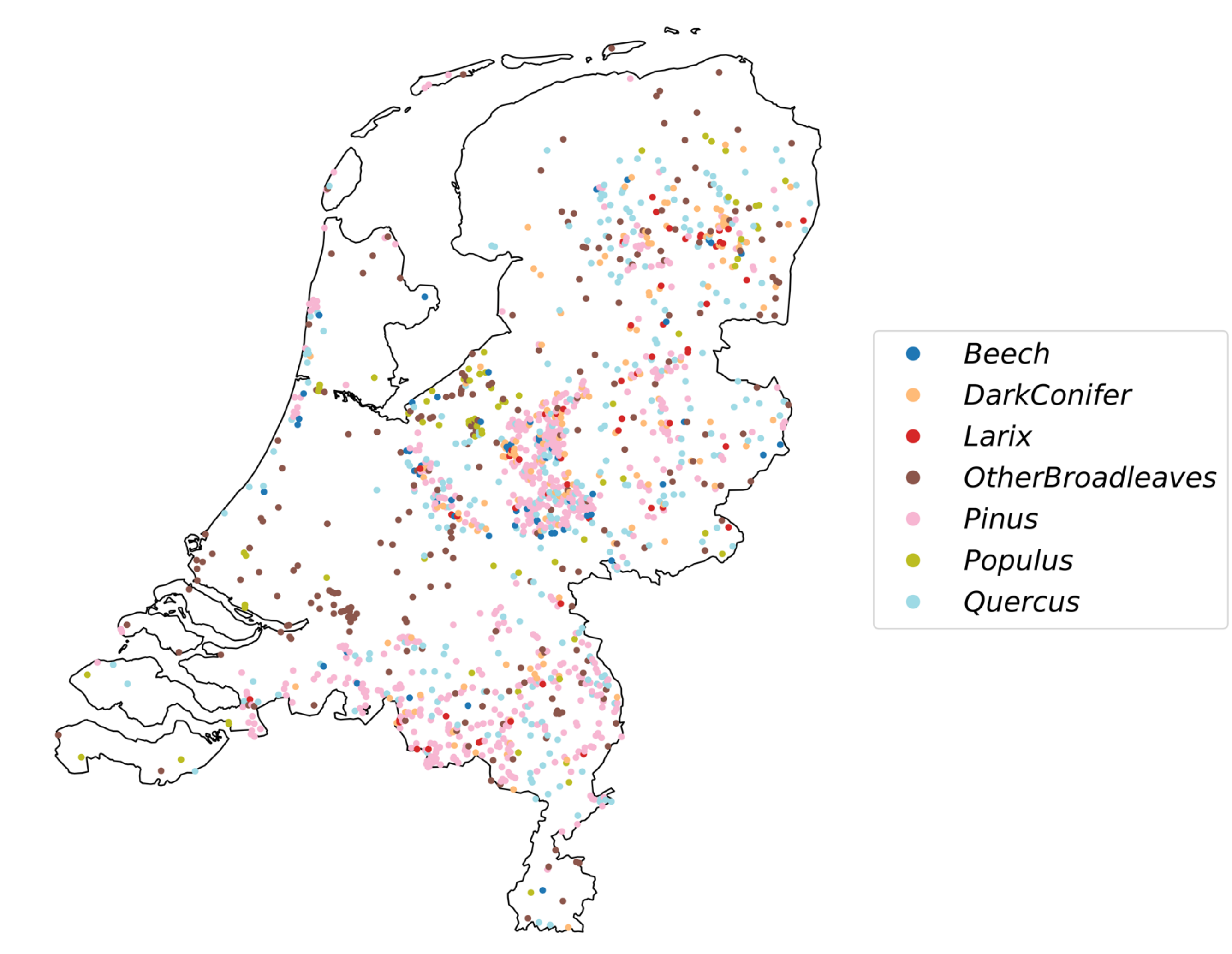}
        \caption{SIMB (7 classes \& imbalanced)}
        \label{fig:study:simb}
    \end{subfigure}
    
    \vspace{0.5cm}
    
    \begin{subfigure}{.5\textwidth}
        \centering
        \includegraphics[width=\textwidth, height=7cm, keepaspectratio]{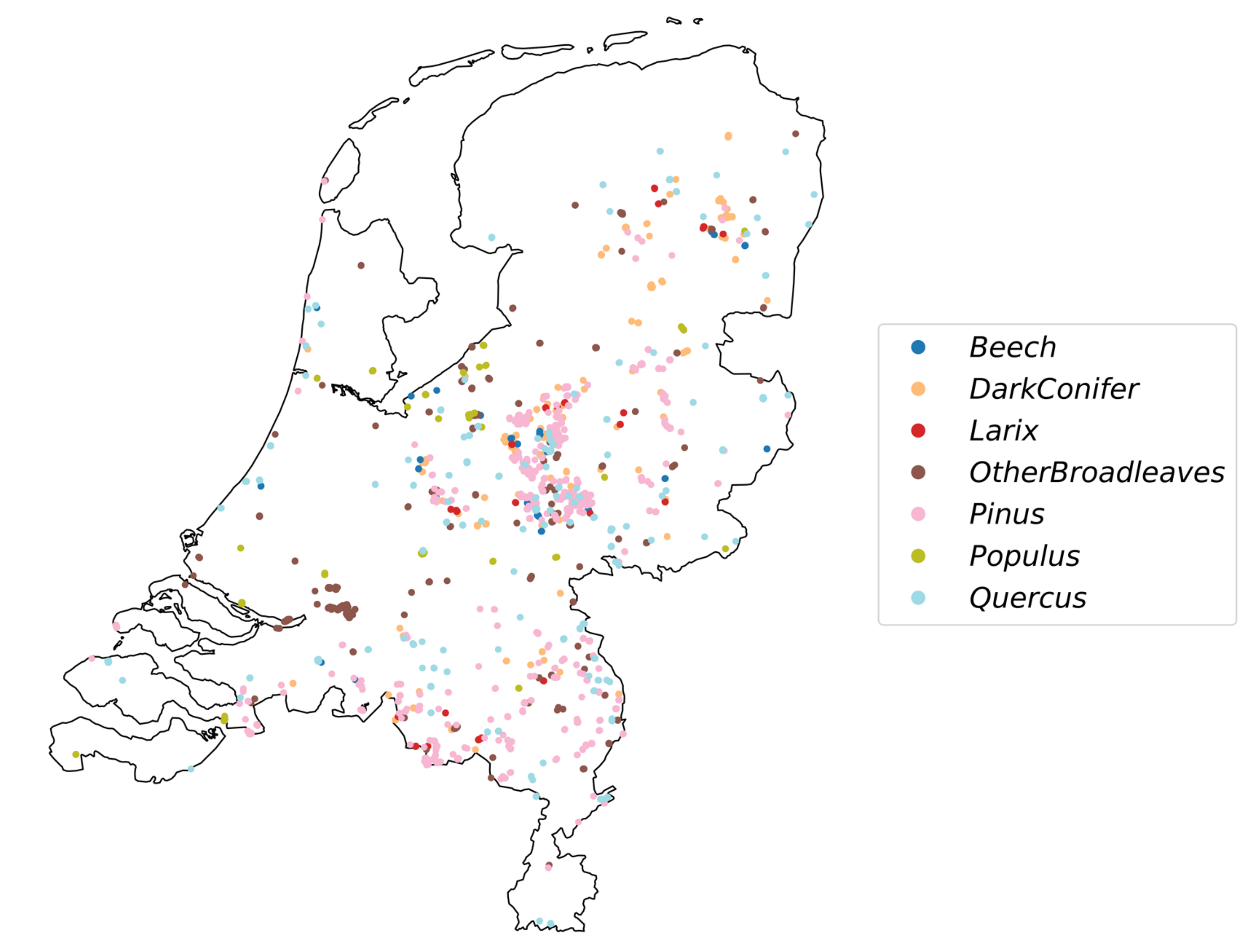}
        \caption{SIBA (7 classes \& balanced)}
        \label{fig:study:siba}
    \end{subfigure}
    \hfill
    \begin{subfigure}{.5\textwidth}
        \centering
        \includegraphics[width=\textwidth, height=7cm, keepaspectratio]{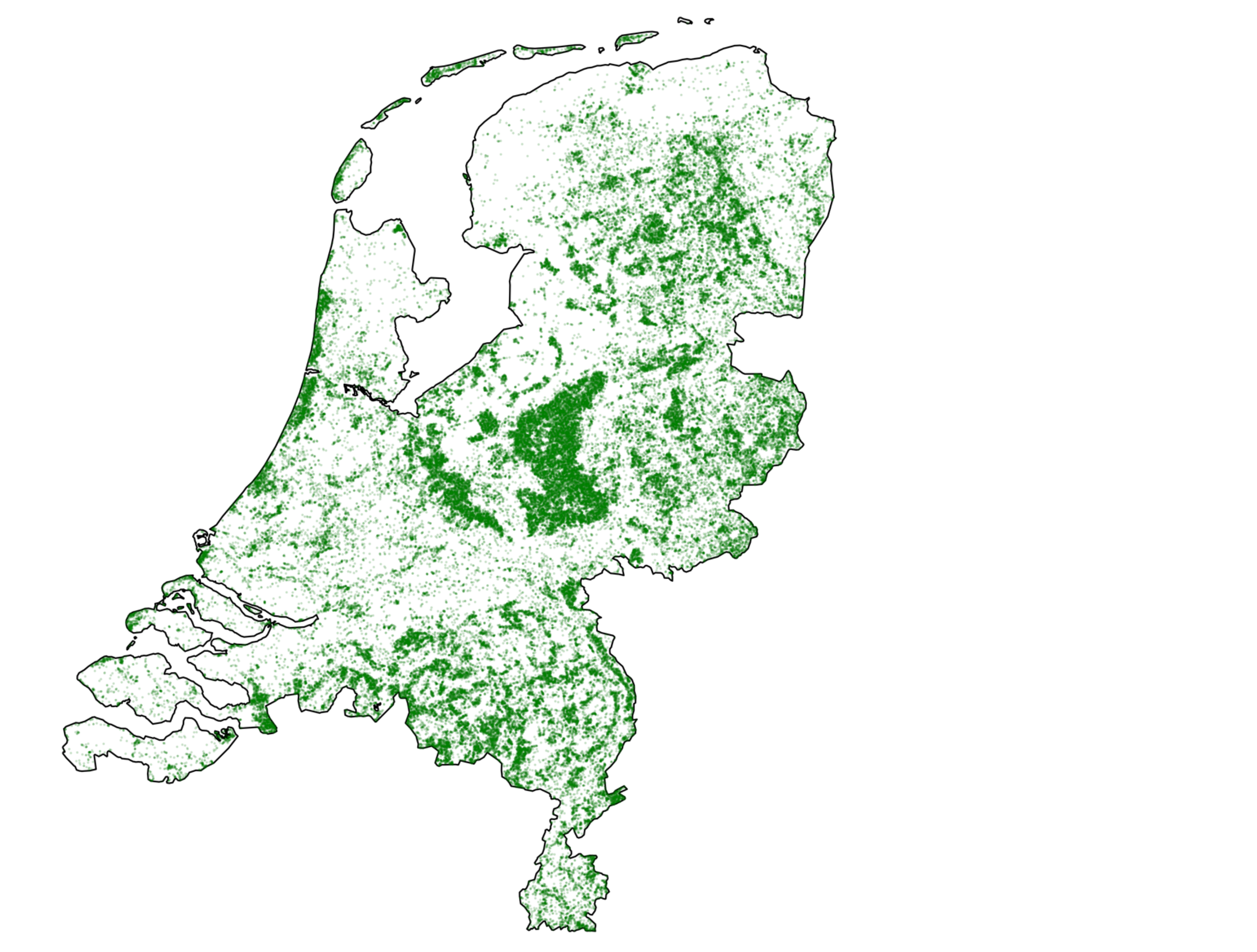}
        \caption{Forests in the Netherlands}
        \label{fig:study:forest}
    \end{subfigure}
    
    \caption{Overview of the study area and the three datasets used in this study. The simplified, class-balanced dataset is identical to related work by \citet{francini_DutchForestSpeciesMap_2024}.}
    \label{fig:study:overview}
\end{figure*}

This study focuses on forests in the Netherlands, which were described in the annual Greenhouse Gas accounting report \citep{arets_greenhouse_2023}. According to the latest \newreplace{\acl{NFI} 7}{\ac{NFI}7}, forests covered 363,801 ha in 2021, corresponding to 11\% of the land use in the Netherlands \citep{schelhaas_DutchNFI7_2022} and comprising approximately 36.4 million 10 $\times$ 10m pixels, as shown in \cref{fig:study:forest}.

\subsection{National Forest Inventory}

\noindent
We use three datasets from the Dutch \newreplace{National Forest Inventory (NFI)}{\ac{NFI}} at different complexity levels\newreplace{ that we will use to compare the proposed deep embeddings with the RF by Francini et al., 2024 fitted on harmonic time series features}{}, as detailed later in \cref{sec:methods}.
Despite the relatively small overall forest area in the Netherlands (approximately 10–11 \% of land cover), forests display spatial variability in species composition, age structure, and mixed stand types, reflecting responses to site conditions, management history, and ecological processes \citep{schelhaas_NFI6_2014}. 

\begin{description}
    \item[Complex \& imbalanced (COMB)] 13 tree species classes, imbalanced distribution, 1,462 time series samples.
    \item[Simplified \& imbalanced (SIMB)] 7 tree species classes, imbalanced class distribution, 1,479 time series samples.
    \item[Simplified \& balanced (SIBA)] 7 species classes, a balanced class distribution, as proposed by \citet{francini_DutchForestSpeciesMap_2024}. In total 13,790 samples and contains additional samples augmented by photointerpretation from Dutch NFI samples.
\end{description}

\newreplace{Orginal}{Original} Dutch NFI samples (COMB and SIMB) are derived by random sampling with a density of 1 point per 100 ha \citep{schelhaas_NFI6_2014}, which means that each random point is designated per square kilometer\newreplace{, and SIBA removed adjacent pixels with minimum distance among sample pixels was set to 15m, which reduced}{. For SIBA, adjacent pixels were removed by setting a minimum distance of 15 m among sample pixels, which reduced} issues associated with spatial autocorrelation \citep{francini_DutchForestSpeciesMap_2024} \new{while \citet{hermosilla_CanadaTreeSpeciesMapping_2022} uses a minimum distance of 45 m}.
\Cref{tab:Species_count} shows a detailed list of the tree species across all three datasets.

\textbf{Label Data Origin}
Ground truth data were collected manually through field measurements at 3,062 plots for the Dutch \newreplace{\acl{NFI} 6}{\ac{NFI}6} between 2012 and 2013. Each plot contained a circular area with a variable radius (5 to 20 m) to ensure inclusion of at least 20 trees \citep{schelhaas_DutchNFI7_2022}. These plots correspond to 1 to 16 pixels at a $10 \times 10$m resolution of S-1 and S-2.
\newreplace{Due to privacy considerations, the precise coordinates of the plots were obtained under a confidentiality agreement with the Dutch government, with the requirement that the data be discarded after project completion.} Prior to data disposal, we utilized the plot center coordinates from the total 3,062 pixel-level data points to extract satellite data at $10 \times 10$ m pixel resolution from \ac{GEE} \citep{gorelick_GoogleeEarthEngine_2017}.

\textbf{Data Quality Filtering}\newreplace{. }{}
Following the identical protocol to \citet{francini_DutchForestSpeciesMap_2024}, we selected plots where a single dominant species represented more than 80\% of the \ac{BA}, which is the cross-sectional area of trees at breast height. This process removed the \textit{Castanea spp} dominant species class that had no samples at this threshold. 

\textbf{Class Aggregation}
The original \ac{NFI} data contains 19 dominant tree species classes. However, we noticed that six dominant species classes have fewer than 10 samples and we decided to remove these classes to not extensively rely on underrepresented tree species in our evaluation, similar to \citet{Kang_UnderSamplingScheme_2017}. This resulted in the dataset \textbf{Complex \& Imbalanced (COMB)} with 1,462 samples with 13 classes for the dominant species classification task. For the \textbf{Simplified \& Imbalanced (SIMB)}, we further aggregated the classes into the 7 tree species identical to \citet{francini_DutchForestSpeciesMap_2024}. This resulted in 1\newreplace{}{,}479 data samples.

Finally, we also acquired the identical dataset used by \citet{francini_DutchForestSpeciesMap_2024} which contains 13,790 data points evenly distributed across the aggregated classes (1,970 points per class). This dataset originated from the same \ac{NFI} data but was augmented with additional labeled data points through visual interpretation of satellite imagery. We call \newreplace{}{this dataset }\textbf{Simplified \& Balanced (SIBA)} compared to \newreplace{aforementioned}{the aforementioned} ones, as it contains 7 tree species classes and an even number of samples per class. \newreplace{To augment the original Dutch NFI plots, the authors performed a rigorous three-step validation process: (1) visual photointerpretation using high-resolution orthophotos and Google satellite imagery to remove plots affected by land-cover change or mixed-species compositions; (2) stand-polygon identification to extract homogenous spectral samples; and (3) spatial thinning, where a minimum distance of 15 m was maintained between pixels to minimize spatial autocorrelation and label noise. The final dataset comprises 13,790 samples (1,970 per class).}

\subsection{Remote Sensing Time Series for Presto and harmonic + medoid calculation}

We extracted time-series data spanning January \newreplace{through}{to} December 2020 \newreplace{through}{via} Google Earth Engine \citep{gorelick_GoogleeEarthEngine_2017}. This temporal range aligns with both the \ac{Presto} model requirements \citep{tseng_Presto_2023} and previous Dutch tree species classification research \citep{francini_DutchForestSpeciesMap_2024}. 

\textbf{Input Data Modalities S-2, S-1, ERA5, SRTM}. This study employs a comprehensive multi-source remote sensing (RS) approach that integrates satellite imagery and environmental data to characterize forest vegetation across space and time. The model relies on several key data sources: S-1 SAR GRD data with VH and VV polarizations at a 6-day revisit frequency; S-2 multispectral Level 1C data with 5–10 day revisit frequency, including 10 spectral bands (B2–B8, B8A, B11, B12) and NDVI computed during preprocessing; ERA5 climate data providing monthly measurements of 2m air temperature and total precipitation; and static SRTM terrain data comprising elevation and slope. Using the plot-level coordinates, we extracted a single set of time-series data (or embeddings) for each plot; consequently, the \newreplace{COMB}{SIBA} dataset comprises 13,790 plots and 13,790 corresponding samples. During preprocessing, satellite data with coarser spatial resolutions than 10\,m (i.e., \newreplace{S-1, }{}ERA5, and SRTM) were resampled to a common spatial resolution of 10\,m to ensure consistency with Presto input \citep{tseng_Presto_2023} and harmonic + medoid feature input \citep{francini_DutchForestSpeciesMap_2024}.

\textbf{Remote Sensing Data Preparation}
For S-2 data, cloud masking was performed using the S-2 cloud probability dataset \citep{pasquarella_QualityAssessment_CloudProbability_2023}, excluding pixels with cloud cover probability exceeding 65\% \citep{francini_DutchForestSpeciesMap_2024}. 
Geographic coordinates were utilized both for satellite data extraction from the \ac{GEE} archive and as input features for the model. Raw time-series data was downloaded as individual CSV files for each plot location across all datasets (SIMB, COMB and SIBA), ensuring direct correspondence between remote sensing observations and ground truth measurements.

\section{Experimental Setup}
\label{sec:experiments}

This study compares classic machine learning with hand-crafted feature engineering against deep learning \new{embeddings }approaches for tree species classification. Our experimental design follows a systematic progression to evaluate different methodological approaches and their combinations.

\subsection{Comparative evaluation framework}
Our experimental design implements four distinct comparison scenarios to systematically evaluate feature representations and classifier architectures:

\begin{enumerate}
    \item \textbf{Sensor contribution analysis}: S-1 vs S-2 vs S-1+S-2 using harmonic + medoid features with \ac{RF}
    \item \textbf{Feature representation comparison}: Harmonic + medoid features vs deep embeddings using RF
    \item \textbf{Classifier architecture evaluation}: \ac{RF} vs \newreplace{Multi Layer Perceptron (MLP)}{deep-learning baselines} using deep embeddings
    \item \new{\textbf{Spatial context ablation}: single pixel vs.\ $3\times3$ neighbourhood-patch pooling (flatten, mean-pool, stats-pool)}
\end{enumerate}

This systematic approach allows us to isolate the contributions of different methodological components while building toward the optimal configuration for operational NFI applications, extending the work of \citet{francini_DutchForestSpeciesMap_2024} with comprehensive deep learning comparisons.

\subsection{Model fine-tuning\new{ and deep learning baselines}}
The fine-tuning approach leverages the pre-trained Presto encoder's learned representations while adapting the model to the specific characteristics of Dutch forest species. This transfer learning strategy balances computational efficiency with task-specific optimization.

For fine-tuning the pre-trained models, the encoder block from the pre-trained model was reused and trained with newly attached \acf{MLP} classifier for the downstream task. This fine-tuning model is trained on the 1,479 and 13,790 samples for 7 aggregated groups classification task (SIMB, SIBA), and the 1,462 samples for the 13 dominant species classification task (COMB). Pre-trained encoder of Presto was set as trainable in this process. 

Based on the current state-of-the-art classifier for tree species classification \citep{Mouret_TreeSpeciesClassificationPixelTimeSeriesInbalanced_2024}, the \newreplace{\ac{MLP}}{Multi Layer Perceptron (MLP)} classifier architecture consists of a 3-layer \ac{MLP} with no dropout (rate: 0.0)\new{, with 1024, 512, and 256 nodes in successive layers, batch normalization, and ReLU activation}. \newreplace{The layer configuration comprises 1024 nodes in the first layer, 512 nodes in the second layer, and 256 nodes in the final layer, with batch normalization and ReLU activation applied throughout. }Training parameters \newreplace{are configured as follows:}{use} a learning rate of 0.0001 \newreplace{with weight decay of 0.00746}{and no weight decay}. \newreplace{The model is trained for a maximum of 100 epochs using a batch size of 64, with cross-entropy loss for optimization. }{An adaptive learning-rate schedule halved the learning rate whenever the validation loss did not improve for 20 epochs, and training was stopped after 40 epochs without improvement.} The best fine-tuned model and its accuracy were selected based on validation loss. \new{The complete training configuration for all deep-learning models evaluated in this study (the fine-tuned Presto \ac{MLP} head and the two from-scratch baselines), including epoch budget, batch size, and loss function, is summarized in \cref{tab:hyperparams}.}

\new{As a deep-learning baseline trained entirely from scratch, we additionally evaluated a Temporal Convolutional Neural Network (TempCNN), an established architecture for satellite image time-series classification \citep{Pelletier_TempCNN_2019}, configured following \citet{Mouret_TreeSpeciesClassificationPixelTimeSeriesInbalanced_2024} for tree-species classification: three one-dimensional convolutional blocks, each followed by batch normalization, ReLU activation, and dropout, then a global max-pooling layer over the temporal dimension and a fully connected layer with softmax output. The TempCNN is trained from random initialization on the same 12-timestep, 17-band pixel time series, train/validation/test splits, random seeds, and training schedule as the fine-tuned Presto model (full architecture and training hyperparameters in \cref{tab:hyperparams}). Because it shares the same inputs, splits, and training schedule as the fine-tuned Presto model and differs only in the absence of pre-trained representations, this from-scratch baseline isolates the contribution of self-supervised pre-training.}

\new{As a second from-scratch baseline, designed to isolate the contribution of self-supervised \emph{pre-training} specifically rather than deep learning in general, we evaluate a from-scratch \ac{MLP} that shares the \emph{same} head architecture as the fine-tuned Presto classifier, applied to the flattened raw time series (the $12\times17$ input reshaped to a 204-dimensional vector) instead of the Presto embedding, and trained from random initialization with no pre-trained encoder. It uses the same training schedule, train/validation/test splits, and seeds as the other deep-learning models (full configuration and parameter count in \cref{tab:hyperparams}). Comparing this from-scratch \ac{MLP} against fine-tuned Presto with the \emph{identical} \ac{MLP} head therefore provides an architecture-controlled ablation of the value of pre-training: the only difference is whether the representation is learned from scratch on the raw input or inherited from the pre-trained Presto encoder.}

\newreplace{All fine‑tuning and embedding extraction experiments were conducted on Google Colab using an NVIDIA Tesla T4 GPU (16GB GPU memory). Each fine‑tuning run used a maximum of 100 training epochs, but due to early stopping based on validation performance, most runs terminated in fewer than 30 epochs in SIMB (1,479 samples) and COMB (1,462 samples), 50 epochs in SIBA (13,790 samples). As a result, a single fine‑tuning run on the smaller SIMB and COMB datasets takes approximately 1 minute, with embedding extraction requiring an additional ~1 minute per dataset. For the larger SIBA dataset, a single fine‑tuning run takes approximately 10 minutes and embedding extraction takes ~3 minutes.}{Compute and wall-clock cost for fine-tuning and embedding extraction are reported in \cref{app:compute_cost}.}

\subsection{Data splitting strategy}
Out-of-bag (OOB) error estimate with RF algorithm was applied in \cite{francini_DutchForestSpeciesMap_2024}\newreplace{,}{;} however\newreplace{}{,} we also need to evaluate \newreplace{MLP}{the MLP and TempCNN} \newreplace{and}{}classifier\newreplace{}{s} in a \newreplace{comparable setting with}{setting directly comparable to} \newreplace{}{the }RF classifier.
Based on prior research \citep{blickensdorfer_GermanTreeSpeciesMap_2024, Freudenberg_S2-TreeSpeciesClassificationGermany_2025, hermosilla_CanadaTreeSpeciesMapping_2022}, a 70/30 (approximately 2:1 ratio) train-test split was applied for fine-tuning and \ac{RF} classification. In addition, for MLP, training data is split into MLP train data and MLP validation data with a ratio of 80:20 (56\%, 14\% of the total dataset) to select the best model by lowest validation loss, which ensures that test data is not used during training and reduces overfitting. During fine-tuning and classification, the same training and test data were used across datasets and models. This practice ensures that comparable results across the model with same train-test split and extracted embeddings are not trained on test data. By adopting stratified sampling, each class is equally split into train (MLP train data and MLP validation data) and test data at this ratio. Appendix table \ref{tab:class_distribution_splits} shows species class distribution across train (MLP train and MLP validation) and test splits for three datasets.
 As described in \Cref{sec:data}, \newreplace{our datasets have already mitigated spatial autocorrelation.}{both data sources include measures to reduce spatial autocorrelation (the NFI sampling grid for SIMB and COMB; 15\,m pixel thinning for SIBA), although for SIBA this thinning is weaker than the 45\,m minimum distance recommended by \citet{hermosilla_CanadaTreeSpeciesMapping_2022}.}
In addition, our datasets\newreplace{}{,} especially NFI data (SIMB and COMB)\newreplace{}{,} are small (1,462 to 1,479) and \newreplace{have a lot of classes}{have many classes} (6 to 13 classes), which makes a spatial\newreplace{}{ly} aware split with class-wise stratified sampl\newreplace{e}{ing} difficult. Therefore, we apply stratified random splits. \new{Because the SIBA samples originate from a limited number of photo-interpreted stand polygons, a purely pixel-level split can place near-duplicate pixels of the same polygon in both the training and test partitions \citep{francini_DutchForestSpeciesMap_2024}. For SIBA we therefore additionally evaluate a plot-grouped (spatially blocked) split, in which all pixels belonging to a given polygon are assigned exclusively to either training or testing; whole polygons are selected per class by a pixel budget so that the class balance of the standard split is preserved. We regard the standard pixel-level split as our primary protocol, because it follows the sampling design of the original SIBA study \citep{francini_DutchForestSpeciesMap_2024} and keeps SIBA comparable to the NFI-based SIMB and COMB datasets, whose systematic sampling grid already enforces large inter-plot distances and good spatial coverage. National forest inventories, including the Dutch NFI, are typically established on a systematic grid or by stratified/unstratified random sampling rather than on clustered photo-interpreted polygons, so plot-level blocking is not the conventional evaluation protocol for such data and is not applied by \citet{francini_DutchForestSpeciesMap_2024} or comparable NFI tree-species study \citep{hermosilla_CanadaTreeSpeciesMapping_2022}. The plot-grouped split is therefore a deliberately conservative, and for operational NFI mapping somewhat extreme, stress test, which we report in \cref{app:siba_spatial} rather than in the main results.} We repeat each experiment five times, applying different random seeds to generate independent data splits.

\subsection{Training pipeline}
    In each random seed, we split each dataset for 70\% training (56\% MLP training, 14\% MLP validation) and 30\% test. MLP training dataset is used for fine-tuning Presto decoder with MLP classifier, and MLP validation data is used to decide the best model. Using this best fine-tune\new{d} Presto model with MLP classifier, we evaluate test data (MLP fine-tune\new{d} Presto result). In addition, by using fine-tune\new{d} Presto decoder, we extracted embeddings for all samples in the dataset. Those fine-tune presto embeddings as well as other embeddings (AEF and TESSERA) are splitted in the exact same manner as MLP fine-tuning. 70\% of those embeddings are used for RF training and 30\% are used for RF evaluation. 

\subsection{Random Forest classifier}
    Following the experimental setup of \citet{francini_DutchForestSpeciesMap_2024}, we implemented a Random Forest (\ac{RF}) classifier \citep{Breiman_RandomForest_2001} consisting of 500 trees. The number of features considered at each split was set to the square root of the total number of input predictors.

\subsection{Performance evaluation}
Model performance was evaluated using a comprehensive set of established accuracy metrics: per-class recall, F1 scores (Macro, Weighted), confusion matrices, and overall accuracy. While overall accuracy provides a general assessment of classification performance, it exhibits reduced sensitivity to class imbalance issues that are prevalent in ecological datasets. Therefore, the F1 score (Macro, Weighted), which represents the harmonic mean of precision and recall, was employed as a supplementary metric to provide a more balanced evaluation of model performance when addressing the inherent class imbalances in tree species distribution data. Confusion matrices were used to visualize the model's performance at the class level.
These metrics are widely used in tree species classification studies \citep{Goutte_ProbabilisticInterpretationPrecision_2005, hermosilla_CanadaTreeSpeciesMapping_2022,francini_DutchForestSpeciesMap_2024, blickensdorfer_GermanTreeSpeciesMap_2024} and provide comprehensive assessment of classification performance across all species classes.

\subsection{Statistical validation}
To evaluate variability of model performance, each model was evaluated five times with different random seeds, which affect to train-test split and model initialization. The mean and standard deviation of the accuracy metrics were calculated for each model.

\subsection{\new{Spatial context: $n\times n$ neighbourhood-patch pooling}}
\label{sec:spatial_patch_methods}

\new{In order to investigate whether incorporating spatial context into pixel-based model improves classification performance, we test this for AEF, TESSERA, and fine-tuned Presto on the \ac{NFI} datasets (SIMB, COMB) and, under both the pixel-level and plot-grouped splits (\cref{sec:experiments}), on SIBA.}

\new{For each plot we retrieved the embedding (AEF, TESSERA) or the raw $12\times17$ time series (Presto) of the eight $10\,\mathrm{m}$ pixels neighbouring the plot's own pixel, forming a $3\times3$ ($30\,\mathrm{m}$) window. Because pre-computed Presto embeddings do not exist (\cref{sec:methods:deepfeatures}), already fine-tuned Presto encoders (one per random seed, \cref{sec:experiments}) extracted embeddings to obtain nine 128-dimensional embeddings per plot.}

\new{We combine the nine per-position embeddings into a single feature vector per plot in three ways: (i) \emph{flatten}, concatenating all nine embeddings (e.g. 128 x 9 = 1152-dimensional embeddings for input); (ii) \emph{mean-pool}, averaging the nine embeddings element-wise, preserving the original embedding dimensionality (128-dimensional embeddings); and (iii) \emph{stats-pool}, computing, independently for each of the 128 embedding dimensions, the minimum, maximum, mean, and standard deviation of that dimension's value across the nine positions, then concatenating the four resulting 128-dimensional vectors into one 512-dimensional feature (four times the embedding dimensionality, independent of patch size); its mean component is identical to mean-pool, with the minimum, maximum, and standard deviation retaining additional per-dimension spread information across the neighbourhood that mean-pool alone discards. Stats-pooling follows the pooling-strategy benchmark of \citet{Corley_PoolingEarthEmbeddings_2026}, who evaluate exactly the \newreplace{AlphaEarth}{AEF} and TESSERA embedding families and find that statistics-based pooling improves geographic generalisation over plain mean-pooling under spatial distribution shift; spatial-statistic features are also established practice in RF-based land-cover classification more broadly \citep{Ghimire_ContextualRF_2010}. For AEF and TESSERA, each pooled representation is evaluated with the same \ac{RF} protocol used for the center-pixel embeddings (\cref{sec:experiments}). For fine-tuned Presto, the same already fine-tuned, fixed encoder used for the center-pixel result (\cref{sec:experiments}) embeds every neighbour pixel as well and is not re-trained at any point in this analysis; we train a freshly initialised \ac{RF} and \ac{MLP} on the pooled embeddings for every pooling strategy, including mean-pool, so that only the fixed encoder is shared with the center-pixel result and the comparison isolates the effect of pooling itself, rather than confounding it with whether the classifier was ever exposed to pooled-feature statistics during its own training (the reasoning for why the existing center-pixel classifiers cannot simply be reused on the pooled features is given in \cref{app:spatial_patch_classifier_rationale}).}

\subsection{\new{Model interpretability analysis}}
\new{To complement the accuracy-based comparisons above with insight into \emph{what} the model relies on, we conduct a gradient-based, per-band and per-month attribution analysis of the fine-tuned Presto \ac{MLP} model, following the multi-perspective interpretation framework of \citet{Xu_InterpretingMultiTemporalCropMapping_2021} and, in particular, the gradient-backpropagation approach to input-importance estimation used by \citet{Russwurm_SelfAttentionRawOpticalSITS_2020} for optical satellite image time series. }
% \new{Among the pipelines evaluated in this study, only the fine-tuned Presto \ac{MLP} model is differentiable end-to-end from the raw $12\times17$ input to the predicted class score: the \ac{RF}-based pipelines (harmonic + medoid, frozen and fine-tuned Presto, AEF, and TESSERA) all use a Random Forest classifier, which breaks the differentiable chain, so the analysis is restricted to this one representation.}

\new{For each held-out test sample that the model does not use for training process, we compute the derivative of the predicted score for the true class, $\hat{y}_{\text{true}}$, with respect to every element of the $12\times15$ input by backpropagation, $\mathbf{g} = \partial \hat{y}_{\text{true}} / \partial \mathbf{x} \in \mathbb{R}^{12 \times 15}$, giving one value per month and per band (S-1, S-2, and \ac{ERA5}; the static \ac{SRTM} elevation and slope bands have no monthly variation and are excluded from this sum). A positive (negative) value indicates that increasing that input element increases (decreases) the predicted score for the true class while zero value does not affect the prediction. This raw derivative is a signed, real-valued quantity. Because we are interested in \emph{how strongly} the model responds to a given month or band, irrespective of direction, we report its magnitude $|\mathbf{g}|$ throughout this analysis rather than the signed value. This also avoids a cancellation artefact when aggregating: summing signed derivatives across the 15 dynamic bands to obtain one value per month could let oppositely-signed contributions offset one another and understate genuine model sensitivity in that month. We restrict the analysis to correctly classified samples so that the attribution reflects what the model relies on when it succeeds and average $|\mathbf{g}|$ per species to obtain a per-species, per-month, per-band sensitivity profile. Test-set sizes for each dataset are detailed in \cref{tab:class_distribution_splits}.}

\new{We complement this model-derived attribution with an analysis of the underlying phenological signal: per-species mean monthly trajectories, with 95\% confidence intervals of the class mean, for each input band, following the data-driven interpretability approach recently applied to tree-species satellite image time series by \citet{Bressant_InterpretabilityUrbanTreeSITS_2026}. Both analyses are reported for the 7-class SIMB dataset (\cref{sec:interpretability_results}), which offers the clearest per-species phenological signal among input bands.}

\subsection{\new{National-scale wall-to-wall mapping}}
\label{sec:national_map_methods}

\new{To complementarily assess whether the pre-computed embeddings evaluated above can support operational, country-scale products, we applied the same trained classifiers to every forest pixel in the Netherlands rather than only the labelled plots.}

\new{The mapped domain is restricted to forest using a 10\,m Copernicus High Resolution Layer Forest Type mask (2021 vintage), clipped to the Netherlands' administrative boundary, yielding 37,880,448 forest pixels (378,805\,ha; this remote-sensing-derived estimate differs slightly from the \ac{NFI}7-reported 363,801\,ha, \cref{sec:study_area}, a discrepancy of the same order as those documented between remote-sensing- and inventory-based forest-area estimates elsewhere \citep{francini_DutchForestSpeciesMap_2024}). }

\new{Map accuracy is assessed following \citet{Olofsson_GoodPracticesAreaEstimation_2014}: overall, user's, and producer's accuracy, and per-class area, each with a 95\% confidence interval, computed from a stratified, area-weighted confusion matrix built from the held-out 30\% test data used for the point-level results (never used in training), classified by the same trained model that produced the map. Assessing accuracy from a held-out partition never seen during training follows \citet{hermosilla_CanadaTreeSpeciesMapping_2022} and \citet{blickensdorfer_GermanTreeSpeciesMap_2024}, rather than the out-of-bag estimate used by \citet{francini_DutchForestSpeciesMap_2024} (\cref{sec:national_map_discussion}). We additionally compare the resulting maps against the \ac{NFI} using \citeauthor{francini_DutchForestSpeciesMap_2024}'s own $10\times10$\,km grid-cell method: per-cell species-area proportions from the classified map are compared against per-cell proportions derived from \ac{NFI} basal-area data, independently of whichever data trained the classifier being evaluated. To implement a relatively comparable analysis with \citeauthor{francini_DutchForestSpeciesMap_2024}'s one, we test the 7-classes datasets (SIMB and SIBA).}

\new{Before mapping every forest pixel, we first assessed map soundness on a cheaper random sample of 1M forest pixels (about one sampled pixel per 60\,m within forest), classified with candidate methods based on the model performance. }
% For AEF and TESSERA this 1M-pixel sample reproduces the full-census overall accuracy and \ac{NFI}-area deviation to within about one percentage point (\cref{app:national_map_1M_validation}), confirming it as a low-cost proxy for the full map; we then produced the full wall-to-wall maps for AEF and TESSERA.

%%%%%%%%%%%%%%%%%%%%%%%%%
%%%%%%%%%%%%%%%%%%%%%%%%%
\section{Results}
\label{sec:results}

\subsection{Influence of multi-source input features in harmonic and medoid features}
\label{sec:results:harmfeatures}

Following the approach of \citet{francini_DutchForestSpeciesMap_2024}, \newreplace{we first evaluated the contribution of S-1 radar data to optical S-2 classification. Using \ac{RF} with harmonic + medoid features, the addition of S-1 provided modest improvements: 0.4\% for the 13-class COMB dataset and 1.4\% for the 7-class SIMB dataset. These results justify the inclusion of S-1 features in our subsequent comparisons.

W}{w}e first evaluated whether combining optical and radar satellite data improves tree species classification using hand-crafted features in a \ac{RF} classifier. \newreplace{Feature sets were derived from seasonal medoid composites and harmonic fits, using either S-2 alone or in combination with S-1. }The goal \newreplace{was}{is} to assess whether radar backscatter, which captures structural and moisture-related canopy properties, offers complementary information to optical reflectance data.

As shown in Table \ref{tab:combination_S1_S2} the inclusion of S-1 provided modest but consistent improvements across both datasets. For the 13-class dataset (COMB), accuracy increased from 66.34\% to 66.71\% when adding S-1 to S-2 derived features. In the 7-class dataset (SIMB), accuracy improved from 72.87\% to 74.22\%. 

These results suggest that the radar signal contributes marginally to distinguishing between species, especially under a more simplified class grouping. While the absolute gains remain small, incorporating S-1 ensures a more robust feature representation, particularly in settings where optical signals may be limited by cloud cover or seasonal variability. \newreplace{The modest accuracy gains observed here are also useful to contextualize the more substantial differences in performance reported in the next sections, which stem from the feature representations rather than the data sources themselves.}

\begin{table*}[htbp]
    \centering
    \small
    \caption{Classification accuracy comparison across different feature combinations and satellite data sources in \ac{NFI} data (in \%). Bold values indicate best performance for each label type.}
    \label{tab:combination_S1_S2}
    \begin{threeparttable}
    \begin{tabular}{
        >{\raggedright}p{2.0cm}
        >{\raggedright}p{2.4cm}
        >{\centering}p{0.8cm}
        >{\raggedleft}p{1.2cm}
        >{\centering}p{0.8cm}
        >{\raggedleft}p{1.2cm}
        >{\centering}p{1.4cm}
        >{\raggedleft\arraybackslash}p{1.2cm}}
    \toprule
    Label & Features & \multicolumn{2}{c}{Sentinel-1} & \multicolumn{2}{c}{Sentinel-2} & \multicolumn{2}{c}{\makecell[c]{S-1 + S-2\\Combined}} \\
    \cmidrule(lr){3-4} \cmidrule(lr){5-6} \cmidrule(lr){7-8}
    & & Band & Acc. & Band & Acc. & Band & Acc. \\
    \midrule
    \multirow{3}{*}{\makecell[l]{COMB\\(13 classes)}} 
    & Seasonal (S) & S-1 & 51.03 & S-2 & 64.03 & S-1 + S-2 & 64.52 \\
    & Harmonic (H) & S-1 & 55.04 & S-2 & 63.79 & S-1 + S-2 & 64.64 \\
    & All (S+H) & S-1 & 57.72 & S-2 & 66.34 & S-1 + S-2 & \textbf{66.71} \\
    \midrule
    \multirow{3}{*}{\makecell{SIMB\\(7 classes)}}
    & Seasonal (S) & S-1 & 55.60 & S-2 & 70.18 & S-1 + S-2 & 71.93 \\
    & Harmonic (H) & S-1 & 60.19 & S-2 & 71.39 & S-1 + S-2 & 71.52 \\
    & All (S+H) & S-1 & 62.75 & S-2 & 72.87 & S-1 + S-2 & \textbf{74.22} \\
    \bottomrule
    \end{tabular}
    \end{threeparttable}
\end{table*}

\subsection{\new{Pixel-level classification performance evaluation}}
\label{sec:pixel_level_evaluation}

We compared classification performance between traditional hand-crafted features (harmonic and medoid features)\new{, deep learning model from scratch and} several deep embeddings as described in \newreplace{t}{T}able \ref{tab:RQ1_combined}. \new{\Cref{tab:input_mapping} (Appendix) summarizes the input, feature representation, and downstream classifier for each approach compared here.} 

\subsubsection{Deep Embeddings vs Harmonic features}
\label{sec:deepembeddings_vs_harmonic}

We first analyze the effect of different feature representations while fixing the classifier to Random Forest (RF), allowing a direct comparison between deep embeddings and traditional harmonic + medoid features, as well as between frozen and fine-tuned Presto representations.

% RF comparison
Across all datasets, RF models trained on deep embeddings \newreplace{consistently}{mostly} outperform \newreplace{those using}{the RF model trained on} harmonic + medoid features. \new{The exeptions of this trend are TESSERA (no-spatial pooling) in SIBA and the frozen Presto for all datasets.}
In particular, results obtained from embeddings with the RF classifier show that fine-tune\new{d} Presto embeddings consistently outperform frozen Presto, AEF, and TESSERA embeddings across all datasets and evaluation metrics, including overall accuracy, F1 macro, and F1 weighted scores. This highlights the benefit of task-specific fine-tuning for adapting general-purpose deep embeddings to forest species classification.  

    % SIMB
    On the SIMB dataset, which is small and moderately imbalanced, RF with fine-tuned Presto embeddings achieved an overall accuracy of \newreplace{73.02\% (±1.55)}{78.59\% (±1.86)}, an F1 macro score of \newreplace{57.60\% (±1.74)}{68.46\% (±2.13)}, and an F1 weighted score of \newreplace{71.80\% (±1.69)}{78.17\% (±2.00)}, compared to 69.48\% (±0.73) accuracy, 51.03\% (±1.71) F1 macro, and 66.98\% (±1.06) F1 weighted for harmonic + medoid features. This corresponds to gains of approximately \newreplace{3 percentage points in accuracy, over 6 points in F1 macro, and nearly 5 points in F1 weighted}{9 percentage points in accuracy, 17 points in F1 macro, and 11 points in F1 weighted}, highlighting the advantage of learned representations in low-sample regimes while accounting for class frequency.

    % COMB
    For the more challenging COMB dataset, characterized by a larger number of classes and stronger class imbalance, the performance gap remains evident. RF with fine-tuned Presto embeddings reached \newreplace{65.91\% (±1.72)}{70.29\% (±0.60)} overall accuracy, \newreplace{46.11\% (±3.00)}{53.23\% (±1.34)} F1 macro, and \newreplace{62.34\% (±1.97)}{67.96\% (±0.70)} F1 weighted, compared to 62.80\% (±0.91) accuracy, 38.95\% (±2.02) F1 macro, and 57.13\% (±1.25) F1 weighted using harmonic + medoid features. The substantial improvement of \newreplace{3 points in accuracy, 7 points in F1 macro and 5 points in F1 weighted}{about 7 points in accuracy, 14 points in F1 macro, and 11 points in F1 weighted} indicates that deep embeddings better capture discriminative information for minority classes under increased class complexity, while gains in F1 weighted reflect consistent improvements across dominant classes.

    % SIBA
    The largest absolute performance differences are observed on the SIBA dataset, which is both balanced and substantially larger. Here, RF with fine-tuned Presto embeddings achieved \newreplace{93.37\% (±0.69)}{94.50\% (±0.45)} accuracy, \newreplace{93.33\% (±0.69) F1 macro}{94.48\% (±0.45) F1 macro}, and \newreplace{93.33\% (±0.69) F1 weighted}{94.48\% (±0.45) F1 weighted}, compared to 84.47\% (±0.41) accuracy, 84.40\% (±0.40) F1 macro, and 84.40\% (±0.40) F1 weighted for harmonic + medoid features. The gains of \newreplace{nearly 9 points}{over 10 points} across evaluation matrices reflect the balanced class distribution of SIBA and demonstrate that deep embeddings scale effectively with data availability and clearer class separation.

    % Overall
    Within RF-based deep models, fine-tuning Presto consistently improves performance over frozen embeddings, particularly for SIMB and SIBA. The gains are more modest for COMB, suggesting that while fine-tuning enhances representation quality, class imbalance and higher label complexity still limit achievable performance.

    For direct comparison between fine-tune\new{d} Presto and Harmonic + medoid features using identical S-1 + S-2 sensor inputs (detailed in Appendix Table \ref{tab:RQ1_results1_S1S2only}), deep embeddings show F1 score improvements of \newreplace{9.0}{19.1} percentage points (SIMB), \newreplace{5.7}{16.6} percentage points (COMB), and \newreplace{7.5}{10.2} percentage points (SIBA), confirming consistent performance advantages across different input configurations for embeddings. Appendix Table \ref{tab:RQ1_results3_All_S1S2_comparison} compares models using all satellite data against those using only S-1 and S-2 data for fine-tuned Presto embeddings with the MLP classifier. \newreplace{The results show no significant differences in input data for fine-tuned Presto embeddings.}{The results show only modest differences: minimal for SIMB and COMB, and minimal for SIBA as well, with the two band configurations effectively indistinguishable there too.}

\subsubsection{MLP vs RF classifier}
\label{sec:mlp_vs_rf}
    %MLP vs RF
    We next examine the influence of classifier architecture by comparing the MLP classifier for tree species classification \newreplace{}{proposed by }\citet{Mouret_TreeSpeciesClassificationPixelTimeSeriesInbalanced_2024} with an RF classifier, both trained on the same deep embeddings extracted from the fine-tuned Presto model. This isolates the effect of the classifier while keeping the input representation fixed.

    \newreplace{Across all datasets, the MLP classifier slightly outperforms RF when using fine-tuned Presto embeddings, with the magnitude of the performance differences varying according to dataset size and class balance.}{Across the datasets, the MLP and RF classifiers are closely competitive when using fine-tuned Presto embeddings, and their relative ordering depends on dataset size: RF matches or slightly exceeds MLP on the small NFI datasets, whereas MLP pulls clearly ahead on the large, balanced SIBA dataset.}

    On the SIMB dataset, MLP achieved \newreplace{73.67\% (±1.29)}{78.05\% (±2.20)} accuracy, \newreplace{62.29\% (±4.60)}{67.94\% (±3.33)} F1 macro, and \newreplace{73.05\% (±1.52)}{77.67\% (±2.32)} F1 weighted, compared to RF’s \newreplace{73.02\% (±1.55)}{78.59\% (±1.86)} accuracy, \newreplace{57.60\% (±1.74)}{68.46\% (±2.13)} F1 macro, and \newreplace{71.80\% (±1.69)}{78.17\% (±2.00)} F1 weighted. \newreplace{While overall accuracy is comparable and F1 macro exhibits higher variance due to the small and imbalanced nature of SIMB, MLP attains marginally higher F1 macro and weighted scores than RF, indicating slightly improved average performance across classes and stronger performance on majority classes, resulting in better overall balance when accounting for class frequency.}{}
    % {The two classifiers are statistically indistinguishable on this dataset: the differences are well below one point and smaller than the seed-to-seed standard deviation, so neither classifier holds a meaningful advantage in this small, imbalanced regime.}

    \newreplace{For the COMB dataset, the advantage of MLP becomes more pronounced.}{On the COMB dataset, RF is the stronger classifier.} MLP achieved \newreplace{67.40\% (±1.47)}{69.35\% (±1.08)} accuracy, \newreplace{50.76\% (±2.23)}{48.59\% (±2.20)} F1 macro, and \newreplace{65.51\% (±1.29)}{66.53\% (±1.08)} F1 weighted, \newreplace{outperforming}{compared with} RF’s \newreplace{65.91\% (±1.72)}{70.29\% (±0.60)} accuracy, \newreplace{46.11\% (±3.00)}{53.23\% (±1.34)} F1 macro, and \newreplace{62.34\% (±1.97)}{67.96\% (±0.70)} F1 weighted. \newreplace{This suggests that the non-linear decision boundaries learned by MLPs are slightly better suited for complex, multi-class classification tasks with imbalanced distributions.}{}
    % {On this small, highly imbalanced 13-class task, RF, which is low-variance and robust on limited samples, reads the fine-tuned embeddings more effectively than the MLP head, most clearly in F1 macro (a gap of about 4.6 points), indicating better handling of minority classes.}

    \newreplace{In the SIBA dataset, both classifiers achieve strong and nearly identical performance due to the large sample size and balanced class distribution. MLP reached 93.54\% accuracy, F1 macro, and F1 weighted, compared to 93.37\% accuracy, 93.33\% F1 macro, and 93.33\% F1 weighted for RF. The negligible differences indicate diminishing returns from classifier complexity in data-rich scenarios.}{On the large, balanced SIBA dataset the MLP head pulls clearly ahead, reaching 97.26\% (±0.43) accuracy and 97.26\% (±0.43) F1 macro and weighted, compared with 94.50\% (±0.45) accuracy and 94.48\% (±0.45) F1 macro and weighted for RF, an advantage of about 2.8 points that exceeds the seed-to-seed variation.}    

    \newreplace{Overall, while RF offers robust and competitive performance—particularly in smaller datasets, MLPs consistently extract slightly more discriminative power from deep embeddings, especially in more complex or imbalanced classification settings, as reflected by modest improvements in both F1 macro and F1 weighted scores.}{Overall, the choice of downstream classifier matters far less than the representation: on the small NFI datasets, RF and the MLP head are statistically tied, and the MLP's advantage emerges only on the large, balanced SIBA dataset. }
    % In other words, once a strong fine-tuned representation is available, a robust classifier such as RF is sufficient under label scarcity, whereas the deep classifier head pays off mainly with data volume.}

\subsubsection{from-scratch vs deep embeddings}
\label{sec:from_scratch_vs_deep_embeddings}
    \new{Finally, the two from-scratch deep-learning baselines (TempCNN and the from-scratch MLP), trained directly on the raw $12\times17$ time series, are competitive with AEF and TESSERA embeddings, but underperform the fine-tuned Presto models on the small NFI datasets (SIMB).} \new{For example, on SIMB they reach 73.47\% (TempCNN) and 73.24\% (from-scratch MLP) overall accuracy, similarly to AEF (72.74\%) and TESSERA (73.65\%), but below the 78.59\% of fine-tuned Presto with RF. }
    % \new{This supports the premise that self-supervised pre-training is most valuable when labelled data are scarce. }
    \new{On the large SIBA dataset these baselines appear better under the standard pixel-level split (92--98\% overall accuracy), but this largely reflects within-polygon spatial autocorrelation: their performance drops sharply under an extreme spatially blocked (plot-grouped) evaluation (\cref{tab:siba_spatial}).}

% Marc: I removed the data size column to make it fit the paper width
\begin{table*}[ht]
    \centering
    \footnotesize
    \setlength{\tabcolsep}{2pt}
    \caption{Classification performance comparison (in \%) across feature types (deep embedding vs. harmonic + medoid) and classifiers \newreplace{(MLP vs. RF)}{(MLP, RF, and from-scratch deep-learning baselines)}. Each model was run five times and the mean and standard deviation (±) are reported. \newreplace{Bold values indicate the best performance among MLP and RF per dataset type.}{Bold values indicate the best performance per dataset and metric across all classifiers.} \new{TempCNN and MLP (scratch) are trained from random initialisation on the raw time series; for SIBA these pixel-level scores are spatially optimistic owing to within-polygon autocorrelation, and a spatially blocked (plot-grouped) evaluation is reported in \cref{tab:siba_spatial}.}}
    \label{tab:RQ1_combined}
    \begin{threeparttable}
    \begin{tabular}{
        >{\raggedright}p{1.0cm}
        >{\raggedright}p{2.3cm}
        >{\centering}p{1.5cm}
        >{\centering}p{1.5cm}
        >{\centering}p{1.5cm}
        >{\centering}p{1.5cm}
        >{\centering}p{1.5cm}
        >{\centering}p{1.5cm}
        >{\centering}p{1.5cm}
        >{\centering\arraybackslash}p{1.5cm}}
    \toprule
    & & \multicolumn{8}{c}{Classification Results} \\
    \cmidrule(lr){3-10}
    & & MLP & \multicolumn{5}{c}{Random Forest} & \multicolumn{2}{c}{\new{From scratch}} \\
    \cmidrule(lr){3-3}\cmidrule(lr){4-8}\cmidrule(lr){9-10}
    & Dataset type & Fine-tune Presto with MLP Classifier & Presto\newline (Frozen) & Fine-tune Presto & Harm.+med.\newline \citep{francini_DutchForestSpeciesMap_2024} & AEF \newline (Frozen ) \newline \citep{brown2025alphaearth} & TESSERA \newline (Frozen) \newline \citep{feng2025tessera} & \new{TempCNN\newline (scratch)\newline \citep{Pelletier_TempCNN_2019}} & \new{MLP\newline (scratch)} \\
    \midrule
    \multirow{3}{*}{\begin{tabular}{@{}l@{}}Overall\\Accuracy\end{tabular}} 
    & SIMB (7 classes)
        & \newreplace{\textbf{73.67 ± 1.29}}{78.05 ± 2.20}
        & 69.39 ± 2.10 
        & \newreplace{73.02 ± 1.55}{\textbf{78.59 ± 1.86}} 
        & 69.48 ± 0.73 
        & 72.74 ± 1.61 
        & 73.65 ± 1.28 & \new{73.47 ± 1.38} & \new{73.24 ± 1.14} \\
    & COMB (13 classes)
        & \newreplace{\textbf{67.40 ± 1.47}}{69.35 ± 1.08}
        & 60.68 ± 1.51 
        & \newreplace{65.91 ± 1.72}{\textbf{70.29 ± 0.60}} 
        & 62.80 ± 0.91 
        & 62.71 ± 0.56 
        & 65.28 ± 1.61 & \new{64.65 ± 1.07} & \new{63.93 ± 2.34} \\
    & SIBA (7 classes)
        & \newreplace{\textbf{93.54 ± 1.27}}{97.26 ± 0.43}
        & 84.51 ± 0.33
        & \newreplace{93.37 ± 0.69}{94.50 ± 0.45}
        & 84.47 ± 0.41
        & 90.32 ± 0.65
        & 75.65 ± 0.64 & \new{92.19 ± 0.46} & \new{\textbf{98.32 ± 0.26}} \\
    \midrule
    \multirow{3}{*}{\begin{tabular}{@{}l@{}}F1\\Macro\end{tabular}}
    & SIMB (7 classes)
        & \newreplace{\textbf{62.29 ± 4.60}}{67.94 ± 3.33}
        & 47.17 ± 2.03 
        & \newreplace{57.60 ± 1.74}{\textbf{68.46 ± 2.13}} 
        & 51.03 ± 1.71 
        & 52.67 ± 2.76 
        & 57.36 ± 1.81 & \new{58.49 ± 1.42} & \new{58.37 ± 2.02} \\
    & COMB (13 classes)
        & \newreplace{\textbf{50.76 ± 2.23}}{48.59 ± 2.20}
        & 36.30 ± 2.10 
        & \newreplace{46.11 ± 3.00}{\textbf{53.23 ± 1.34}} 
        & 38.95 ± 2.02 
        & 38.05 ± 1.97 
        & 43.23 ± 3.16 & \new{43.45 ± 3.20} & \new{45.25 ± 2.25} \\
    & SIBA (7 classes)
        & \newreplace{\textbf{93.54 ± 1.26}}{97.26 ± 0.43}
        & 84.32 ± 0.36
        & \newreplace{93.33 ± 0.69}{94.48 ± 0.45}
        & 84.40 ± 0.40
        & 90.20 ± 0.65
        & 75.66 ± 0.62 & \new{92.08 ± 0.49} & \new{\textbf{98.32 ± 0.26}} \\
    \midrule
    \multirow{3}{*}{\begin{tabular}{@{}l@{}}F1\\Weighted\end{tabular}}
    & SIMB (7 classes)
        & \newreplace{\textbf{73.05 ± 1.52}}{77.67 ± 2.32}
        & 66.58 ± 2.19
        & \newreplace{71.80 ± 1.69}{\textbf{78.17 ± 2.00}}
        & 66.98 ± 1.06
        & 70.38 ± 1.78
        & 71.12 ± 1.40 & \new{71.52 ± 1.24} & \new{72.01 ± 1.44} \\
    & COMB (13 classes)
        & \newreplace{\textbf{65.51 ± 1.29}}{66.53 ± 1.08}
        & 55.37 ± 1.67
        & \newreplace{62.34 ± 1.97}{\textbf{67.96 ± 0.70}}
        & 57.13 ± 1.25
        & 57.18 ± 0.88
        & 60.15 ± 1.96 & \new{60.68 ± 1.95} & \new{61.72 ± 1.87} \\
    & SIBA (7 classes)
        & \newreplace{\textbf{93.54 ± 1.26}}{97.26 ± 0.43}
        & 84.32 ± 0.36
        & \newreplace{93.33 ± 0.69}{94.48 ± 0.45}
        & 84.40 ± 0.40
        & 90.20 ± 0.65
        & 75.66 ± 0.62 & \new{92.08 ± 0.49} & \new{\textbf{98.32 ± 0.26}} \\
    \bottomrule
    \end{tabular}
    \end{threeparttable}
\end{table*}

\subsubsection{\new{Effect of spatial context: $n\times n$ patch pooling}}
\label{sec:spatial_patch_results}

\new{\cref{tab:spatial_patch_headline} compares the single center pixel against $3\times3$-neighbourhood mean-pooling (\cref{sec:spatial_patch_methods}) for AEF, TESSERA, and fine-tuned Presto (\ac{RF} and \ac{MLP}), on \ac{NFI} (SIMB, COMB) and on SIBA; the full breakdown across all three pooling strategies under both the pixel-level and plot-grouped (spatially blocked) splits is given in Appendix Table \ref{tab:spatial_patch_aef_tess_full}.}

\new{For AEF, spatial pooling on all datasets yields a modest gain or loss. For TESSERA, however, mean-pooling gives a large, consistent gain everywhere: overall accuracy rises 5.9 points on SIMB (72.96\%~$\to$~78.88\%) and 6.4 points on COMB (65.37\%~$\to$~71.78\%), with even larger F1 macro gains (+11.3 and +11.4 points).}
\new{For fine-tuned Presto, every pooling variant improves over the single-pixel baseline on both \ac{NFI} targets and classifiers -- but gives no benefit on SIBA.}
% : every pooling variant and classifier is statistically indistinguishable from or slightly below the single-pixel baseline on either split, the largest change being a 0.4-point \ac{OA} \emph{drop} (\ac{MLP}, pixel-level, 97.26\%~$\to$~96.88\%).}

\new{Mean-pooling is the most consistently strong of the three pooling strategies for TESSERA and fine-tuned Presto on \ac{NFI}, and for AEF/TESSERA on SIBA's pixel-level split, though not the single best cell in every combination (Appendix Table \ref{tab:spatial_patch_aef_tess_full}); flatten, despite carrying strictly more information (9$\times$ the embedding dimensionality), underperforms mean-pooling in most cases, plausibly because concatenating nine spatially autocorrelated copies of each dimension dilutes the per-split feature subsampling that \ac{RF} relies on at these sample sizes ($\sqrt{p}$ features drawn per split from an input that is now mostly redundant copies). }

\begin{table*}[ht]
    \centering
    \footnotesize
    \setlength{\tabcolsep}{3pt}
    \caption{\new{Effect of $3\times3$ neighbourhood mean-pooling versus the single center pixel on classification performance (in \%), under the standard (pixel-level) split, for AEF, TESSERA, and fine-tuned Presto (\ac{RF} and \ac{MLP} classifiers). Mean and standard deviation (±) over five seeds. Bold values indicate the single highest value per metric (Overall Accuracy, F1 Macro, F1 Weighted) within each dataset. The full breakdown across all three pooling strategies (flatten, mean-pool, stats-pool), including the SIBA plot-grouped (spatially blocked) split, is given in Appendix Table \ref{tab:spatial_patch_aef_tess_full}.}}
    \label{tab:spatial_patch_headline}
    \begin{newtable}
    \begin{threeparttable}
    \begin{tabular}{
        >{\raggedright}p{1.4cm}
        >{\raggedright}p{3.2cm}
        >{\centering}p{1.5cm}>{\centering}p{1.5cm}
        >{\centering}p{1.5cm}>{\centering}p{1.5cm}
        >{\centering}p{1.5cm}>{\centering\arraybackslash}p{1.5cm}}
    \toprule
    & & \multicolumn{2}{c}{Overall Accuracy} & \multicolumn{2}{c}{F1 Macro} & \multicolumn{2}{c}{F1 Weighted} \\
    \cmidrule(lr){3-4}\cmidrule(lr){5-6}\cmidrule(lr){7-8}
    & Embedding & Center & \makecell{$3\times3$\\Mean-pool} & Center & \makecell{$3\times3$\\Mean-pool} & Center & \makecell{$3\times3$\\Mean-pool} \\
    \midrule
    \multirow{4}{*}{\makecell[l]{SIMB\\(7 classes)}}
    & AEF (RF) & 72.06 ± 1.68 & 72.11 ± 1.33 & 52.86 ± 3.08 & 54.69 ± 2.47 & 69.83 ± 1.95 & 70.17 ± 1.56 \\
    & TESSERA (RF) & 72.96 ± 1.42 & 78.88 ± 0.96 & 57.46 ± 1.72 & 68.79 ± 1.27 & 70.40 ± 1.54 & 77.76 ± 1.02 \\
    & Fine-tuned Presto (RF) & 78.59 ± 1.86 & \textbf{81.57 ± 1.29} & 68.46 ± 2.13 & 71.70 ± 1.99 & 78.17 ± 2.00 & \textbf{81.01 ± 1.47} \\
    & Fine-tuned Presto (MLP) & 78.05 ± 2.20 & 81.03 ± 1.33 & 67.94 ± 3.33 & \textbf{72.95 ± 2.22} & 77.67 ± 2.32 & 80.67 ± 1.36 \\
    \midrule
    \multirow{4}{*}{\makecell[l]{COMB\\(13 classes)}}
    & AEF (RF) & 62.62 ± 0.91 & 62.44 ± 1.08 & 36.51 ± 1.71 & 37.63 ± 2.52 & 56.81 ± 1.15 & 56.93 ± 1.24 \\
    & TESSERA (RF) & 65.37 ± 1.83 & 71.78 ± 1.87 & 43.95 ± 2.89 & 55.39 ± 2.79 & 60.59 ± 1.95 & 68.64 ± 2.02 \\
    & Fine-tuned Presto (RF) & 70.29 ± 0.60 & 71.56 ± 1.25 & 53.23 ± 1.34 & 54.47 ± 1.84 & 67.96 ± 0.70 & 68.97 ± 1.51 \\
    & Fine-tuned Presto (MLP) & 69.35 ± 1.08 & \textbf{72.10 ± 1.07} & 48.59 ± 2.20 & \textbf{57.69 ± 1.61} & 66.53 ± 1.08 & \textbf{70.86 ± 1.00} \\
    \midrule
    \multirow{4}{*}{\makecell[l]{SIBA\\(7 classes)}}
    & AEF (RF) & 88.86 ± 0.55 & 90.67 ± 0.40 & 88.74 ± 0.55 & 90.56 ± 0.41 & 88.74 ± 0.55 & 90.56 ± 0.41 \\
    & TESSERA (RF) & 75.33 ± 0.68 & 87.71 ± 0.83 & 75.33 ± 0.67 & 87.67 ± 0.85 & 75.33 ± 0.67 & 87.67 ± 0.85 \\
    & Fine-tuned Presto (RF) & 94.50 ± 0.45 & 94.48 ± 0.41 & 94.48 ± 0.45 & 94.46 ± 0.41 & 94.48 ± 0.45 & 94.46 ± 0.41 \\
    & Fine-tuned Presto (MLP) & \textbf{97.26 ± 0.43} & 96.88 ± 0.17 & \textbf{97.26 ± 0.43} & 96.88 ± 0.17 & \textbf{97.26 ± 0.43} & 96.88 ± 0.17 \\
    \bottomrule
    \end{tabular}
    \end{threeparttable}
    \end{newtable}
\end{table*}

\subsection{Class level performances}
\label{sec:class_level_performance}

\begin{figure*}[ht]
    \centering
    \includegraphics[width=19cm]{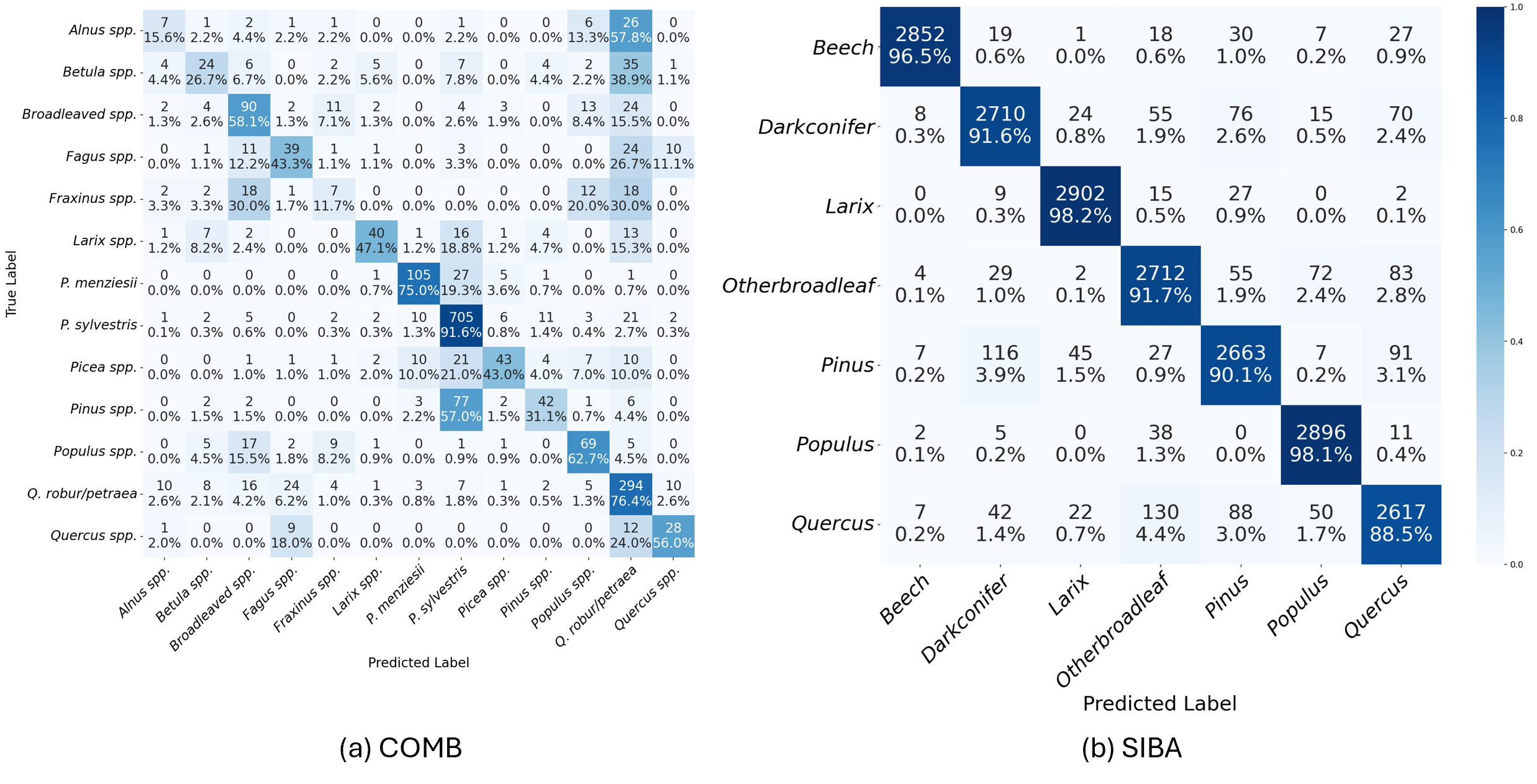}
    \caption{\textit{Confusion matrices for the best performing models with deep embeddings by \ac{MLP} classifier. \textbf{Complex \& Imbalanced (COMB)}(left), \textbf{Simplified \& Balanced (SIBA)} (right). Each cell has average count and recall percentage.}}
    \label{fig:ConfusionMatrix_MLP_deepfeature}
\end{figure*}

To better understand class-level behavior, we visualized the confusion matrices of the best-performing model -- an MLP fine-tuned on deep embeddings -- for two datasets (COMB, SIBA). These matrices are shown in Figure \ref{fig:ConfusionMatrix_MLP_deepfeature}, with each cell reporting the number of predicted samples and the corresponding recall (i.e., percentage of correct predictions among all true samples for that class).

In the complex 13-class dataset (COMB), misclassifications are pronounced. Within the broadleaved group (\textit{Alnus spp.}, \textit{Betula spp.}, “Broadleaved spp.”, \textit{Fagus spp.}, \textit{Fraxinus spp.}, \textit{Populus spp.}, \textit{Q. robur/petraea}, \textit{Quercus spp.}), there is substantial confusion among species whose phenology or seasonal spectral signatures overlap. For example, \textit{Q. robur/petraea}, being the most abundant class in this group, serves as a frequent “sink” — many other broadleaved species are misclassified as oak. Less common broadleaved classes, such as \textit{Alnus spp.} and \textit{Fraxinus spp.}, often suffer from very poor classification performance. In our results, \textit{Alnus spp.} achieves only 2.2 \% recall, with 62 \% of its true instances misclassified as \textit{Q. robur/petraea}. Phenologically, both oak (\textit{Q. robur/petraea}) and alder (\textit{Alnus spp.}) are classified among the late-leafing deciduous species according to remote-sensing studies \citep{Grabska-szwagrzyk_Sentinel2TimeSeriesPhenology_2024}. In addition, species such as ash (\textit{Fraxinus spp.}) are known to be difficult to discriminate reliably from other broadleaved deciduous species in remote-sensing based classification \citep{Lisein_DiscriminationDeciduousTree_2015}. Because of these overlapping phenological and spectral patterns, misclassification among broadleaved species is observed.

With conifer species (\textit{Larix spp}, \textit{P. menziesii}, \textit{P. sylvestris}, \textit{Picea spp} and \textit{Pinus spp}), this group often exhibits markedly different spectral and phenological behavior compared to deciduous broadleaves, which can facilitate broad-leaf vs conifer discrimination \citep{Vorovencii_LocalscaleMappingTree_2023}. However, misclassification still occurs within the conifer group. For instance, the majority of \textit{Pinus spp} were misclassified as \textit{P. sylvestris}, which is also a type of Pinus (pine). Thus, while the model likely performs well at the coarse level (broadleaved vs conifer), its reliability decreases when discriminating among species within each group, especially for rare species and those with phenological and spectral overlap.

In the 7-class NFI dataset (SIMB), the model exhibits generally strong performance across most classes. Recall values range from approximately 90 \% up to 98 \% in several cases. For the aggregated class \textit{DarkConifer} and \textit{Populus}, the model yields a recall of around 98 \%. Within the broadleaved aggregate classes, performance is more variable. Most of the residual confusion is observed among \textit{Quercus} (Oak), \textit{Beech}, and \textit{Other Broadleaf}, reflecting the inherent spectral and phenological similarity within the broadleaved group.  

\subsection{\new{Phenological trajectories and gradient-based temporal attribution}}
\label{sec:interpretability_results}

\begin{figure*}[ht]
    \centering
    \includegraphics[width=18cm]{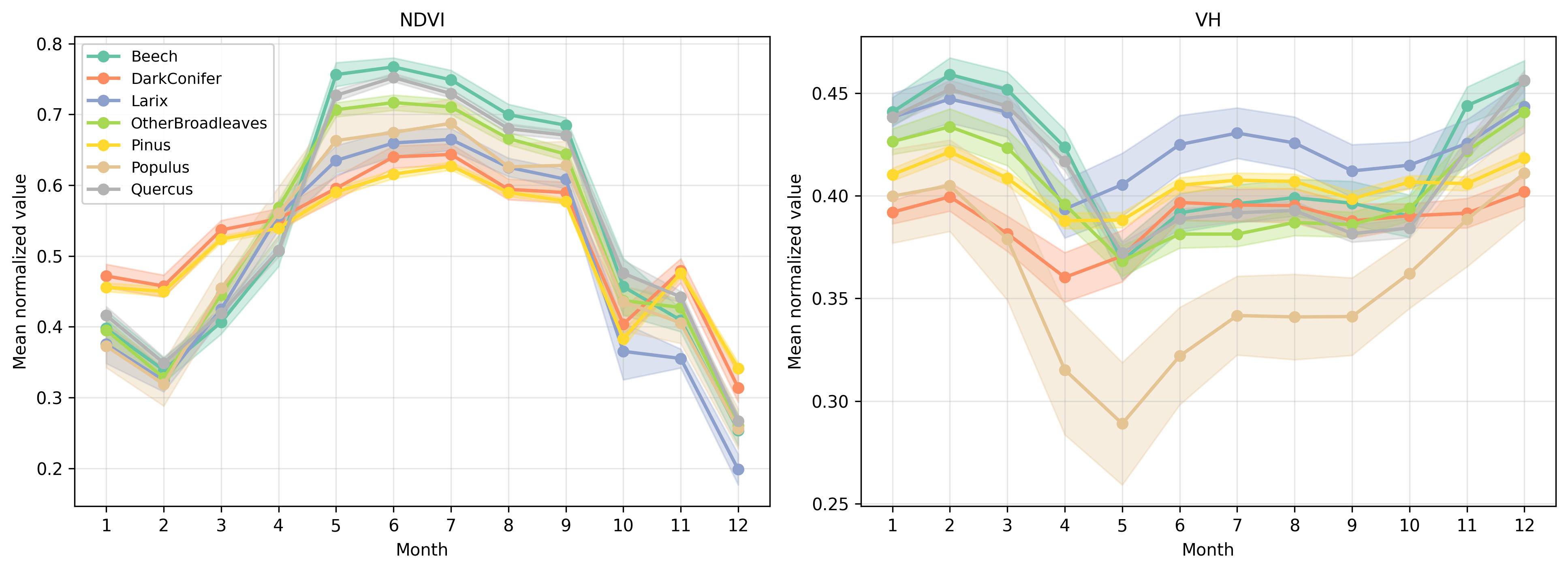}
    \caption{\new{\textit{Per-species mean monthly trajectories (SIMB, 7 classes; shaded band: 95\% confidence interval of the class mean) for \ac{NDVI} (left) and Sentinel-1 VH backscatter (right).}}}
    \label{fig:phenology_trajectory}
\end{figure*}

\new{\cref{fig:phenology_trajectory} shows per-species mean monthly trajectories (SIMB, 7 classes; shaded band: 95\% confidence interval of the class mean) for two representative bands: \ac{NDVI} and Sentinel-1 VH backscatter. \ac{NDVI} separates the deciduous broadleaved classes (Quercus, Beech, Other Broadleaves, and Populus), which show a stronger growing-season peak, from the conifer classes (Pinus, DarkConifer, and Larix). Sentinel-1 VH additionally isolates Populus, whose backscatter drops markedly relative to all other species between April and September. A structural signal that \ac{NDVI} alone does not carry illustrates the complementary value of combining optical and radar inputs. The corresponding trajectories for all 15 dynamic input bands are provided in \cref{app:phenology_annex}.}

\begin{figure}[ht]
    \centering
    \includegraphics[width=\linewidth]{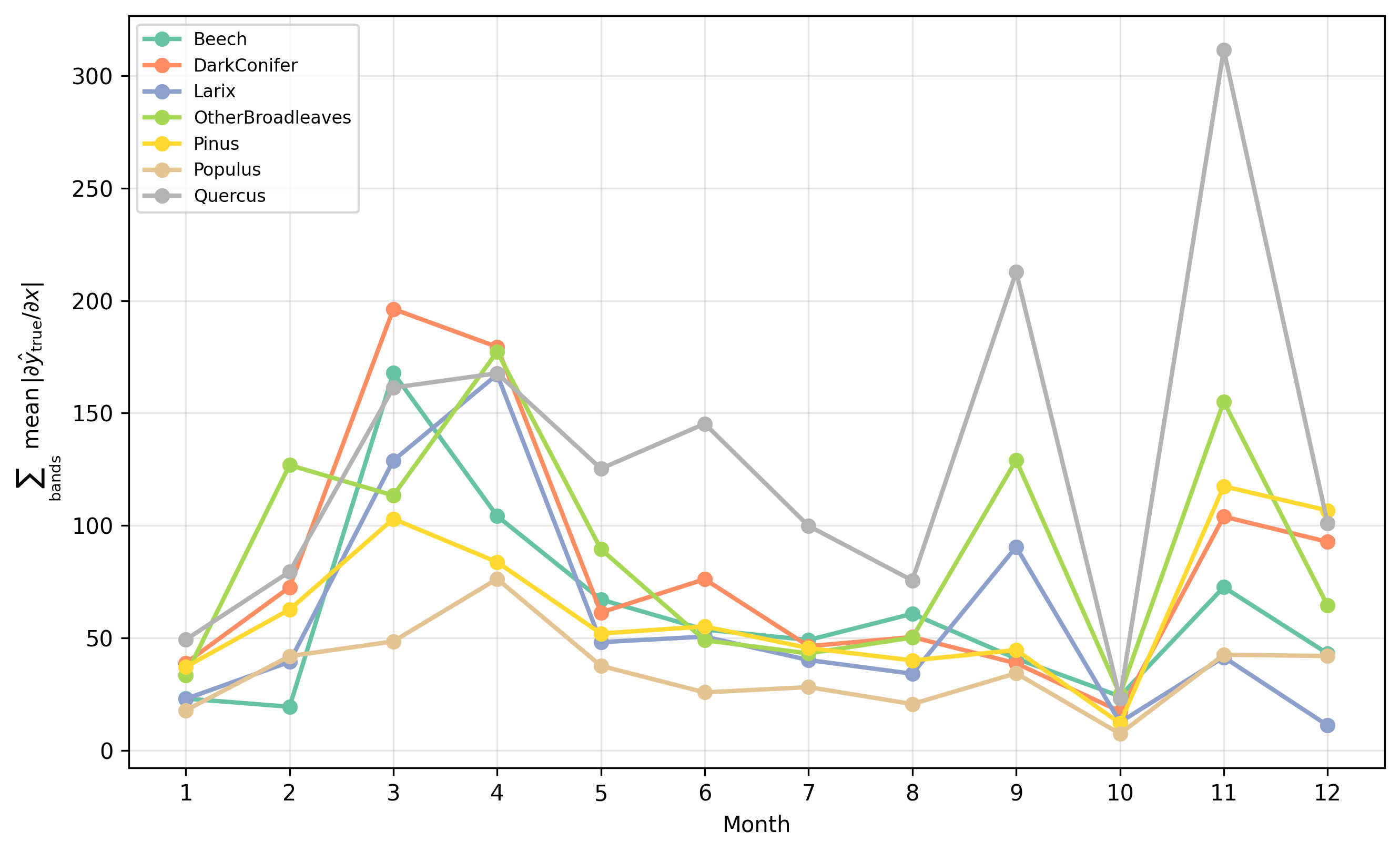}
    \caption{\new{\textit{Gradient-based temporal-response attribution for the fine-tuned Presto \ac{MLP} model (SIMB, correctly classified test samples, rs~=~42): per-species $\sum_{\text{bands}} \text{mean}\,|\partial \hat{y}_{\text{true}}/\partial x|$ across the 15 dynamic input bands, by month.}}}
    \label{fig:gradient_temporal_response}
\end{figure}

\new{\cref{fig:gradient_temporal_response} shows the gradient-based temporal-response profile of the fine-tuned Presto \ac{MLP} model (\cref{sec:experiments}), summed across the 15 dynamic input bands for correctly classified test samples of one representative seed. Model sensitivity shows three peaks across most species: a spring peak in March--April, coinciding with the onset of spring green-up visible in \cref{fig:phenology_trajectory}; a secondary rise in September; and, for several species, a further peak in November. These three windows are separated by a drop in October shared by every species, and a shallower, more variable minimum around May--August. This indicates that the fine-tuned model's temporal sensitivity concentrates in specific phenological transition periods -- spring green-up, an early-autumn transition, and the onset of leaf-fall, providing direct model-level evidence for the phenology-driven classification behaviour that \cref{fig:phenology_trajectory} shows in the raw signal.}

\begin{figure*}[ht]
    \centering
    \includegraphics[width=18cm]{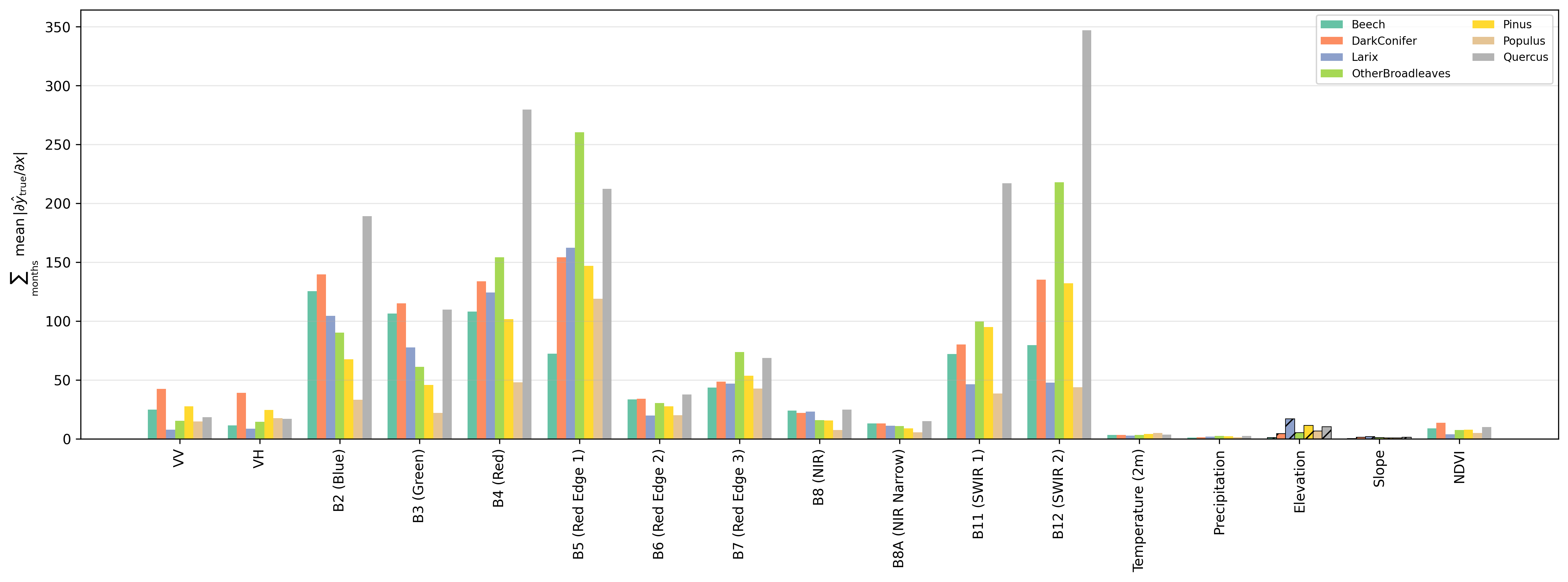}
    \caption{\new{\textit{Gradient-based band importance for the fine-tuned Presto \ac{MLP} model (SIMB, correctly classified test samples, rs~=~42), grouped by species within each band: per-species $\sum_{\text{months}} \text{mean}\,|\partial \hat{y}_{\text{true}}/\partial x|$ for each of the 17 input bands (hatched bars: static \ac{SRTM} bands).}}}
    \label{fig:gradient_band_importance_grouped}
\end{figure*}

\new{\cref{fig:gradient_band_importance_grouped} complements this temporal view with the per-band breakdown, summed over months and grouped by species within each band for direct cross-species comparison. The S-2 short-wave infrared (B11, B12), red (B4), and red-edge/blue (B2, B5) bands dominate the attribution for every species, while \ac{NDVI} itself contributes comparatively little, indicating that the fine-tuned model draws more directly on raw reflectances than on the derived vegetation index. \ac{ERA5} and \ac{SRTM} bands show comparatively low gradient magnitude across species, corroborated by the S-1+S-2-only ablation in Appendix Table \ref{tab:RQ1_results3_All_S1S2_comparison}, where excluding these auxiliary bands entirely changes accuracy by less than one percentage point on all three datasets, including the larger SIBA dataset. Quercus shows markedly higher gradient magnitude than the other six species on several of these bands (B2, B4, B5, B11, and B12). A plausible interpretation is that discriminating Quercus draws more heavily on fine-grained spectral information, consistent with Quercus belonging to the broadleaved group (Quercus, Beech, Other Broadleaves) that shows the most residual confusion in the SIMB confusion matrix (\cref{sec:class_level_performance}).} 

\new{We repeated this gradient-based attribution analysis on the more complex, 13-class COMB dataset as a robustness check (\cref{app:comb_gradient}). The temporal-response profile (\cref{fig:gradient_temporal_response_comb}) reproduces the same three-window structure found for SIMB, most clearly for the two best-sampled classes: \textit{Pinus sylvestris} peaks sharply in March--April, and \textit{Quercus robur/petraea}  shows elevated attribution in both September and November, where its attribution goes beyond every other species-month combination in the figure. All classes share a drop in October. The corresponding per-band breakdown (\cref{fig:gradient_band_importance_grouped_comb}) reproduces the SWIR/red/red-edge dominance and low \ac{NDVI} weight found for SIMB, alongside additional noise from COMB's many small classes.}

\subsection{\new{National-scale species mapping}}
\label{sec:national_map_results}

\subsubsection{\new{1M-pixel mapping before full mapping}}
\label{sec:1M_pixel_mapping}
\new{We first evaluate candidate methods on the randomly extracted 1M-pixel sample (\cref{sec:national_map_methods}), which exposes a failure mode that the plot-level accuracy alone does not. We contrast two encoder-adapting deep models (fine-tuned Presto + RF, and a from-scratch \ac{MLP}) with three fixed-feature methods that keep their representation frozen and only fit the downstream \ac{RF} (the pre-computed AEF and TESSERA embeddings, and the hand-crafted harmonic + medoid features) based on \cref{tab:RQ1_combined}. For TESSERA, the same $3\times3$ mean-pooling method is used (\cref{tab:spatial_patch_headline}). }
\new{For each method, the same \ac{RF} classifier is trained on the same 70\% train partition used for the pixel-level results above (\cref{tab:RQ1_combined}), once for SIMB (\ac{NFI}, 1,479 plots) and once for SIBA (13,790 pixels, pixel-level split).}
% \new{AEF is classified based on pixel-trained model while TESSERA is applied $3\times3$ mean-pooling (\cref{sec:spatial_patch_methods}), since this project's own spatial-pooling result (\cref{sec:spatial_patch_results}) shows improvement.}

\new{On the \ac{NFI} (SIMB) source, all methods produce plausible maps (\cref{fig:national_map_1M_simb}). Most methods achieve higher overall accuracy than the hand-crafted one (except AEF; \cref{tab:national_map_1M}), and lower grid-cell species-proportion deviation from the NFI (\cref{tab:national_map_1M_gridcell}).}
% , and the two deep models in fact match the \ac{NFI} area distribution slightly better than the fixed-feature methods (mean per-class area deviation 3.8--4.6\,\% versus 5.1--5.8\,\%). 

\new{On SIBA, however, the two encoder-adapting models degenerate: they classify 61\,\% (fine-tuned Presto) and 88\,\% (from-scratch \ac{MLP}) of all forest as \textit{Other Broadleaves}, map \textit{Pinus} at only 0.5--1.7\,\% against the \ac{NFI}'s 30.7\,\%, and lose the conifers almost entirely (\cref{tab:national_map_1M_share}). Every fixed-feature method (the frozen AEF and TESSERA embeddings and the hand-crafted features) stays sound, and AEF and TESSERA perform consistently well compared with hand-crafted features.}

\new{This degeneration in SIBA is invisible in both the reference accuracy and map accuracy. The SIBA-trained deep models post the \emph{highest} map overall accuracy (92.2\,\% fine-tuned Presto, 97.9\,\% from-scratch \ac{MLP}), but these figures come from SIBA's spatially autocorrelated pixel-level split (\cref{sec:experiments}) and are the same leakage-inflated results. Against the independent \ac{NFI} basal-area distribution, the same SIBA maps by Presto and \ac{MLP} are the \emph{worst}.
% : their per-class area proportions deviate from the national \ac{NFI} distribution by 15.1--18.5\,\%, roughly two to three times the 5.2--7.6\,\% of the fixed-feature methods. 
The degeneracy scales with the overfitting -- the from-scratch \ac{MLP} produces the most degenerate map with SIBA dataset. The grid-cell deviation makes the same point directionally: for the sound maps it falls as cells are better sampled (AEF SIBA 11.0\,\%$\rightarrow$6.6\,\% from all cells to $\geq$20-plot cells), whereas for the degenerate SIBA maps it \emph{rises} (fine-tuned Presto 14.7\,\%$\rightarrow$19.3\,\%), because the degenerate SIBA maps miss the conifer classes.}

\begin{table}[htbp]
    \centering
    \small
    \caption{\new{1M-pixel-sample map overall accuracy (Olofsson et al., 2014) by method and training source. SIBA overall accuracy is computed on SIBA's leakage-prone pixel-level split (\cref{sec:experiments}) and is therefore optimistic for the encoder-adapting models; the honest arbiter is the grid-cell agreement with the \ac{NFI} (\cref{tab:national_map_1M_gridcell}).}}
    \label{tab:national_map_1M}
    \begin{newtable}
    \begin{threeparttable}
    \begin{tabular}{lrr}
    \toprule
    Method & \makecell{SIMB OA\\(\%, 95\% CI)} & \makecell{SIBA OA\\(\%, 95\% CI)} \\
    \midrule
    Presto-ft + RF            & $76.3 \pm 4.6$ & $92.2 \pm 1.5$ \\
    MLP-scratch               & $69.5 \pm 4.8$ & $97.9 \pm 1.0$ \\
    AEF                       & $65.9 \pm 5.2$ & $86.6 \pm 1.2$ \\
    TESSERA ($3\times3$ mean) & $73.9 \pm 4.9$ & $85.0 \pm 1.4$ \\
    Harmonic+Medoid           & $66.5 \pm 4.9$ & $84.1 \pm 1.3$ \\
    \bottomrule
    \end{tabular}
    \end{threeparttable}
    \end{newtable}
\end{table}

% \new{\cref{tab:national_map_1M_gridcell} gives the grid-cell deviation from the \ac{NFI} for all five methods on the 1M sample. On the SIMB source the deep maps from fine-tuned Presto and from-scratch \ac{MLP} are the best-agreeing at every threshold. On SIBA the deviation \emph{falls} with plot count for the three sound fixed-feature maps but \emph{rises} for the two degenerate encoder-adapting maps (fine-tuned Presto and from-scratch \ac{MLP}) -- a threshold-direction flip that is itself a diagnostic of the degeneration, independent of the leakage-inflated overall accuracy.}

\begin{table}[htbp]
    \centering
    \small
    \caption{\new{Grid-cell species-proportion deviation (\%) from the \ac{NFI} on the 1M-pixel sample, all five methods, by training source and \ac{NFI} plot-count threshold.}}
    \label{tab:national_map_1M_gridcell}
    \begin{newtable}
    \begin{threeparttable}
    \begin{tabular}{lrrrrrr}
    \toprule
     & \multicolumn{3}{c}{SIMB} & \multicolumn{3}{c}{SIBA} \\
    \cmidrule(lr){2-4}\cmidrule(lr){5-7}
    Method & All & $\geq$10 & $\geq$20 & All & $\geq$10 & $\geq$20 \\
    \midrule
    Presto-ft + RF            & 8.88 & 5.77 & 5.10 & 14.65 & 16.90 & 19.33 \\
    MLP-scratch               & 8.83 & 5.57 & 4.77 & 17.56 & 20.88 & 22.75 \\
    AEF                       & 9.38 & 6.74 & 6.48 & 11.01 &  7.92 &  6.61 \\
    TESS. ($3\times3$) & 9.07 & 6.31 & 6.34 &  9.92 &  7.38 &  5.77 \\
    Harm+Med           & 9.54 & 6.62 & 6.98 & 12.69 &  8.58 &  7.17 \\
    \bottomrule
    \end{tabular}
    \end{threeparttable}
    \end{newtable}
\end{table}

\begin{table}[htbp]
    \centering
    \small
    \caption{\new{National class-area share (\% of classified points) on the 1M-pixel sample, SIBA-trained models. The fine-tuned Presto and from-scratch \ac{MLP} maps collapse onto \textit{Other Broadleaves} and lose the conifers, while the three fixed-feature methods (frozen AEF/TESSERA embeddings and hand-crafted harmonic + medoid), trained on the same data, retain all classes.}}
    \label{tab:national_map_1M_share}
    \begin{newtable}
    \begin{threeparttable}
    \begin{tabular}{lrrrrr}
    \toprule
    Class & \makecell{Presto\\FT + RF} & \makecell{MLP\\scratch} & AEF & \makecell{TESS.\\$3\times3$} & \makecell{Harm\\+Med} \\
    \midrule
    OtherBroadleaves & 61.4 & 87.7 & 20.5 & 22.6 & 14.0 \\
    Quercus          & 38.0 &  5.3 & 27.4 & 31.9 & 31.2 \\
    Pinus            &  0.5 &  1.7 & 15.4 & 16.0 & 15.4 \\
    DarkConifer      &  0.0 &  0.1 & 10.1 & 10.5 & 11.2 \\
    Larix            &  0.0 &  1.3 &  6.8 &  8.4 & 11.4 \\
    Populus          &  0.0 &  3.9 & 16.8 &  8.0 & 13.4 \\
    Beech            &  0.0 &  0.0 &  2.9 &  2.7 &  3.4 \\
    \bottomrule
    \end{tabular}
    \end{threeparttable}
    \end{newtable}
\end{table}

\begin{figure*}[!ht]
    \centering
    \includegraphics[width=\linewidth]{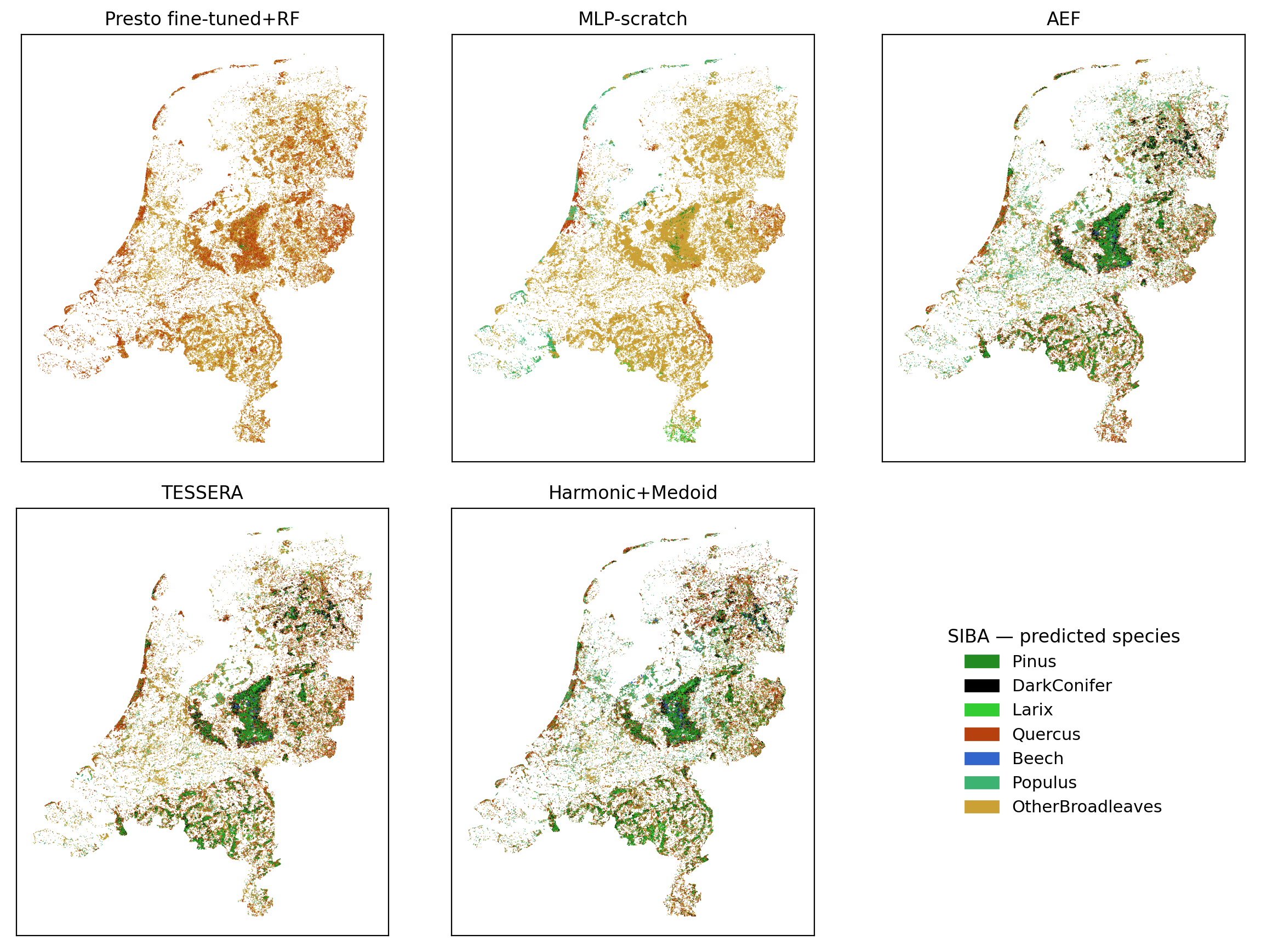}
    \caption{\new{Predicted species at 1,000,000 random forest pixels, SIBA-trained: fine-tuned Presto, from-scratch \ac{MLP}, AEF, TESSERA ($3\times3$ pool), and hand-crafted harmonic + medoid. The two encoder-adapting models (fine-tuned Presto, from-scratch \ac{MLP}) collapse -- the Veluwe Scots-pine belt, visible in every other panel, is erased -- while the three fixed-feature methods stay sound. Corresponding SIMB-trained maps in \cref{fig:national_map_1M_simb} (Appendix).}}
    \label{fig:national_map_1M_siba}
\end{figure*}

\subsubsection{\new{Full Dutch forest species mapping}}
\label{sec:full_dutch_map}
\new{Considering 1M-pixel mapping results, we extend our mapping with AEF and TESSERA ($3\times3$ mean-pooling) for the full wall-to-wall map. They (i) have better results than the hand-crafted method at the plot level and map level for both datasets, (ii) are the methods that remain computationally inexpensive at full 37.9 million pixels -- unlike Presto, from-scratch and hand-crafted, whose per-pixel time-series extraction would be more costly (\cref{sec:national_map_discussion}) -- \emph{and} (iii) produce sound maps for both datasets.
% \new{This is a complementary demonstration of practical, wall-to-wall deployability for pre-computed embeddings, not a claim that AEF or TESSERA is the single best-performing representation overall -- fine-tuned Presto remains the strongest point-level model on \ac{NFI} (\cref{tab:RQ1_combined}). }
% They are nonetheless not simply a lower-accuracy fallback: mean-pooled TESSERA substantially narrows its point-level gap with fine-tuned Presto on \ac{NFI} (\cref{tab:spatial_patch_headline}), and both AEF and TESSERA outperform every fine-tuned Presto variant under SIBA's spatially blocked evaluation (\cref{tab:siba_spatial}).
}

\begin{figure*}[!ht]
    \begin{subfigure}{.48\textwidth}
        \centering
        \includegraphics[width=\textwidth]{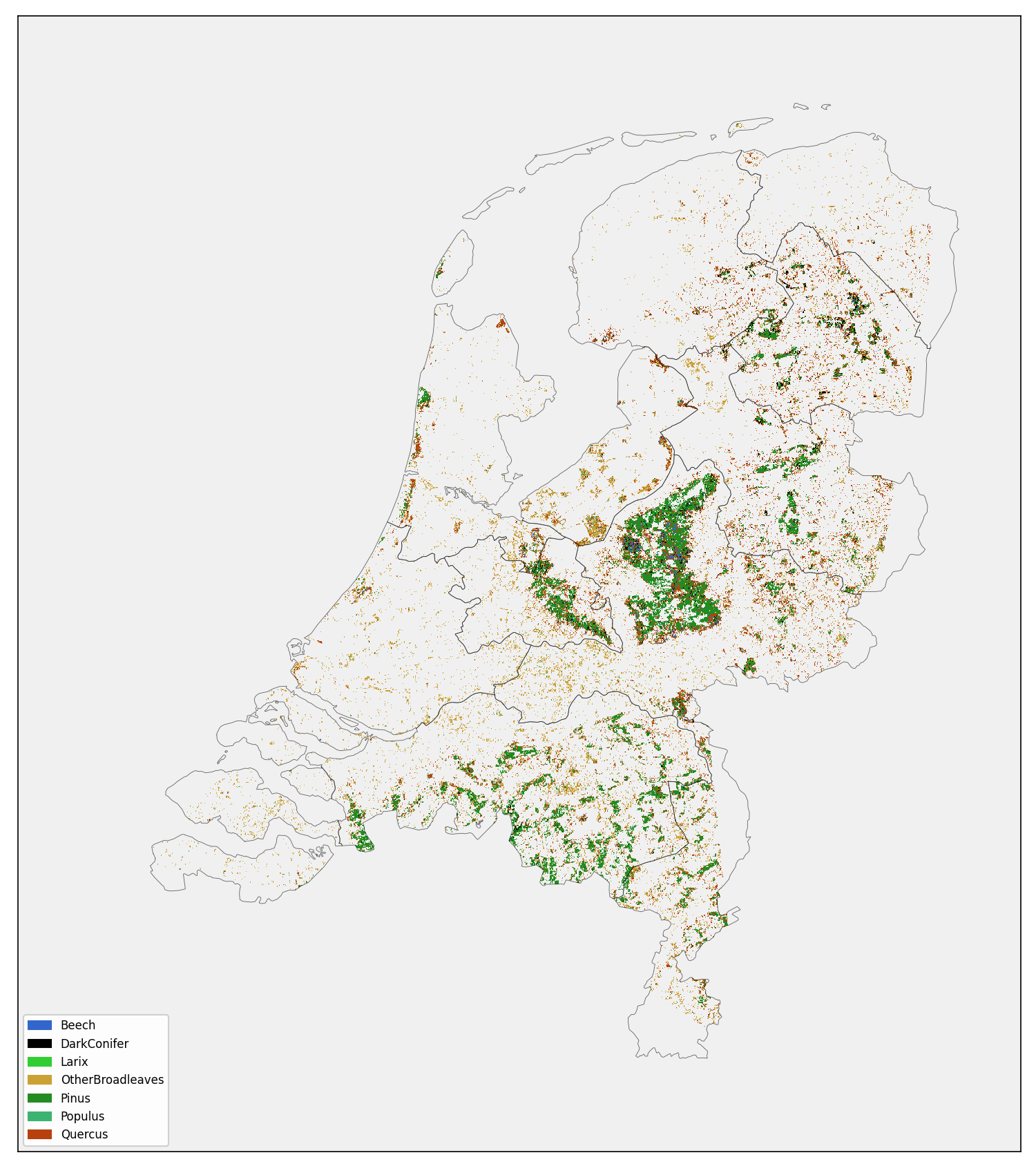}
        \caption{\new{TESSERA, trained on SIMB}}
        \label{fig:national_map:tessera_simb}
    \end{subfigure}
    \hfill
    \begin{subfigure}{.48\textwidth}
        \centering
        \includegraphics[width=\textwidth]{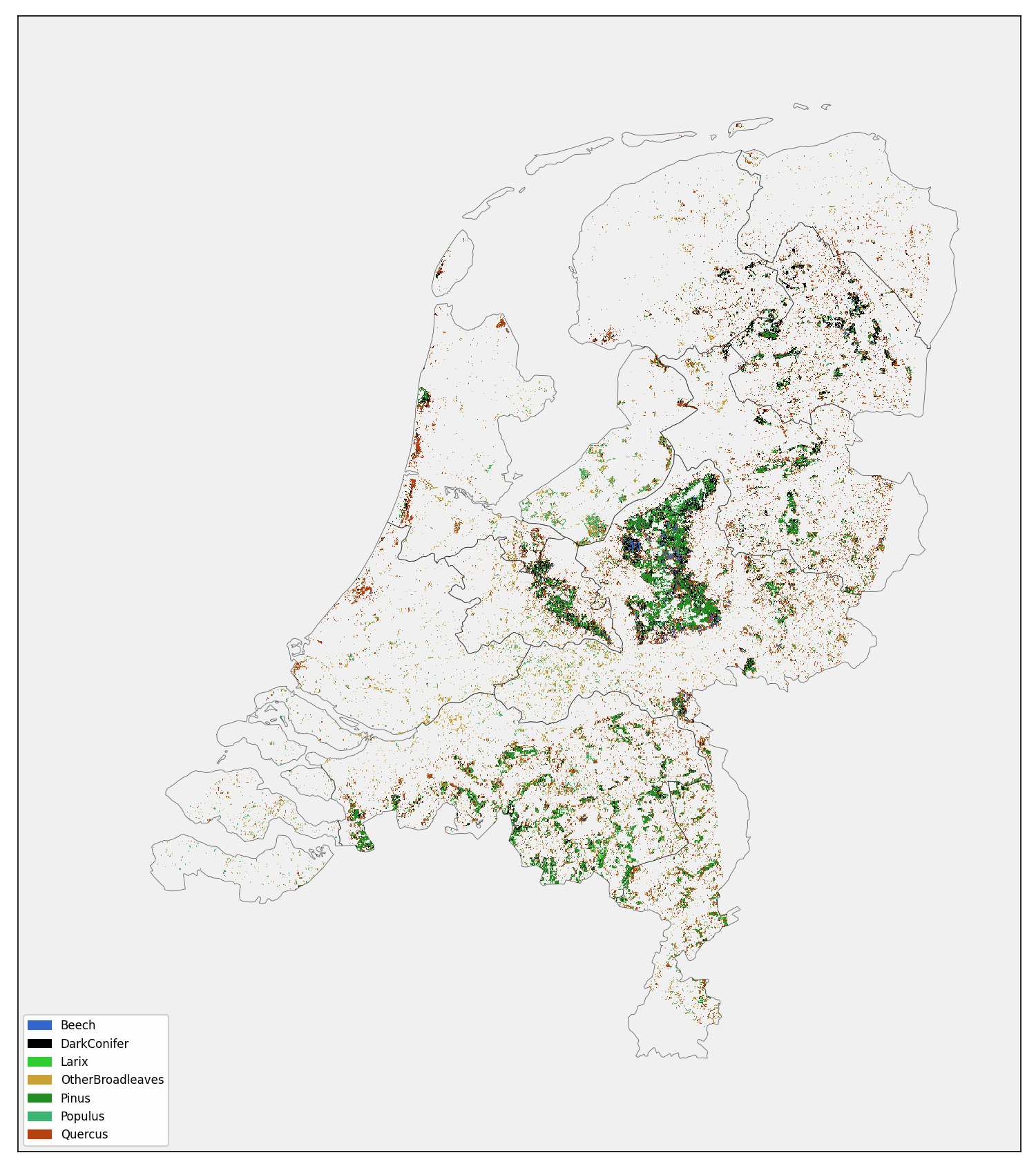}
        \caption{\new{TESSERA, trained on SIBA}}
        \label{fig:national_map:tessera_siba}
    \end{subfigure}
    \caption{\new{National-scale species classification (10\,m resolution, forest pixels only), TESSERA embeddings trained on SIMB (left, 74.69\% $\pm$ 4.47\% overall accuracy) and SIBA (right, 84.68\% $\pm$ 1.36\%). See \cref{app:national_map_annex} for the corresponding AEF maps.}}
    \label{fig:national_map_tessera}
\end{figure*}

\new{\cref{fig:national_map_tessera} shows the resulting national species maps for TESSERA, trained on SIMB (left) and SIBA (right). \cref{tab:national_map_headline} summarises overall map accuracy for all four embedding/training-source combinations.}
% \new{The two SIBA-trained maps achieve markedly higher, near-identical overall accuracy (86.15\% AEF, 84.68\% TESSERA).}
\new{Comparing the classified maps pixel counts against the \ac{NFI} basal-area proportion using the $10\times10$\,km grid-cell method (\cref{sec:national_map_methods}) gives a second, independent check (\cref{tab:national_map_gridcell}).}
\new{These maps and statistics indicate that the mappings are sound, and the results are similar to the 1M-pixel analysis.}
% \new{Average per-cell species-proportion deviation is 8.5--8.7\% for the two SIMB-trained maps and 10.4--11.0\% for the two SIBA-trained maps, falling to 6.0--6.8\% when restricted to cells with $\geq$20 \ac{NFI} plots, closely matching \citeauthor{francini_DutchForestSpeciesMap_2024}'s own reported 9\% (7.5\%~/~6\% at the same plot-count thresholds).} 

\begin{table}[htbp]
    \centering
    \small
    \caption{\new{National-scale (full 37.9M-pixel) map overall accuracy (Olofsson et al., 2014) by embedding and training source.}}
    \label{tab:national_map_headline}
    \begin{newtable}
    \begin{threeparttable}
    \begin{tabular}{lrr}
    \toprule
    Embedding & \makecell{SIMB OA\\(\%, 95\% CI)} & \makecell{SIBA OA\\(\%, 95\% CI)} \\
    \midrule
    AEF                       & $67.03 \pm 4.74$ & $86.15 \pm 1.24$ \\
    TESSERA ($3\times3$ mean) & $74.69 \pm 4.47$ & $84.68 \pm 1.36$ \\
    \bottomrule
    \end{tabular}
    \end{threeparttable}
    \end{newtable}
\end{table}

\begin{table}[htbp]
    \centering
    \small
    \caption{\new{Mean per-cell species-proportion deviation (\%) from \ac{NFI} basal-area data on the full 37.9M-pixel wall-to-wall maps, by embedding and training source, at three \ac{NFI} plot-count thresholds ($n=325$ / $100$ / $39$ cells for all / $\geq$10 / $\geq$20 plots).}}
    \label{tab:national_map_gridcell}
    \begin{newtable}
    \begin{threeparttable}
    \begin{tabular}{lrrrrrr}
    \toprule
     & \multicolumn{3}{c}{SIMB} & \multicolumn{3}{c}{SIBA} \\
    \cmidrule(lr){2-4}\cmidrule(lr){5-7}
    Embedding & All & $\geq$10 & $\geq$20 & All & $\geq$10 & $\geq$20 \\
    \midrule
    AEF                       & 8.70 & 6.10 & 6.03 & 11.00 & 8.12 & 6.78 \\
    \makecell{TESSERA\\($3\times3$ mean)} & 8.50 & 5.82 & 6.07 & 10.43 & 7.95 & 6.05 \\
    \bottomrule
    \end{tabular}
    \end{threeparttable}
    \end{newtable}
\end{table}

\new{\cref{tab:national_map_cm_tessera_siba} gives the reference confusion matrix for the SIBA-trained TESSERA map. This also indicates similar confusion in the pixel level classification (\cref{sec:class_level_performance}).}

\begin{table*}[htbp]
    \centering
    \small
    \caption{\new{TESSERA $\times$ SIBA reference confusion matrix ($n=4135$).}}
    \label{tab:national_map_cm_tessera_siba}
    \begin{newtable}
    \begin{threeparttable}
    \begin{tabular}{lrrrrrrrrr}
    \toprule
     & \multicolumn{7}{c}{Reference (true) class} & \\
    \cmidrule(lr){2-8}
    Map (predicted) class & Be & DC & La & OB & Pi & Po & Qu & User's acc. (\%) \\
    \midrule
    Beech & 544 & 16 & 4 & 16 & 0 & 1 & 30 & $89.0 \pm 2.5$ \\
    DarkConifer & 11 & 489 & 25 & 8 & 31 & 1 & 11 & $84.9 \pm 2.9$ \\
    Larix & 4 & 23 & 540 & 13 & 20 & 0 & 10 & $88.5 \pm 2.5$ \\
    OtherBroadleaves & 8 & 1 & 2 & 462 & 4 & 8 & 37 & $88.5 \pm 2.7$ \\
    Pinus & 5 & 37 & 14 & 10 & 515 & 0 & 22 & $85.4 \pm 2.8$ \\
    Populus & 4 & 1 & 1 & 25 & 0 & 576 & 12 & $93.1 \pm 2.0$ \\
    Quercus & 15 & 24 & 5 & 57 & 19 & 5 & 469 & $79.0 \pm 3.3$ \\
    \bottomrule
    \multicolumn{9}{r}{Overall accuracy: $84.68\% \pm 1.36\%$}
    \end{tabular}
    \begin{tablenotes}
    \footnotesize
    \item[] Rows: map (predicted) class; columns: reference (true) class. Be = Beech, DC = DarkConifer, La = Larix, OB = OtherBroadleaves, Pi = Pinus, Po = Populus, Qu = Quercus. $\pm$ indicates 95\% confidence interval of user's accuracy.
    \end{tablenotes}
    \end{threeparttable}
    \end{newtable}
\end{table*}

\begin{figure*}[!ht]
    \centering
    \includegraphics[width=0.9\linewidth]{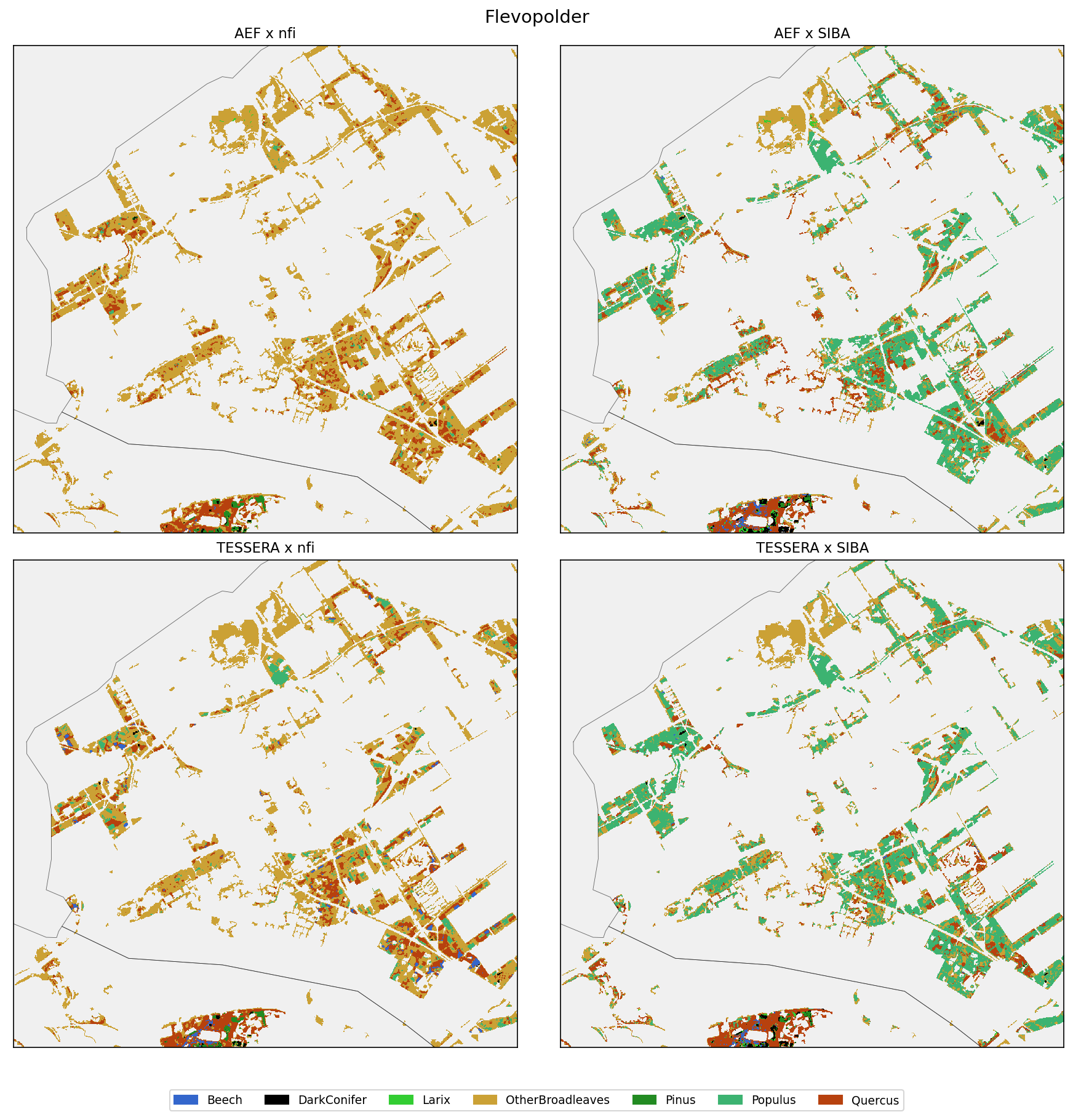}
    \caption{\new{Regional comparison, Flevopolder: AEF $\times$ SIMB, AEF $\times$ SIBA, TESSERA $\times$ SIMB, TESSERA $\times$ SIBA.}}
    \label{fig:national_map_regions_flevo}
\end{figure*}

\new{The clearest concrete illustration of the two training sources' contrasting rare-class behaviour is the Flevopolder region, which has largely poplar plantations \citep{francini_DutchForestSpeciesMap_2024}. \cref{fig:national_map_regions_flevo} shows all four classified maps for this region side by side: the two SIMB-trained maps classify the area as predominantly \textit{Other Broadleaves}, almost entirely missing \textit{Populus}, while the two SIBA-trained maps correctly identify it as \textit{Populus}-dominant.}

\new{The corresponding AEF full national maps, two further regional comparison maps \citep{francini_DutchForestSpeciesMap_2024}, remaining confusion matrices and full per-class accuracy/area tables are provided in \cref{app:national_map_annex}.}
% the three  (AEF $\times$ SIMB, AEF $\times$ SIBA, TESSERA $\times$ SIMB) are given in \cref{app:national_map_annex}
% \new{The qualitative regional comparison against \citeauthor{francini_DutchForestSpeciesMap_2024}'s own published map is provided in \cref{app:national_map_annex}.}

%%%%%%%%%%%%%%%%%%%%%%%%%
\section{Discussion}
\label{discussion}

\subsection{Multi-modal sensor integration for harmonic + medoid features}
Our findings on the modest but consistent benefits of combining S-1 radar with S-2 optical data for harmonic + medoid features align with recent European tree species mapping studies. \citet{blickensdorfer_GermanTreeSpeciesMap_2024} demonstrated that S-1 SAR data effectively fills potential data gaps in cloud-affected regions, particularly important for national-scale applications where clear-sky observation density varies substantially. Similarly, \citet{liu_MappingTreeSpecies_2023} confirmed that S-1 provides supplementary structural information to S-2 optical data, contributing to increased reliability of large-scale tree species mapping. This benefit could be attributed to the complementary nature of the sensors: S-1 SAR captures structural information, while S-2 optical data provides insights into canopy density and foliage characteristics \citep{forkuor_AbovegroundBiomassMapping_2020}. 

Conversely, the addition of S-1 sensors for tree species classification does not always yield substantial accuracy improvements. \citet{Lechner_Combination-S1S2-TreeSpecies_2022} noted that the advantages of SAR are most pronounced when S-2 lacks sufficient cloud-free observations. Given the monthly and seasonal composite nature of our harmonic and medoid features, most S-2 composites in our study contained valid data. This high data availability likely resulted in the marginal improvements observed. 

Our results, showing 0.4-1.4\% accuracy improvements when adding S-1 to S-2 features\new{in addition to phenological and per-band attribution analysis (\cref{sec:interpretability_results})}, corroborate these findings and provide confidence in the robustness of multi-modal approaches for operational NFI applications.

\subsection{Deep vs hand-crafted features: an emerging paradigm}
\label{sec:deepvshandcrafted}
The transition from hand-crafted to learned feature representations marks a significant shift in remote sensing methodology. Even though the interpretability of deep features remains an active research topic \citep{Rao_MeasuringIntrinsicDimension_2025}, learned representations capture complex, distributed temporal and multisensor patterns more effectively than simple harmonics and summary statistics \citep{Bengio_RepresentationLearningReview_2014}. In particular, transformer-based and embedding-field models trained with self-supervision compress irregular, noisy multi-sensor time series into high-dimensional representations that are useful for downstream tasks \citep{tseng_Presto_2023, feng2025tessera, brown2025alphaearth}. Conceptually, the encoder in deep embedding models learns data-adaptive basis components and nonlinear combinations, whereas harmonic coefficients are fixed basis functions chosen a priori and may not adequately capture the subtle differences among very similar classes. Furthermore, self-supervised pre-training on large unlabeled data enhances embedding quality and improves label efficiency and generalization \citep{wang_Self-Supervised-Learning-RS_2022}. These representational advantages likely explain the consistent gains in classification performance we observed across datasets of differing size and class balance\new{, as well as mapping and grid-cell species proportion deviation performance}.

While deep learning has proven transformative in crop type mapping and land cover classification \citep{jakubik_Prithvi_2023, szwarcman_prithvi-eo-20_2024,wang_CropformerNewGeneralized_2023}, its application to tree species classification at NFI scale represents a relatively recent development. Our results demonstrate consistent improvements in evaluation metrics \newreplace{of 2–9 percentage points}{and mapping} across datasets of varying class balance and complexity, suggesting that transformer-derived features capture phenological and structural variability more effectively than engineered metrics by \citet{francini_DutchForestSpeciesMap_2024}. This performance gain supports the ongoing shift in remote sensing from task-specific feature engineering toward transferable, learned representations that can adapt to diverse ecological contexts without domain-specific tuning.

Beyond accuracy, these results carry broader scientific implications. Consistent with \citet{ballGeospatialFoundationModels2026}, who demonstrate this for temperate mountain forests, our findings suggest that geospatial foundation models shift the primary bottleneck in tree species mapping from \emph{feature engineering} toward the \emph{availability, quality, and temporal alignment of ecological reference data}. This has a direct implication for how future NFI monitoring effort should be allocated: rather than investing in designing country-specific spectral-temporal features, the scientific priority should shift toward securing high-quality, spatially representative labeled plots. Furthermore, the use of globally pre-trained encoders \newreplace{fine-tuned on}{or embeddings with} each country's local NFI data suggests a scalable pathway for national forest monitoring: individual countries can adapt shared pre-trained models\new{ or embeddings} to their own labeled inventory, reducing the methodological barrier for nations with limited remote sensing expertise.

\subsection{Frozen \newreplace{vs}{and} fine-tune embeddings}
We include three deep embedding representations: AEF embeddings, TESSERA embeddings, and Presto embeddings (both frozen and fine‑tuned). These representations differ substantially in model scale, training data, and training methodology, which influences their performance in downstream species classification tasks.

Across all datasets and evaluation metrics with the RF classifier\new{ at the reference-pixel level}, we find that embeddings extracted from the fine‑tuned Presto consistently improve downstream classification performance compared to using frozen Presto, AEF, or TESSERA embeddings with RF. \new{This benefit can be observed in 1M pixel mapping for the spatially well-separated SIMB dataset (\cref{sec:1M_pixel_mapping}), where the fine‑tuned Presto embeddings yield higher map accuracy and lower deviation than other methods, and a plausible species map.}
Although AEF embeddings are trained with encoders ranging from 480M to 1B parameters on approximately 3B image observations, and TESSERA embeddings are trained with a 40M-parameter encoder on 0.8B time‑series pixels, both optimized to capture complex spatio‑temporal patterns across diverse land covers, these embeddings are produced from frozen (fixed) encoders. AEF’s model and weights are not publicly available. Therefore, its encoder cannot be fine-tuned for the specific forest species classification task evaluated here. Furthermore, while TESSERA is open and powerful, its large number of parameters requires substantial computational resources for fine-tuning and embedding extraction, which would likely overfit the small-scale training dataset.
By contrast, Presto is trained with a 0.4M parameter encoder on 21.5M time‑series pixels\newreplace{, and its light-weighted pretrained model is openly available for fine‑tuning. Frozen Presto embeddings are extracted from the pretrained encoder without updating any model weights during fine‑tuning. As a relatively small model, the performance of frozen Presto embeddings is similar to or worse than that of harmonic + medoid features. However, }{. }when the same pretrained Presto encoder is further fine‑tuned on the downstream forest species classification objective, the resulting embeddings exhibit improved discriminative power compared to the other \new{frozen }embeddings.

\new{One important thing we need to pay attention to is that this task-specific adaptation carries a risk that the frozen embeddings do not. Because fine-tuning (and training from scratch) updates the encoder on the downstream labels, it can also learn signal specific to how those labels are spatially arranged rather than to the species themselves. On SIBA, whose photo-interpreted pixels are spatially autocorrelated within stand polygons, this manifests at map scale. The fine-tuned Presto and from-scratch models achieve the highest reference accuracy (92--98\,\%) yet produce degenerate national maps, where more than half of all forest collapsed into a single class, the conifers were lost, and grid-cell species proportion deviation increased. While the frozen AEF and TESSERA embeddings, which cannot overfit \emph{through} the encoder, remain sound and yield better statistics on the 1M-pixel map than hand-crafted features in both SIMB and SIBA. Whether fine-tuning helps or hurts is therefore dataset and training setting dependent.}
% \new{On the well-separated \ac{NFI} plots it clearly helps (\cref{sec:deepvshandcrafted}), but on spatially leaky reference data a high test-set accuracy is not a reliable indicator of map quality.}

This result highlights the advantage of \newreplace{task-specific adaptation, even though the fine-tuning process necessitates additional steps and computational effort}{deep embeddings including task-specific adaptation under careful consideration for training datasets and setting}. \newreplace{While f}{F}oundation embeddings like AEF and TESSERA offer rich, general representations that serve as a solid baseline for \newreplace{many tasks}{tree species classification and its mapping}, since they are pre-computed and easily extracted from coordinates without requiring raw time-series data processing\newreplace{, fine-tuned embeddings tailored to the specific classification task capture discriminative information more effectively}.

\subsection{Neural network architectures for learned features}

    \newreplace{The generally modest yet consistent performance differences observed between \ac{MLP} and \ac{RF} classifiers when trained on deep embeddings may also reflect differences in the training protocol and hyperparameter choices, not only intrinsic classifier inductive biases.}{The dataset-dependent differences observed between \ac{MLP} and \ac{RF} classifiers when trained on deep embeddings, with RF matching or slightly exceeding MLP on the small NFI datasets and MLP leading only on the large SIBA dataset, may also reflect differences in the training protocol, not only intrinsic classifier inductive biases.} In our experiments the RF models were trained using 70\% of the labeled data, whereas the fine-tuning process for Presto (which includes MLP training) used a smaller effective training fraction because we split that 70\% into 80\%/20\% train/validation folds for early stopping (i.e., 70\% \(\times\) 80\% = 56\% used for optimizer updates, with the remaining 14\% used to select the lowest-validation-loss checkpoint). Using an internal validation split and early stopping is a common strategy to prevent overfitting in neural models, but it reduces the amount of data used for weight updates and can therefore affect measured performance, especially on small datasets \citep{Mahsereci_EarlyStoppingWithoutValidation_2017}. 

    \deleted{We also adopted hyperparameter settings from \cite{Mouret_TreeSpeciesClassificationPixelTimeSeriesInbalanced_2024}. while those settings are reasonable starting points, hyperparameter sensitivity is known to substantially affect deep learning performance and optimal settings can depend on dataset size, class aggregation, and geographical region. Systematic hyperparameter search, nested cross-validation, or adaptive optimization (e.g., Bayesian) can therefore be important to conclusively ascribe performance differences to classifier architecture rather than to training procedure \citep{Raiaan_SystematicReviewHyperparameter_2024}.}

    \newreplace{Taken together, the observed small gains of MLP over RF are likely the result of both (i) the MLP's ability to exploit non-linear structure in high-dimensional embeddings and (ii) procedural factors (reduced effective training set for MLP due to the validation split and reliance on transferred hyperparameters).}{Taken together, the dataset-dependent ordering of MLP and RF is likely the result of both (i) the MLP's ability to exploit non-linear structure in high-dimensional embeddings, which becomes advantageous mainly when training data are abundant (as on SIBA), and (ii) procedural factors, in particular the smaller effective training set used by the MLP (56\% versus 70\% for RF, owing to the internal validation split), which penalises the MLP most on the small NFI datasets, together with reliance on transferred hyperparameters.}
    \deleted{To strengthen these conclusions, we recommend follow-up experiments such as increasing the effective training fraction (e.g., by using cross-validation or a larger outer training split), running a targeted hyperparameter search for the MLP on each dataset, and reporting stability across different validation splits and seeds.}

\subsection{Computational efficiency and operational scalability}

From an operational perspective, \newreplace{this approach}{the pre-computed embeddings such as AEF and TESSERA} offers \newreplace{significant}{practical} advantages for national-scale \new{classification} implementations. 
\deleted{The computational efficiency is evidenced by \ac{Presto}'s lightweight architecture, which enables fine-tuning on a single GPU or CPU \citep{tseng_Presto_2023}, making it accessible to organizations with limited computational resources. The pre-training process itself demonstrates remarkable efficiency, processing 21.5 million pixel time-series in only 2 hours 12 minutes per epoch (43 hours 15 minutes for 20 epochs), indicating that similar approaches could be readily applied to other countries' NFI systems. For The Netherlands specifically, with its 36.4 million 10$\times$10\,m pixels, large-scale pixel-label mapping appears computationally feasible using this methodology.}
\new{We produced a national map for both embeddings and both training sources (\cref{sec:national_map_methods,sec:national_map_results,sec:national_map_discussion}), including map-level accuracy and area-proportion estimation \citep{Olofsson_GoodPracticesAreaEstimation_2014} and a comparison against the National Forest Inventory at $10\times10$\,km resolution. Wall-clock cost differs by embedding: AEF's server-side classification completed in 1-2~hours (backed by up to 486,000 EECU-seconds of parallelised cloud compute in \ac{GEE}), while TESSERA's fully local classification took approximately 3--4~hours per map after a one-time $\sim$10-hour, 71\,GB local tile-cache download. In contrast, for Presto, from-scratch MLP and harmonic + medoid, a national map would require extracting a full 12-month time series for all 37.9 million forest pixels, which requires substantial computational time.}
\deleted{While incorporating spatial and temporal context is known to benefit tree species classification \citep{Mu_NationalscaleTreeSpecies_2025}, training such patch-based architectures typically requires substantial computational resources. By contrast, \citet{tseng_Presto_2023} explicitly designed Presto as a pixel-based model with approximately 1,000 times fewer parameters than comparable ViT- or ResNet-based models, enabling pre-training and fine-tuning on a single consumer-grade CPU or GPU and national-scale inference without high-performance computing infrastructure.}

\subsection{\new{National-scale mapping: validation, spatial patterns}}
\label{sec:national_map_discussion}

\new{We produced national-scale species maps with two pre-computed embeddings and two datasets (SIMB, SIBA). The grid-cell comparison between classified species counts and NFI average basal area proportion (\cref{sec:national_map_methods}) indicates similar deviations produced by  \citeauthor{francini_DutchForestSpeciesMap_2024}'s own published deviation figures (9\%/7.5\%/6\%) closely for both SIMB-trained maps. This shows practical pathway to use embeddings for national tree species mapping, with the caveat that the training data source can substantially affect rare-class representation. The SIBA-trained maps achieve higher overall accuracy and better rare-class representation than SIMB-trained maps, which is consistent with the point-level results (\cref{sec:experiments}).}

\new{The different mapping patterns are observed between training data sources rather than the embeddings. SIMB-trained maps systematically over-predict \textit{Other Broadleaves} while the SIBA-trained maps identify \textit{Populus} better in the Flevopolder region.}
\new{This is not a limit of the available spectral-temporal signal: per-species mean monthly trajectories (\cref{fig:phenology_trajectory}) already show Populus separating clearly from \textit{Other Broadleaves} and every other species in Sentinel-1 VH backscatter between April and September, and the full-band breakdown (\cref{fig:phenology_annex}) shows the same separation in VV. The bottleneck could be training-sample size: NFI/SIMB's \textit{Populus} class has only 50 training samples (\cref{tab:class_distribution_splits}), which is too few for the classifier to learn to exploit an otherwise usable signal. This strengthens the case for class-balanced, sufficiently large training data such as SIBA's design: the discriminating signal for a spectrally similar rare class is present in the input data, but is only learned when enough labelled examples exist to expose it. This augmented SIBA dataset demonstrably improves rare-class representation at the country scale.}
\new{One additional implication is derived from the close agreement between the 1M-pixel sample and the full census (\cref{app:national_map_1M_vs_census}): a cheap 1M-point map can screen whether a full wall-to-wall map will be sound before committing to the far more expensive full-country extraction.}

\subsection{\new{Spatial context integration to pixel-based embeddings}}
\label{sec:spatial_patch_discussion}

\new{\cref{sec:spatial_patch_results} (\cref{tab:spatial_patch_headline}) shows that spatial context can straightforwardly be added to a pixel-based approach via $n\times n$ patches, but the benefit is not uniform -- it depends systematically on how much spatial context a representation already captures internally, and, for a fine-tuned representation, on whether fine-tuning itself already had access to that context. AEF's Space-Time Precision encoder explicitly combines spatial and temporal attention (\cref{sec:methods:deepfeatures}), so a given pixel's embedding already reflects information from its surroundings; concatenating or pooling the literal neighbour pixels therefore adds essentially nothing on \ac{NFI}, and only a modest amount on SIBA. TESSERA, by contrast, is an explicitly \emph{pixel-wise} encoder with no spatial context integration step \citep{feng2025tessera}, and complementary spatial signal in the neighbourhood pixel enhance the performance (\cref{sec:spatial_patch_results}).}
%  \new{This is consistent with, and extends, our existing architectural comparison of these three models (\cref{sec:deepvshandcrafted,sec:methods:deepfeatures}): the same design choices that determine how each model is pre-trained also determine how much it stands to gain from a spatial-pooling step.}

\new{Fine-tuned Presto, also a pixel-wise encoder \citep{tseng_Presto_2023}, shows benefits in NFI (SIMB, COMB) but not in SIBA. 
% no pooling variant beats the single-pixel baseline on either split (\cref{sec:spatial_patch_results}). 
Center-pixel fine-tuned Presto is affected by spatial correlation in SIBA's pixel-level split and overfitting, so its accuracy is likely already saturated, leaving little headroom for pooling. 
% The difference is that, unlike AEF and TESSERA (fixed embeddings, identical regardless of downstream task or split), fine-tuned Presto is retrained separately for SIBA on the exact same closely spaced plots (\cref{sec:experiments}); if adjacent, same-stand pixels are frequently split across its own train and validation sets during fine-tuning, the encoder may already absorb much of that local spatial context during training itself, leaving an explicit $3\times3$ neighbourhood with little further information to add. \ac{NFI}'s well-separated plots ($\geq$160.8\,m apart) offer well-sparated fine-tuning, so its neighbouring pixels remain a genuine source of new spatial context there. 
% Under SIBA's plot-grouped split specifically, this leaves AEF and TESSERA's pooled representations as the strongest embeddings overall, exceeding every fine-tuned Presto variant (Appendix Table \ref{tab:spatial_patch_aef_tess_full}) -- consistent with, and extended by spatial pooling to a wider margin than, \cref{tab:siba_spatial}'s finding that foundation-model embeddings overtake fine-tuned Presto once SIBA is evaluated honestly.
}

% \new{Spatial-context integration sometimes outperforms the pixel-wise baseline. On \ac{NFI}, TESSERA with $3\times3$ mean-pooling (78.88\% overall accuracy, SIMB; 71.78\%, COMB) exceeds a baseline fine-tuned-Presto-\ac{MLP} result at the single pixel (78.05\%, 69.35\%) even though mean-pooling Fine-tuned Presto reclaims and extends the lead (81.57\%~\ac{RF} / 81.03\%~\ac{MLP}, SIMB; 71.56\%~\ac{RF} / 72.10\%~\ac{MLP}, COMB).}

% \new{This evaluation uses the paper's standard random 70/30 split rather than a spatially blocked one; \citet{Corley_PoolingEarthEmbeddings_2026}, whose stats-pooling method we adopt as a comparison arm, find that plain mean-pooling is in fact the \emph{weakest} pooling strategy under spatial distribution shift, the opposite ranking from what we observe here under \ac{NFI}'s random split (\cref{tab:spatial_patch_aef_tess_full}). This is not a contradiction -- \ac{NFI}'s systematic sampling grid is the reason we already treat random splits as a reasonable proxy for spatial independence for this dataset (\cref{sec:experiments}) -- but it does mean the relative ranking of pooling strategies reported here should not be assumed to transfer to a genuinely spatially held-out evaluation. SIBA's plot-grouped split is one step in that direction, and there flatten/stats-pool already edge out mean-pool for AEF/TESSERA (\cref{sec:spatial_patch_results}), a small early signal in the direction \citeauthor{Corley_PoolingEarthEmbeddings_2026} report; a fully spatially held-out \ac{NFI} evaluation remains for future work.}

\subsection{Data quality, class balance, and spatial autocorrelation considerations}

{
    The results in \cref{tab:RQ1_combined}\new{ and \cref{sec:national_map_results}} indicate the role of class balance and sampling design (e.g. number of samples, spatial separation) in determining classification \new{and map }performance. Large \new{accuracy }performance gaps are observed between the original Dutch NFI datasets (SIMB and COMB) and the augmented SIBA dataset, despite using identical feature representations and classifiers.

    \deleted{The class-balanced SIBA dataset achieves substantially higher performance (93.54\,\% F1 macro with the MLP classifier) compared to the imbalanced SIMB and COMB datasets (62.29\,\% and 50.76\,\% F1 macro, respectively).}
    % \delete{The class-balanced SIBA dataset achieves substantially higher performance (\newreplace{93.54\,\%}{97.26\,\%} F1 macro with the MLP classifier) compared to the imbalanced SIMB and COMB datasets (\newreplace{62.29\,\% and 50.76\,\%}{67.94\,\% and 48.59\,\%} F1 macro, respectively).} 
    \newreplace{While SIBA also benefits from a larger sample size (13,790 samples versus approximately 1,460 for SIMB and COMB), the magnitude of improvement indicates that c}{C}lass balance is \newreplace{a primary }{one of the }driver\new{s} of model performance rather than data volume alone. This is particularly evident for the macro-averaged F1 score, which is sensitive to underrepresented classes and more representative of operational performance across species\new{, and in the SIMB map of the Flevopolder region}. 
    %  \new{SIBA's photo-interpreted pixels are only separated by a minimum of 15\,m to each other, and this distance is less than \citet{hermosilla_CanadaTreeSpeciesMapping_2022}'s recommended 45\,m minimum distance. As a result, overfitted fine-tuning Presto and from-scratch MLP produces degerated maps with inflated accuracy estimates.}
    These findings suggest that future NFI data collection strategies could benefit from explicitly targeting underrepresented tree species to improve class balance. The data augmentation strategy proposed by \citet{francini_DutchForestSpeciesMap_2024}, which supplements field-based NFI observations with visually interpreted samples from high-resolution satellite imagery, provides a practical pathway to achieve this goal. 

    \new{On the other hand, spatial autocorrelation of the data is also important to consider.}
    From a spatial perspective, both the NFI and SIBA datasets incorporate measures to reduce spatial autocorrelation. While the original Dutch NFI sampling design selects only one plot per square kilometer, the SIBA dataset is augmented by delineating polygons around these NFI plots. Although SIBA enforces a minimum inter-sample distance of 15\,m \citep{francini_DutchForestSpeciesMap_2024}, this may not be sufficient to fully eliminate spatial autocorrelation in heterogeneous forest landscapes\newreplace{. In such environments, species composition and canopy structure can remain correlated over larger distances;}{, and} for instance, \citet{hermosilla_CanadaTreeSpeciesMapping_2022} recommended a minimum sampling distance of 45\,m.
    In spatial machine learning, failure to account for spatial dependence between training and testing samples can lead to overly optimistic performance estimates \citep{Karasiak_SpatialDependenceTrain-Test_2022}. Consequently, \newreplace{the use of a spatially independent validation dataset would be highly beneficial for ensuring a robust evaluation of NFI-scale applications.}{a spatially independent evaluation is needed for a robust assessment of NFI-scale applications \citep{Ploton_SpatialValidation_2020}.}
    \new{To quantify this autocorrelation effect for SIBA datasets, we tested a plot-grouped (spatially blocked) split that assigns all pixels augmented through photo-interpretation from a single plot exclusively to one partition (train, test, or validation) (\cref{tab:siba_spatial}). This grouped split creates an extreme scenario where certain areas lack training data entirely, while others have a high density of pixels. Then, its impact is substantial: overall accuracy for every method drops from the near-ceiling pixel-level values (up to 94--98\%) to 45--62\%, and the ranking changes qualitatively. Under the pixel-level split, models that operate on the raw time series and can memorise a polygon's spectral signature dominate (a from-scratch MLP attains 98.3\% overall accuracy, exceeding fine-tuned Presto), whereas under the spatially blocked split these from-scratch baselines become the weakest models and the frozen, pre-computed embeddings (AEF and TESSERA) rank highest (62.0\% and 60.8\% overall accuracy). This indicates that a portion of the apparent SIBA performance reflects within-polygon spatial autocorrelation.} 
    
    \new{The original NFI datasets (SIMB and COMB), whose plots are separated by the inventory grid and are not pixel-augmented, are unaffected by spatial autocorrelation. On these datasets, fine-tuned models including Presto embeddings achieve higher accuracy scores, demonstrating that the advantage of fine-tuned embeddings remains robust. In addition, pre-computed embeddings (AEF and TESSERA) still outperform hand-crafted harmonic + medoid features in spatially blocked split in SIBA, confirming that learned representations show a robust performance even in such a scenario.}
}

    \subsection{Ecological Interpretation of Classification Patterns}
    \label{sec:ecological_interpretation}
    Beyond quantitative performance metrics, the observed classification patterns have meaningful ecological implications that reflect the underlying biophysical and phenological characteristics of the plot types. 

    While deep embeddings capture key phenological signals, they have intrinsic limits for discriminating species that share very similar seasonal trajectories: when two taxa leaf-out, peak and senesce at similar times and have comparable canopy optical properties, spectral-temporal separability is inherently low \citep{Persson_TreeSpeciesClassification-multi-temporalS2_2018}. As detailed in \cref{sec:class_level_performance}, this is particularly evident for broadleaved species such as \textit{Alnus}, \textit{Betula}, and \textit{Fraxinus}, which exhibit the highest inter-class confusion consistent with their overlapping deciduous phenologies and similar canopy reflectance in the Sentinel spectral range. \newreplace{Critically, this}{This} persistent confusion is not specific to deep embeddings but reflects a fundamental separability ceiling that any spectral-temporal classifier would face. \citet{tan_everagingSentinel1_2025} independently corroborate this, finding that deciduous species cluster closely in feature space, confirming reduced separability among phenologically similar taxa. More fundamentally, at the 10m spatial resolution of Sentinel data, mixed-species pixels impose a classification ceiling that affects pixel-based and patch-based spatial methods alike \citep{Mu_NationalscaleTreeSpecies_2025}. 
    
    This limitation can be mitigated by integrating complementary sensors that encode different trait axes such as canopy height, vertical structure from airborne LiDAR or GEDI, and spectral information \citep{lang_high-resolution_2023, wagner_highresolutiontreeheight_2025} or higher resolution images \citep{Mu_NationalscaleTreeSpecies_2025}.
    \newreplace{Second, the superior performance of deep embeddings in our experiments is consistent with recent findings in the time-series remote sensing literature: self-supervised, pre-trained encoders learn high-dimensional phenological and contextual representations that generalize better across classes and recording conditions than a fixed set of harmonic + medoid statistics. These learned representations can therefore resolve subtle temporal patterns and interactions across sensors that hand-crafted features miss, explaining the empirical gains we report.
    Taken together, these interpretations not only provide deeper ecological context for the classification results, but also highlight how deep embeddings capture phenological patterns better than hand-crafted harmonic + medoid features.}

\subsection{\new{Model interpretability analysis: multi-band, phenology-concentrated attribution}}
\label{sec:interpretability_discussion}

\new{The interpretability analysis in \cref{sec:interpretability_results} lets us move beyond simply observing that fine-tuned Presto embeddings outperform hand-crafted features (\cref{sec:results:harmfeatures,sec:deepembeddings_vs_harmonic}) to characterising \emph{how} the model achieves this. Two findings stand out.}

\new{First, the model does not rely on a single dominant band or index: \cref{fig:gradient_band_importance_grouped} shows that attribution is spread across the short-wave infrared (B11, B12), red (B4), and red-edge/blue (B2, B5) bands for every species, with comparatively little weight on the pre-computed \ac{NDVI} channel itself. This is direct, model-internal evidence for the conceptual argument made in \cref{sec:deepvshandcrafted}: rather than being restricted to a fixed, pre-specified index or a small set of hand-selected harmonic coefficients, the fine-tuned encoder constructs its own nonlinear combinations across the full multi-spectral, multi-polarisation input, and distributes discriminative weight across many bands simultaneously.} 

\new{Second, this multi-band sensitivity is concentrated in specific, phenologically meaningful months rather than spread uniformly across the year: \cref{fig:gradient_temporal_response} shows three recurring elevated windows -- a spring peak in March--April, coinciding with the spring green-up visible directly in the raw signal (\cref{fig:phenology_trajectory,fig:phenology_annex}), a secondary rise in September, and, for several species, a further peak in November -- separated by a shared trough in October. In other words, the model has learned \emph{when} the inter-species phenological differences are largest, and preferentially draws on those months, rather than weighting the full 12-month sequence uniformly -- it is efficiently using many bands, but concentrated on the temporal windows where those bands are actually informative for species discrimination.}

\new{This pattern echoes gradient- and attention-based attribution studies on other satellite image time series tasks: \citet{Russwurm_SelfAttentionRawOpticalSITS_2020} similarly find that a self-attention model trained on raw optical time series concentrates on a small number of phenologically informative dates rather than the full sequence, and \citet{Xu_InterpretingMultiTemporalCropMapping_2021} report that gradient-based attribution for crop mapping is dominated by the growing-season transition window and by the short-wave infrared band. Our results extend this observation from crop and single-region settings to multi-species tree classification at \ac{NFI} scale together with the temporal trajectory analysis similar to \citet{Bressant_InterpretabilityUrbanTreeSITS_2026} for urban tree species.}

\subsection{Limitations and future directions}
\label{limitations}
Several limitations merit consideration for operational deployment. 
First, \newreplace{deep embeddings were extracted from a model pre-trained on a limited temporal and spatial range (2020-2021 global data), and further domain adaptation might improve performance for underrepresented classes to country's forests. Second, while MLPs demonstrated slightly superior performance in our controlled experiments, their generalization in operational settings with smaller, more imbalanced training sets or under different geographic and ecological conditions remains an open question. Future work should investigate robust training strategies, such as cross-validation, targeted hyperparameter optimization, and data augmentation, to better understand how neural classifiers trained on deep embeddings perform when labeled data are scarce or highly skewed. Such studies will help establish practical guidelines for deploying learned representations and downstream classifiers in real-world national forest inventory and monitoring applications.}{}\new{although deep embeddings proved both effective and practical for national forest species mapping, the encoder-adaptation approach (fine-tuned Presto) did not fully realize this potential on the large, class-balanced SIBA dataset. Fine-tuned Presto (with RF and MLP) performed slightly better than the other embeddings on the small, imbalanced NFI datasets (SIMB, COMB) at both the reference-pixel level and in national mapping, but on SIBA its apparent pixel-level advantage largely reflects within-polygon spatial autocorrelation: under a spatially blocked evaluation the frozen pre-computed embeddings (AEF, TESSERA) rank higher, indicating that fine-tuning the encoder on spatially autocorrelated samples can overfit polygon-specific signatures rather than learn more generalizable features. Whether encoder adaptation can outperform general-purpose pre-computed embeddings on large, class-balanced, spatially well-separated data, and under different geographic and ecological conditions, therefore remains open. Future work should investigate robust training strategies and sampling design (cross-validation, targeted hyperparameter optimization, and spatially well-separated data augmentation) to establish when domain-adapted deep embeddings yield more robust, lower-uncertainty maps than pre-computed embeddings for national forest inventory and monitoring.}

\newreplace{Third}{Second}, our datasets derive exclusively from Dutch NFI data. Nonetheless, Dutch forests represent a particularly demanding mapping context, characterised by high fragmentation, small stand sizes, frequent mixed-species composition, and sharp species transitions at the 10m pixel scale \citep{schelhaas_NFI6_2014}. This heterogeneity contrasts with many boreal or continental forest systems where dominant species occur in larger, more homogeneous patches, suggesting that strong performance in the Netherlands provides evidence of robustness under fine-grained spatial variability. However, applying deep embeddings to other datasets \citep{Freudenberg_GermanyTreeSpeciesDataset_2025}, regions, or ecosystems remains essential for assessing generalization across broader ecological contexts.

Future research directions include: (i) adaptation to new regions or tree species\newreplace{ not present in the training data}{}, testing the transferability of learned representations across different forest ecosystems; (ii) integration of additional modalities such as hyperspectral or LiDAR data where available, potentially capturing structural metrics that complement spectral-temporal patterns; and (iii) development of computational efficient uncertainty quantification methods to assess classification confidence for operational decision-making, (iv) inclusion of canopy height products (spaceborne GEDI or regional airborne LiDAR / CHMs) or/and higher spectral resolution data (hyperspectral or narrow-band indices) for the species that showed highest confusion with similar phenological patterns (e.g., Alnus, Fraxinus vs. Quercus) \new{and (v) better strategy on national mapping with class-balanced, spatially well separated data augmentation for actual applications}.

%%%%%%%%%%%%%%%%%%%%%%%%%
\section{Summary and conclusions}
\label{conclusion}

This study examined the efficacy of deep embeddings for tree species classification in National Forest Inventories, yielding several \newreplace{significant }{}findings. First, deep embeddings extracted from pre-trained models consistently outperformed traditional methods employing harmonic and medoid predictors across varying dataset sizes and classification complexities \new{at the reference-pixel level}. \new{Two from-scratch deep-learning baselines (a TempCNN and a from-scratch MLP) trained on identical inputs and splits did not match fine-tuned Presto on the label-scarce NFI datasets, indicating that this advantage stems from self-supervised pre-training rather than from deep learning alone.} \newreplace{In particular, fine-tuned Presto embeddings achieve the best performance despite their small model size.} This performance advantage was particularly evident with larger and more balanced training datasets\newreplace{, as demonstrated by the results from the augumented NFI dataset (SIBA), indicating superior scalability of deep learning approaches with increased high-quality data availability}{, indicating that deep learning approaches scale with the volume and class balance of high-quality labelled data such as the augmented SIBA dataset}. Second, \newreplace{the MLP classifier demonstrated superior performance compared to RF in our experimental setting, corroborating previous findings (Mouret et al., 2024).}{the choice of downstream classifier proved less important than the representation itself: \ac{RF} and the \ac{MLP} classifier performed comparably on the small \ac{NFI} datasets (with \ac{RF} marginally ahead on the more imbalanced task).} 
% \new{Third, from-scratch deep-learning baselines (a TempCNN and a from-scratch MLP) did not match the fine-tuned pre-trained model on the label-scarce NFI datasets, reinforcing the value of self-supervised pre-training.} 
% On the augmented SIBA dataset, the from-scratch MLP appeared to outperform fine-tuned Presto under the standard pixel-level split; a plot-grouped stress-test split traced this to within-polygon spatial autocorrelation rather than a genuine advantage of training from scratch. 
\new{Third, gradient-based attribution of the fine-tuned embedding model showed that it distributes discriminative weight across several spectral bands rather than one index, concentrated in phenologically informative months such as the spring green-up.}
\new{Fourth, country-scale maps built from the pre-computed embeddings reached accuracy consistent with the plot-level results and with an independent comparison against the National Forest Inventory, demonstrating that the approach is not limited to plot-level evaluation and can support operational, wall-to-wall NFI mapping.} Collectively, these results demonstrate the potential for implementing computationally efficient, large-scale tree species mapping within \ac{NFI} systems.

This research \newreplace{significantly}{} advances our understanding of deep learning applications in forestry by \newreplace{having }{}demonstrat\newreplace{ed}{ing} the effectiveness of \newreplace{fine-tuning pre-trained }deep embeddings for tree species classification, establishing a \newreplace{computationally }efficient approach for improving \ac{NFI} accuracy.

\newreplace{This insight lays the groundwork for future developments in automated forest inventory systems and broader applications in environmental monitoring.}

\section*{CRediT authorship contribution statement}
\textbf{Takayuki Ishikawa:} Writing – review \& editing, Writing – original draft, Visualization, Methodology, Investigation, Formal analysis, Conceptualization. \textbf{Carmelo Bonannella:} Writing – review \& editing, Data curation, Supervision, Conceptualization. \textbf{Bas J. W. Lerink\newreplace{}{:}} Writing – review \& editing, Data curation. \textbf{Marc Rußwurm\newreplace{}{:}} Writing – review \& editing, Methodology, Supervision, Conceptualization.

\section*{Declaration of competing interest}
C.B. reports employment with OpenGeoHub Foundation. Other authors declare that they have no known competing financial interests or personal relationships that could have appeared to influence the work reported in this paper.

\section*{\new{Declaration of generative AI and AI-assisted technologies in the manuscript preparation process}}
\new{During the preparation of this work, the authors used Claude Sonnet 5 and Claude Code for coding assistance, literature search, and improving the writing of the manuscript. After using these tools, all scientific interpretations, analyses, results, and conclusions were reviewed and further edited by the authors, who take full responsibility for the content of the published article.}

\section*{Acknowledgement}
This research is based on data from the Dutch National Forest Inventory (NFI), commissioned by the Ministry of Agriculture, Nature and Food Quality of the Netherlands. We thank the Ministry and the agencies responsible for collecting and providing these nationally representative forest data, which underpin our analysis.
C.B. (Carmelo Bonannella) was supported by the Open‑Earth‑Monitor Cyberinfrastructure project, funded by the European Union’s Horizon Europe research and innovation programme [grant agreement No.101059548].

\section*{Data availability}
The source code is available on:

\href{https://github.com/wildflowers315/tree-species-classification-with-presto}{https://github.com/wildflowers315/tree-species-classification-with-presto}.

\bibliography{References}
\bibliographystyle{elsarticle-harv} 

\newpage
\appendix
\setcounter{table}{0}
\renewcommand{\thetable}{A.\arabic{table}}
\onecolumn
\section{Appendix}

\subsection{Implementation \newreplace{D}{d}etails on Presto \newreplace{E}{e}mbeddings}
\label{sec:presto_implmenentation}

\textbf{Calculation of Presto deep embeddings}.
In contrast to harmonic features that fit mathematical curves to individual bands, we evaluate the effectiveness of deep embeddings obtained by the pre-trained time series foundation model, which is Lightweight, Pre-trained Transformers for Remote Sensing Timeseries (Presto) developed and trained by \citet{tseng_Presto_2023} for general-purpose remote sensing time series classification and regression.
This model is based on transformer architecture and was originally pretrained on 21.5 million pixel time-series with 12-month contiguous intervals and static topographic and coordinates data. Each month's composite satellite data included \ac{S1}, S-2 and its \ac{NDVI}, \ac{ERA5}, \ac{DW}, extracted between 2020-01-01 and 2021-12-31. The static \ac{SRTM} and coordinates data are added to the time-series data as input of the \ac{Presto} encoder. 
The model was originally pretrained through \emph{Masked Auto Encoding (MAE)} \citep{he2022masked}\new{ to} reconstruct masked sequences across different satellite modalities. The advantages of this model include the ability to handle freely available multi-source and multi-temporal data even when some sources (e.g., \ac{ERA5}) or temporal data (e.g., S-2 values in November) are missing, and computational efficiency when processing large areas. We obtained 128 dimensional deep embeddings from the encoder output.

For this study, we used the pre-trained weights to initialize the Presto Encoder. 

\ac{MLP} classifier was attached for fine-tuning of \ac{Presto} encoder to the downstream task of tree species classification, which is described in \cref{sec:experiments}. 
We used the fine-tuned encoder to extract deep embeddings from the test set for the comparison between \emph{Harmonic and Seasonal Features} and \emph{\ac{Presto} deep embeddings}.

We directly input 12-monthly time series ($T = 12$ timesteps) for (i) 2 Sentinel-1 bands (S-1), (ii) 10 Sentinel-2 bands (S-2) and (iii) NDVI, (iv) 2 ERA5 features (ERA5), (v) Dynamic World land cover class probabilities (DW). 
To every pixel-time series we appended two static-in-time products: (i) elevation and slope from the SRTM digital elevation model \citep{farr_SRTM_2007} (TG) and (ii) location coordinates (Loc) of each pixel as input tokens to fine-tuned \ac{Presto} encoder. Location coordinates (Loc) are 3D \newreplace{static in time}{static-in-time} Cartesian coordinates computed from the latitude and longitude of the pixel’s geographical location. 

One training sample $\mathbf{x}$, comprising a pixel-timeseries $\mathbf{t} \in \mathbb{R}^{T \times 15}$, land cover classes $\mathbf{v} \in \mathbb{V}^{T \times 1}$, and static variables $\mathbf{s} \in \mathbb{R}^{1 \times 5}$, is summarized as follows:

\begin{equation}
    \mathbf{x} = \left[ \{t^{\text{S1}}_i ; t^{\text{S2}}_i ; t^{\text{ERA5}}_i ; t^{\text{NDVI}}_i ; t^{\text{DW}}_i \,\middle|\, i = 1, \ldots, 12 \,\} ; \, s^{\text{TG}}; s^{\text{Loc}} \right]
\end{equation}

The pixel-timeseries $\mathbf{x}$ were transformed into a sequence of tokens, each represented by an embedding $\mathbf{e}$, suitable for processing by the Presto transformer. For each timestep $0 \leq i < T$, input variables were split into channel groups $\mathcal{C}$ based on their sensor or data source as follows;
\begin{itemize}
  \item \textbf{$\mathbf{e}_\text{S1}$:} S-1 VV and VH bands.
  \item \textbf{$\mathbf{e}_\text{S2-RGB}$:} S-2 RGB(B2, B3, B4) bands.
  \item \textbf{$\mathbf{e}_\text{S2-RE}$:} S-2 Red Edge (B5, B6, B7) bands.
  \item \textbf{$\mathbf{e}_\text{S2-NIR10}$:} S-2 NIR (B8) 10m band.
  \item \textbf{$\mathbf{e}_\text{S2-NIR20}$:} S-2 NIR (B8A) 20m band.
  \item \textbf{$\mathbf{e}_\text{S2-SWIR}$:} S-2 SWIR(B11, B12) bands.
  \item \textbf{$\mathbf{e}_\text{NDVI}$:} Computed from S-2 B4 and B8.
  \item \textbf{$\mathbf{e}_\text{ERA5}$:} \ac{ERA5} Precipitation and 2m temperature.
  \item \textbf{$\mathbf{e}_\text{TG}$:} Elevation and slope from \ac{SRTM} DEM.
  \item \textbf{$\mathbf{e}_\text{Loc}$:} Cartesian coordinates from latitude/longitude.
\end{itemize}

 Each real-valued group was projected into a common latent space of dimension $d_e$ using group-specific learned linear projections $h_C$, such that $e^{\text{S1}}_i = h_{\text{S1}}(t^{\text{S1}}_i)$. 
Categorical variables like Dynamic World classes were embedded via indexing into a learnable embedding matrix. 
To convey the (i) location, (ii) timestamp, and (iii) channel group of each token, specific encodings were added to the embeddings $\mathbf{e}$. 
The final embedding of dimension $d_e$ is a concatenation of:

\begin{itemize}
\item \textbf{Positional Encoding:} Sinusoidal encoding as in \citet{vaswani_attention_2023}.
\item \textbf{Month Encoding:} To represent seasonal similarity, each month ($0$ to $11$) is encoded as:
\begin{align*}
    p_{\text{month}, 2i} &= \sin\left(\frac{2\pi \cdot \text{month}}{12}\right) \\
    p_{\text{month}, 2i+1} &= \cos\left(\frac{2\pi \cdot \text{month}}{12}\right)
\end{align*}
\end{itemize}
For static variables, positional and month encodings were set to zero.
\begin{itemize}
\item \textbf{Channel Group Encoding:} A learnable encoding was added for each channel group $c \in \mathcal{C} = \{\text{S-1}, \text{S-2 RGB}, \ldots, \text{ERA5}, \text{TG}, \text{Loc}\}$, to capture source-specific information.
\end{itemize}

The final transformer input matrix $\mathbf{E} \in \mathbb{R}^{(T \cdot |\mathcal{C}_{\text{dynamic}}| + |\mathcal{C}_{\text{static}}|) \times d_e}$ is composed of:

\begin{itemize}
\item \textbf{Dynamic Variables:} For each timestep $i < T$ and channel group $c \in \mathcal{C}$:
\[
    \mathbf{e}^c_i = h_c(t^c_i) + [p^c_{\text{channel}}; p_{\text{sin}}(i); p_{\text{month}}(i)]
\]
\item \textbf{Topographical Data:}
\[
    \mathbf{e}_{\text{TG}} = h_{\text{TG}}(s_{\text{TG}}) + [p^{\text{TG}}_{\text{channel}}; 0; 0]
\]
\item \textbf{Coordinates:}
\[
    \mathbf{e}_{\text{Loc}} = h_{\text{Loc}}(s_{\text{Loc}})
\]
\end{itemize}

\textbf{Data Normalization (for deep embeddings)}. 
The data normalization follows the original \textit{Presto} setup \citep{tseng_Presto_2023}, with adjustments based on the observed value ranges for each source. S-1 VH and VV bands, originally ranging from --31 to 17~dB, are normalized by shifting +25 and dividing by 25, resulting in a range of --0.24 to 1.68. Sentinel-2 spectral bands, with raw values from 10 to 15,769, are scaled by dividing by 10{,}000, yielding normalized values between 0.00 and 1.58. ERA5 temperature data, originally 278 to 295~K (equivalent to 6 to 23\textdegree{}C), are shifted by --272.15 and divided by 35, producing a range of 0.17 to 0.66. ERA5 precipitation, ranging from 0.008 to 0.208~m, is normalized by dividing by 0.03, resulting in values from 0.27 to 6.93. SRTM elevation data, spanning --27 to 331~m, are divided by 2{,}000 to give a range of --0.01 to 0.17, while slope values from 0.0 to 39.3\textdegree{} are scaled by dividing by 50, yielding normalized values between 0.00 and 0.79.

\subsection{Details of methodology}
\subsubsection{Data \newreplace{S}{s}plitting and training pipeline}
The detailed breakdown of class distribution across train, validation, and test splits is provided in \cref{tab:class_distribution_splits}.
For fine-tuning Presto, we used MLP-Train data as training data for the MLP classifier and evaluated loss values based on MLP-validation data.

\begin{table}[!ht]
    \centering
    \caption{Species/Group class distribution across train, validation, and test splits for three datasets.}
    \label{tab:class_distribution_splits}
    \footnotesize
    \begin{tabular}{lrrrrrr}
        \toprule
        \textbf{Dataset} & \textbf{Class} & \multicolumn{3}{c}{\textbf{Train}} & \textbf{Test} & \textbf{Total} \\
        \cmidrule(lr){4-5}
        & & \textbf{(train total)} & \textbf{MLP-Train} & \textbf{MLP-Val} & & \\
        \midrule
        \textit{\newreplace{NFI (dominantSpecies)}{COMB (13 classes \& imbalanced)}}           & \textit{Alnus} spp                       &     21 &     16 &      5 &      9 &     30 \\
                                                 & \textit{Betula} spp                      &     40 &     32 &      8 &     18 &     58 \\
                                                 & \textit{Fagus} spp                       &     40 &     32 &      8 &     18 &     58 \\
                                                 & \textit{Fraxinus} spp                    &     28 &     22 &      6 &     12 &     40 \\
                                                 & \textit{Larix} spp                       &     39 &     31 &      8 &     17 &     56 \\
                                                 & \textit{Other Pinus}                     &     62 &     49 &     13 &     27 &     89 \\
                                                 & \textit{Other Quercus}                   &     23 &     18 &      5 &     10 &     33 \\
                                                 & Other broadleaved                        &     71 &     56 &     15 &     31 &    102 \\
                                                 & \textit{Picea} spp                       &     46 &     36 &     10 &     20 &     66 \\
                                                 & \textit{Pinus sylvestris}                &    359 &    287 &     72 &    154 &    513 \\
                                                 & \textit{Populus} spp                     &     50 &     40 &     10 &     22 &     72 \\
                                                 & \textit{Pseudotsuga menziesii}           &     62 &     49 &     13 &     28 &     90 \\
                                                 & \textit{Quercus robur petraea}           &    178 &    142 &     36 &     77 &    255 \\
        \cmidrule(l){1-7}
        & \textbf{Total} &  1,019 &    810 &    209 &    443 &  1,462 \\
        \midrule
        \textit{\newreplace{NFI (Group\_F)}{SIMB (7 classes \& imbalanced)}}                  & Beech                                    &     40 &     32 &      8 &     18 &     58 \\
                                                 & Dark Conifer                             &    112 &     89 &     23 &     48 &    160 \\
                                                 & Larix                                    &     39 &     31 &      8 &     17 &     56 \\
                                                 & Other Broadleaves                        &    169 &    135 &     34 &     73 &    242 \\
                                                 & Pinus                                    &    422 &    337 &     85 &    181 &    603 \\
                                                 & Populus                                  &     50 &     40 &     10 &     22 &     72 \\
                                                 & Quercus                                  &    201 &    160 &     41 &     87 &    288 \\
        \cmidrule(l){1-7}
        & \textbf{Total} &  1,033 &    824 &    209 &    446 &  1,479 \\
        \midrule
        \textit{\newreplace{Francini (group2)}{SIBA (7 classes \& balanced)}}               & Beech                                    &  1,379 &  1,103 &    276 &    591 &  1,970 \\
                                                 & Dark Conifer                             &  1,379 &  1,103 &    276 &    591 &  1,970 \\
                                                 & Larix                                    &  1,379 &  1,103 &    276 &    591 &  1,970 \\
                                                 & Other Broadleaf                          &  1,379 &  1,103 &    276 &    591 &  1,970 \\
                                                 & Pinus                                    &  1,379 &  1,103 &    276 &    591 &  1,970 \\
                                                 & Populus                                  &  1,379 &  1,103 &    276 &    591 &  1,970 \\
                                                 & Quercus                                  &  1,379 &  1,103 &    276 &    591 &  1,970 \\
        \cmidrule(l){1-7}
        & \textbf{Total} &  9,653 &  7,721 &  1,932 &  4,137 & 13,790 \\
        \bottomrule
    \end{tabular}
\end{table}

\subsubsection{\new{Compute and wall-clock cost}}
\label{app:compute_cost}
\new{All fine-tuning and embedding extraction experiments were conducted on Google Colab using an NVIDIA Tesla T4 GPU (16GB GPU memory). Each fine-tuning run used a maximum of 100 training epochs, but due to early stopping based on validation performance, most runs terminated in fewer than 30 epochs on SIMB (1,479 samples) and COMB (1,462 samples), and 50 epochs on SIBA (13,790 samples). As a result, a single fine-tuning run on the smaller SIMB and COMB datasets takes approximately 1 minute, with embedding extraction requiring an additional ~1 minute per dataset. For the larger SIBA dataset, a single fine-tuning run takes approximately 10 minutes and embedding extraction takes ~3 minutes.}

\subsubsection{\new{Why pooled embeddings require freshly trained classifiers}}
\label{app:spatial_patch_classifier_rationale}
\new{At the single center pixel, two classifiers already exist for fine-tuned Presto (\cref{sec:experiments}): an \ac{MLP} head trained jointly with the encoder during the original fine-tuning step (the \emph{Fine-tune Presto with \ac{MLP} Classifier} column of \cref{tab:RQ1_combined}), and a separate \ac{RF} trained afterwards on the resulting 128-dimensional embeddings with the encoder held fixed (the \emph{Fine-tune Presto} \ac{RF} column). Neither of these two pretrained classifiers can be reused directly on the pooled embeddings from \cref{sec:spatial_patch_methods}. For flatten and stats-pool this is a hard constraint, since pooling changes the input dimensionality (1152-d and 512-d, respectively) and neither pretrained classifier's input layer matches. For mean-pool the pooled vector keeps the original 128 dimensions, but its values are no longer a single-pixel embedding: it is the element-wise average of the center pixel's embedding with its eight neighbours, which shifts the feature statistics away from what either pretrained classifier was calibrated to during its own training. We therefore train a freshly initialised \ac{RF} and \ac{MLP} (the latter using the same architecture as \cref{tab:hyperparams}) on the pooled embeddings for every pooling strategy, including mean-pool, so that only the fixed encoder is shared with the center-pixel result and the comparison isolates the effect of pooling itself rather than confounding it with whether the classifier was ever exposed to pooled-feature statistics during its own training.}

\subsection{\new{Model and training configuration}}
\new{\Cref{tab:hyperparams} summarises the training configuration of the fine-tuned Presto \ac{MLP} head and the two from-scratch baselines (from-scratch \ac{MLP} and TempCNN), and \cref{tab:input_mapping} maps each evaluated approach to its input, feature representation, and downstream classifier.}

\begin{table*}[ht]
    \centering
    \footnotesize
    \caption{\new{Training configuration of the deep-learning models: the \ac{MLP} head used for Presto fine-tuning, the from-scratch \ac{MLP}, and the from-scratch TempCNN. The three share an identical training schedule and differ only in input, architecture, and architecture-specific regularization. Settings follow \citet{Mouret_TreeSpeciesClassificationPixelTimeSeriesInbalanced_2024} (training protocol and \ac{MLP} architecture) and \citet{Pelletier_TempCNN_2019} (TempCNN architecture and its dropout and weight decay).}}
    \label{tab:hyperparams}
    \begin{newtable}
    \begin{threeparttable}
    \begin{tabular}{
        >{\raggedright\arraybackslash}p{3.0cm}
        >{\raggedright\arraybackslash}p{3.4cm}
        >{\raggedright\arraybackslash}p{3.4cm}
        >{\raggedright\arraybackslash}p{3.4cm}}
    \toprule
    & \textbf{MLP (Presto fine-tune)} & \textbf{MLP (from scratch)} & \textbf{TempCNN (from scratch)} \\
    \midrule
    Input & Presto embedding (fine-tuned encoder) & flattened raw $12\times17$ (204-d) & raw $12\times17$ time series \\
    Learning rate & $1\times10^{-4}$ & $1\times10^{-4}$ & $1\times10^{-3}$ \\
    Weight decay & 0 & 0 & $1\times10^{-6}$ \\
    Dropout & 0.0 & 0.0 & 0.5 \\
    Hidden units / conv.\ filters & 1024, 512, 256 & 1024, 512, 256 & 128, 128, 128 (kernels 3, 3, 2) \\
    Trainable parameters & $\sim$1.2M (0.4M encoder + 0.79M head) & $\sim$0.87M & $\sim$91K \\
    Epochs / batch size & 100 / 64 & 100 / 64 & 100 / 64 \\
    Adaptive learning rate & $\times$0.5 after 20 epochs without improvement & $\times$0.5 after 20 epochs without improvement & $\times$0.5 after 20 epochs without improvement \\
    Early stopping & 40 epochs without improvement & 40 epochs without improvement & 40 epochs without improvement \\
    Loss / model selection & cross-entropy / best validation loss & cross-entropy / best validation loss & cross-entropy / best validation loss \\
    \bottomrule
    \end{tabular}
    \end{threeparttable}
    \end{newtable}
\end{table*}

\begin{table*}[ht]
    \centering
    \footnotesize
    \caption{\new{Input, feature representation, and downstream classifier for each approach evaluated. RF is the common downstream classifier across all feature representations; the ``MLP (Presto fine-tune)'' result is the end-to-end fine-tuned-Presto head; AEF and TESSERA embeddings are pre-computed and retrieved by coordinate; the harmonic + medoid features, the from-scratch \ac{MLP}, and TempCNN consume the raw $12\times17$ time series directly.}}
    \label{tab:input_mapping}
    \begin{newtable}
    \begin{threeparttable}
    \begin{tabular}{
        >{\raggedright\arraybackslash}p{3.0cm}
        >{\raggedright\arraybackslash}p{2.4cm}
        >{\raggedright\arraybackslash}p{4.3cm}
        >{\raggedright\arraybackslash}p{2.6cm}
        >{\centering\arraybackslash}p{1.4cm}}
    \toprule
    \textbf{Approach} & \textbf{Input} & \textbf{Feature representation} & \textbf{Classifier} & \textbf{Pre-trained} \\
    \midrule
    Harmonic + medoid & raw $12\times17$ & 209 hand-crafted (harmonic + seasonal medoid) & RF & N/A \\
    Frozen Presto & raw $12\times17$ & Presto embedding (frozen) & RF & Yes (frozen) \\
    Fine-tuned Presto & raw $12\times17$ & Presto embedding (fine-tuned) & RF & Yes \\
    AEF & pre-computed & AlphaEarth embedding (frozen, by coordinate) & RF & Yes (frozen) \\
    TESSERA & pre-computed & TESSERA embedding (frozen) & RF & Yes (frozen) \\
    Fine-tuned Presto + MLP & raw $12\times17$ & Presto encoder (fine-tuned, end-to-end) & MLP head & Yes \\
    MLP (from scratch) & raw $12\times17$ (flattened, 204-d) & learned internally & MLP (1024-512-256) & No \\
    TempCNN (from scratch) & raw $12\times17$ & learned internally & TempCNN & No \\
    \bottomrule
    \end{tabular}
    \end{threeparttable}
    \end{newtable}
\end{table*}

\subsection{\new{Spatially blocked (plot-grouped) evaluation of SIBA}}
\label{app:siba_spatial}
\new{\Cref{tab:siba_spatial} reports the SIBA results under the standard pixel-level split and under the plot-grouped (spatially blocked) split described in \cref{sec:experiments}. The pixel-level columns reproduce the SIBA values of \cref{tab:RQ1_combined}.}

\begin{table*}[ht]
    \centering
    \footnotesize
    \setlength{\tabcolsep}{4pt}
    \caption{\new{Spatially blocked evaluation of the SIBA dataset: classification performance (in \%) under the standard pixel-level split versus a plot-grouped (spatially blocked) split that keeps all pixels of a photo-interpreted polygon in a single partition. Mean and standard deviation (±) over five seeds. Bold values indicate the best method per metric and split. The pixel-level columns correspond to the SIBA rows of \cref{tab:RQ1_combined}.}}
    \label{tab:siba_spatial}
    \begin{newtable}
    \begin{threeparttable}
    \begin{tabular}{
        >{\raggedright}p{3.2cm}
        >{\centering}p{1.9cm}
        >{\centering}p{1.9cm}
        >{\centering}p{1.9cm}
        >{\centering}p{1.9cm}
        >{\centering}p{1.9cm}
        >{\centering\arraybackslash}p{1.9cm}}
    \toprule
    & \multicolumn{2}{c}{Overall Accuracy} & \multicolumn{2}{c}{F1 Macro} & \multicolumn{2}{c}{F1 Weighted} \\
    \cmidrule(lr){2-3}\cmidrule(lr){4-5}\cmidrule(lr){6-7}
    Method & Pixel & Grouped & Pixel & Grouped & Pixel & Grouped \\
    \midrule
    MLP (fine-tune\new{d} Presto)  & 97.26 ± 0.43 & 48.97 ± 3.14 & 97.26 ± 0.43 & 46.96 ± 3.03 & 97.26 ± 0.43 & 47.04 ± 3.00 \\
    Presto frozen (RF)      & 84.51 ± 0.33 & 49.84 ± 3.48 & 84.32 ± 0.36 & 47.77 ± 3.86 & 84.32 ± 0.36 & 47.94 ± 3.84 \\
    Fine-tune Presto (RF)   & 94.50 ± 0.45 & 53.03 ± 3.50 & 94.48 ± 0.45 & 50.85 ± 3.41 & 94.48 ± 0.45 & 50.95 ± 3.44 \\
    Harmonic + medoid (RF)  & 84.47 ± 0.41 & 55.20 ± 5.65 & 84.40 ± 0.40 & 53.30 ± 5.84 & 84.40 ± 0.40 & 53.25 ± 6.08 \\
    AEF (RF)                & 90.32 ± 0.65 & \textbf{62.02 ± 4.25} & 90.20 ± 0.65 & 59.68 ± 4.20 & 90.20 ± 0.65 & 59.64 ± 4.72 \\
    TESSERA (RF)            & 75.65 ± 0.64 & 60.78 ± 2.78 & 75.66 ± 0.62 & \textbf{60.20 ± 2.42} & 75.66 ± 0.62 & \textbf{60.22 ± 2.58} \\
    TempCNN (scratch)       & 92.19 ± 0.46 & 44.71 ± 3.69 & 92.08 ± 0.49 & 42.44 ± 3.23 & 92.08 ± 0.49 & 42.50 ± 3.20 \\
    MLP (scratch)           & \textbf{98.32 ± 0.26} & 44.80 ± 2.70 & \textbf{98.32 ± 0.26} & 41.17 ± 3.81 & \textbf{98.32 ± 0.26} & 41.28 ± 3.71 \\
    \bottomrule
    \end{tabular}
    \end{threeparttable}
    \end{newtable}
\end{table*}

\new{The plot-grouped split is an intentionally pessimistic bound rather than a recommended evaluation protocol for inventory data. It removes \emph{all} within-polygon pixels from training, so the classifier must generalise across photo-interpreted stands with as few as roughly 25 polygons in the rarest class; this is considerably stricter than the 15\,m pixel thinning used in the original SIBA study \citep{francini_DutchForestSpeciesMap_2024} and is not applied in that work or in other NFI-based tree-species mapping studies, which rely on systematic-grid or (stratified) random sampling with broad spatial coverage \citep{blickensdorfer_GermanTreeSpeciesMap_2024, hermosilla_CanadaTreeSpeciesMapping_2022, Freudenberg_S2-TreeSpeciesClassificationGermany_2025}. The true generalisation performance is therefore bracketed between the optimistic pixel-level split and this conservative plot-grouped split. The main value of this analysis is not the absolute accuracies but the change in \emph{ranking}: under spatial blocking the foundation-model embeddings (AEF and TESSERA) become the strongest representations, whereas the from-scratch MLP, which dominates the pixel-level split, collapses to near the bottom, indicating that its high pixel-level accuracy is driven by spatial autocorrelation rather than transferable structure \citep{Karasiak_SpatialDependenceTrain-Test_2022, Ploton_SpatialValidation_2020}.}

\subsection{\new{Full-band phenological trajectories (SIMB)}}
\label{app:phenology_annex}

\begin{figure*}[ht]
    \centering
    \includegraphics[width=17cm]{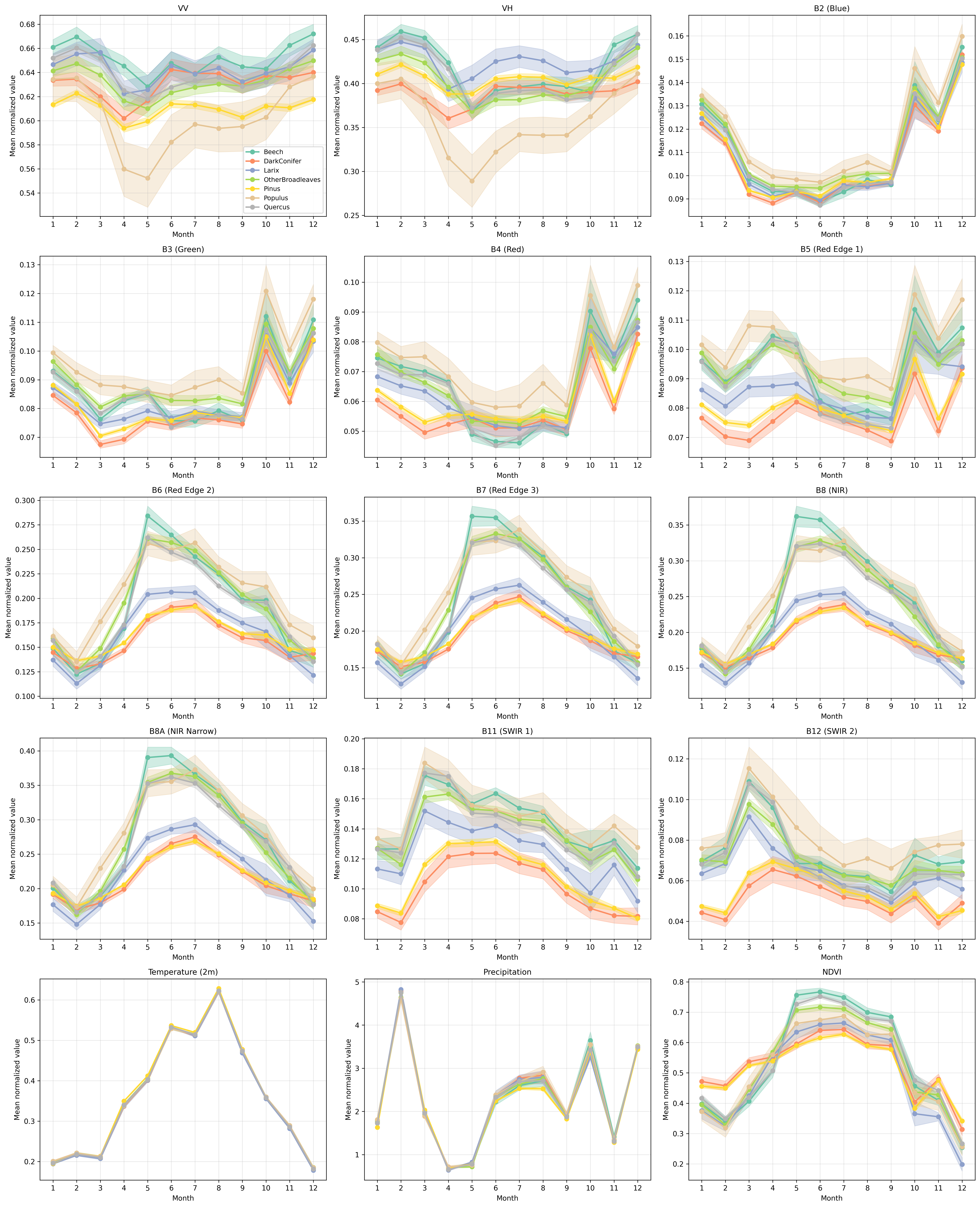}
    \caption{\new{\textit{Per-species mean monthly trajectories for all 15 dynamic bands (SIMB, 7 classes; shaded band: 95\% confidence interval of the class mean).}}}
    \label{fig:phenology_annex}
\end{figure*}

\new{\cref{fig:phenology_annex} extends \cref{fig:phenology_trajectory} (\cref{sec:interpretability_results}) to all 15 dynamic input bands (S-1 VV/VH, the 10 S-2 bands, \ac{NDVI}, and the 2 \ac{ERA5} bands), each panel showing per-species mean monthly trajectories with 95\% confidence intervals of the class mean on the SIMB dataset. The static \ac{SRTM} bands (elevation, slope) are the only ones omitted, as they have no monthly trajectory to plot. The \ac{ERA5} panels confirm, rather than merely assume, that temperature and precipitation are essentially indistinguishable across species: all seven per-species curves overlap almost exactly, consistent with \ac{ERA5}'s coarse spatial resolution relative to the extent of this study, in contrast to the clear inter-species separation visible in the optical and radar bands.}

\subsection{\new{Gradient-based attribution for COMB (13-class), a robustness check}}
\label{app:comb_gradient}

\begin{figure}[ht]
    \centering
    \includegraphics[width=\linewidth]{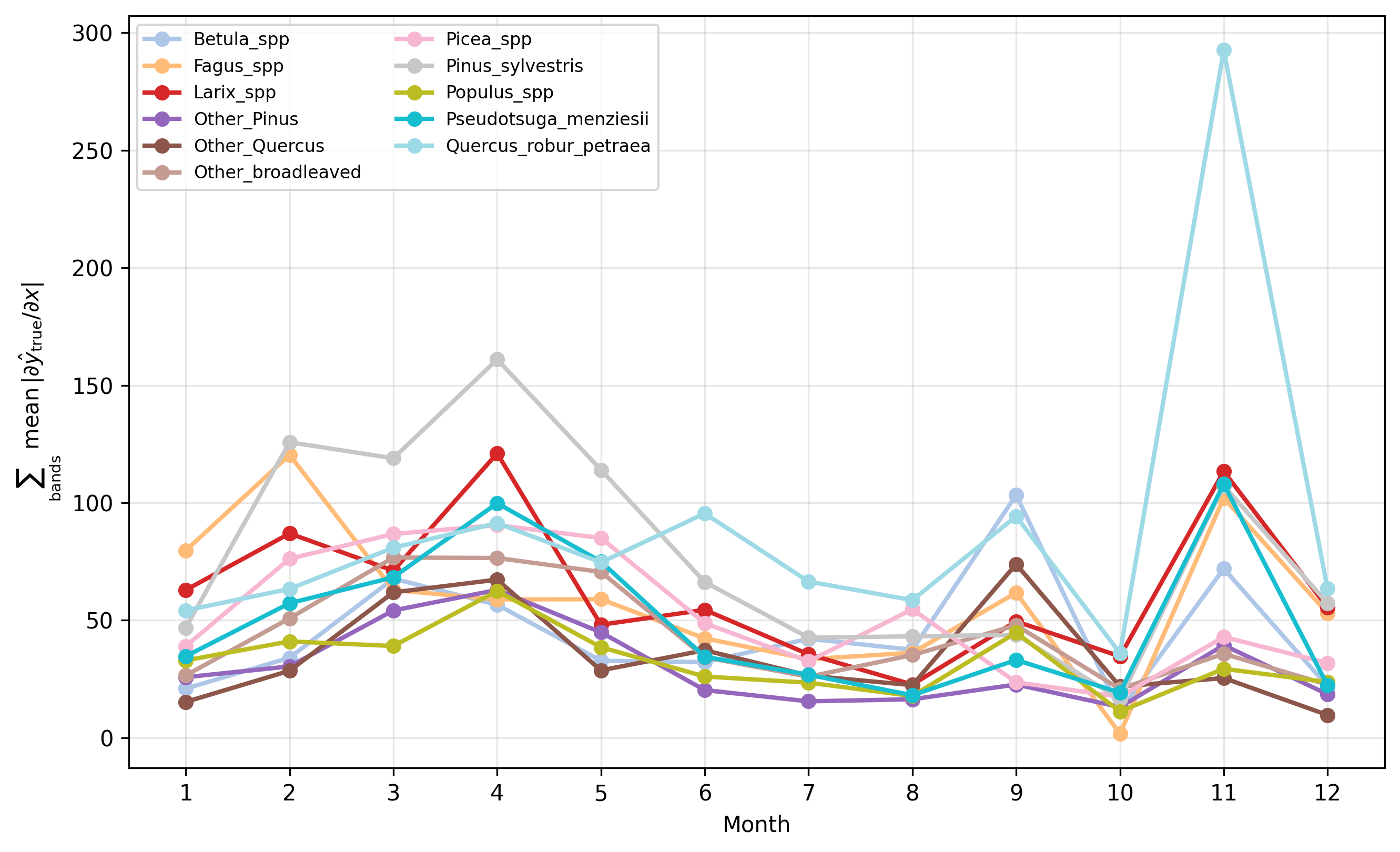}
    \caption{\new{\textit{Gradient-based temporal-response attribution for the fine-tuned Presto \ac{MLP} model (COMB, 13 classes, correctly classified test samples, rs~=~42): per-species $\sum_{\text{bands}} \text{mean}\,|\partial \hat{y}_{\text{true}}/\partial x|$ across the 15 dynamic input bands, by month.}}}
    \label{fig:gradient_temporal_response_comb}
\end{figure}

\begin{figure*}[ht]
    \centering
    \includegraphics[width=18cm]{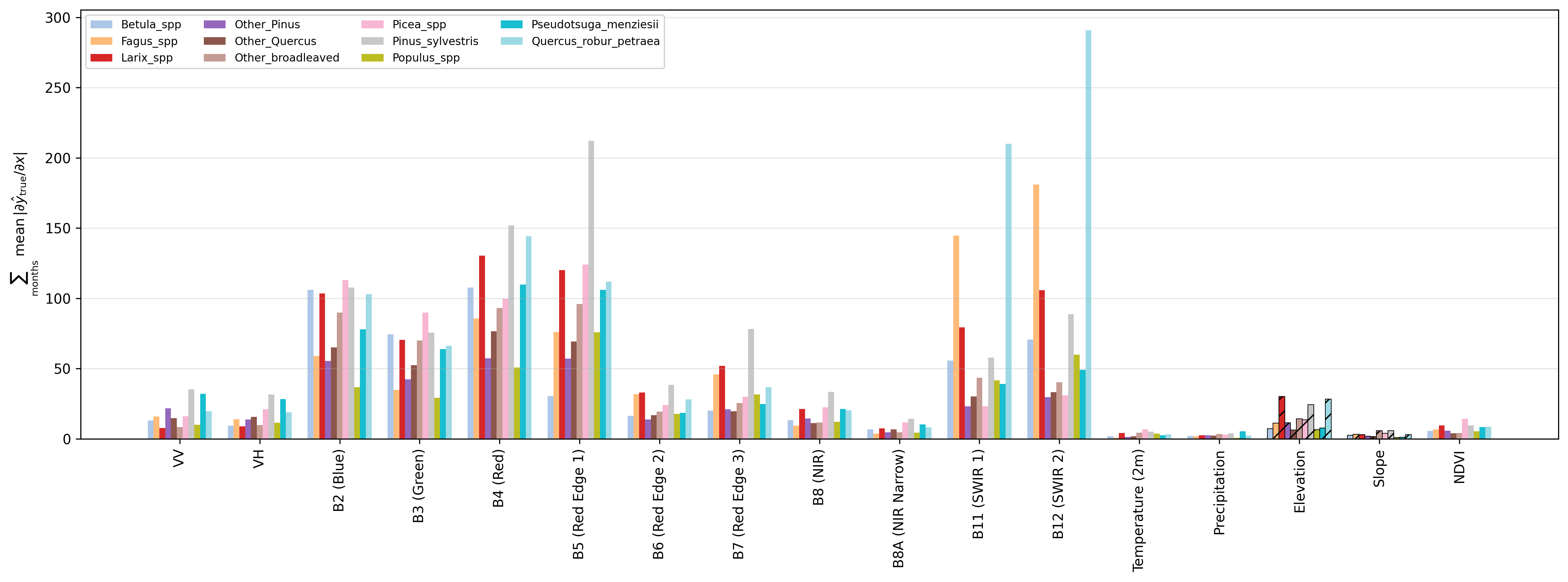}
    \caption{\new{\textit{Gradient-based band importance for the fine-tuned Presto \ac{MLP} model (COMB, 13 classes, correctly classified test samples, rs~=~42), grouped by species within each band: per-species $\sum_{\text{months}} \text{mean}\,|\partial \hat{y}_{\text{true}}/\partial x|$ for each of the 17 input bands (hatched bars: static \ac{SRTM} bands).}}}
    \label{fig:gradient_band_importance_grouped_comb}
\end{figure*}

\new{The gradient-based attribution analysis (\cref{sec:experiments,sec:interpretability_results}) is repeated on the 13-class COMB dataset, using the corresponding fine-tuned Presto \ac{MLP} checkpoint and the same correctly-classified-only restriction. \cref{fig:gradient_temporal_response_comb,fig:gradient_band_importance_grouped_comb} give the corresponding results to \cref{fig:gradient_temporal_response,fig:gradient_band_importance_grouped}. The static \ac{SRTM} bands are again excluded from the temporal sum in \cref{fig:gradient_temporal_response_comb} because they have no month-to-month variation to attribute.}

\new{The well-sampled classes reproduce the qualitative pattern found for SIMB: attribution is dominated by the short-wave infrared (B11, B12), red (B4), and red-edge (B5) bands, and the temporal-response curves for \textit{Pinus sylvestris} and \textit{Quercus robur/petraea} reproduce the same three-window pattern described for SIMB in \cref{sec:interpretability_results}, with elevated attribution in the spring green-up months and again around September, separated by a shared trough in October. Two differences from SIMB are worth noting. First, COMB's long tail of small classes (several with fewer than 10 correctly-classified test samples, e.g.\ \textit{Fagus spp}, n~=~2) makes the corresponding curves too noisy to interpret individually; this is an expected consequence of the finer-grained, more imbalanced class scheme (\cref{sec:data}) rather than a property of the attribution method itself. Second, \textit{Quercus robur/petraea} shows a pronounced attribution peak in November that is not visible for any other class -- a pattern also present, though less pronounced, in the SIMB \textit{Quercus} curve (\cref{fig:gradient_temporal_response}). A plausible ecological explanation is marcescence: oaks characteristically retain dead leaves later into autumn/winter than most other broadleaved species in this study, which could make November a genuinely discriminating month for this class; we flag this as a candidate hypothesis rather than a confirmed mechanism, consistent with our single-seed, single-pipeline scope (\cref{sec:interpretability_discussion}).}

\subsection{Additional results}
\subsubsection{Results of models using only S-1 and S-2 data for fine-tuning and embeddings extraction}
Our primary models incorporated all available satellite data as input for fine-tuning and extraction of deep embeddings from Presto. In this section, we present results from models using only S-1 and S-2 data as input for fine-tuning and embeddings extraction, excluding \ac{SRTM} and \ac{ERA5} data.

\paragraph{\newreplace{d}{D}eep embeddings model performance} \leavevmode\\
Table\newreplace{}{~}\ref{tab:RQ1_results1_S1S2only} reports classification performance \newreplace{in}{using the} RF classifier when fine-tuning Presto using only S-1 and S-2 data for fine-tune and embeddings extraction, excluding auxiliary topographic (\ac{SRTM}) and climatic (\ac{ERA5}) inputs. Across all datasets, fine-tuned Presto embeddings substantially outperform harmonic + medoid features for all evaluation metrics, highlighting the strong representational capacity of deep embeddings even when limited to optical and radar data. These results indicate that S-1 + S-2 only fine-tuning Presto embeddings remain\newreplace{s}{} highly effective and robust.

\begin{table*}[htbp]
    \centering
    \small
    \caption{Classification performance comparison between fine-tuned Presto embeddings using only S-1 and S-2 inputs and harmonic + medoid features (in \%). Results are reported for a RF classifier. Each model was run five times and the mean and standard deviation (±) are reported. Bold values indicate the best performance between the models per dataset type.}
    \label{tab:RQ1_results1_S1S2only}
    \begin{threeparttable}
    \begin{tabular}{
        >{\raggedright}p{1.5cm}
        >{\raggedright}p{2.5cm}
        >{\raggedleft}p{1.2cm}
        >{\centering}p{2.6cm}
        >{\centering\arraybackslash}p{2.2cm}}
    \toprule
    & & & \multicolumn{2}{c}{RF} \\
    \cmidrule(lr){4-5}
    & Dataset type & Data size 
    & Fine-tune Presto\newline embeddings
    & Harm.+med.\newline features\newline (\citep{francini_DutchForestSpeciesMap_2024})\\
    \midrule
    \multirow{3}{*}{\begin{tabular}{@{}l@{}}Overall\\Accuracy\end{tabular}}
    & SIMB (7 classes) & 1,479
        & \textbf{\newreplace{74.01 ± 1.83}{79.46 ± 0.96}}
        & 69.48 ± 0.73 \\
    & COMB (13 classes) & 1,462
        & \textbf{\newreplace{65.42 ± 1.92}{71.15 ± 2.14}}
        & 62.80 ± 0.91 \\
    & SIBA (7 classes) & 13,790
        & \textbf{\newreplace{92.31 ± 0.18}{94.64 ± 0.36}}
        & 84.47 ± 0.41 \\
    \midrule
    \multirow{3}{*}{F1 Macro}
    & SIMB (7 classes) & 1,479
        & \textbf{\newreplace{60.00 ± 2.12}{70.13 ± 1.85}}
        & 51.03 ± 1.71 \\
    & COMB (13 classes) & 1,462
        & \textbf{\newreplace{44.67 ± 3.25}{55.55 ± 2.34}}
        & 38.95 ± 2.02 \\
    & SIBA (7 classes) & 13,790
        & \textbf{\newreplace{91.95 ± 0.24}{94.62 ± 0.36}}
        & 84.40 ± 0.40 \\
    \midrule
    \multirow{3}{*}{F1 Weighted}
    & SIMB (7 classes) & 1,479
        & \textbf{\newreplace{72.96 ± 1.87}{79.09 ± 1.13}}
        & 66.98 ± 1.06 \\
    & COMB (13 classes) & 1,462
        & \textbf{\newreplace{61.32 ± 2.10}{69.00 ± 2.28}}
        & 57.13 ± 1.25 \\
    & SIBA (7 classes) & 13,790
        & \textbf{\newreplace{92.23 ± 0.20}{94.62 ± 0.36}}
        & 84.40 ± 0.40 \\
    \bottomrule
    \end{tabular}
    \end{threeparttable}
\end{table*}

%% MLP vs RF

Table \ref{tab:RQ1_results2_S1S2only} compares the performance of \ac{MLP} and \ac{RF} classifiers using fine-tuned Presto embeddings when only S-1 and S-2 data are provided as input. \newreplace{The \ac{MLP} classifier consistently outperforms \ac{RF} on the SIMB and COMB datasets across all metrics, with particularly notable improvements in F1 macro and weighted scores for the more class-imbalanced COMB dataset. In contrast, performance differences between the two classifiers are marginal for the SIBA dataset, where both models achieve similarly high accuracy and F1 scores. This suggests that classifier choice has a stronger impact under reduced input modalities and class imbalance, while its effect diminishes for larger and more balanced datasets.}{\ac{RF} slightly outperforms the \ac{MLP} classifier on the SIMB and COMB datasets across all metrics, whereas the \ac{MLP} classifier pulls clearly ahead on the SIBA dataset. This mirrors the pattern observed with all available bands (\cref{sec:deepembeddings_vs_harmonic}): a robust classifier such as RF is sufficient under label scarcity, whereas the parametric MLP head benefits more from the larger, balanced SIBA dataset.}

\begin{table*}[htbp]
    \centering
    \small
    \caption{Classification performance comparison between \ac{MLP} and \ac{RF} classifiers using fine-tuned Presto embeddings with only S-1 and S-2 inputs (in \%). Each model was run five times and the mean and standard deviation (±) are reported. Bold values indicate the best performance between the \ac{MLP} and \ac{RF} classifiers per dataset type.}
    \label{tab:RQ1_results2_S1S2only}
    \begin{threeparttable}
    \begin{tabular}{
        >{\raggedright}p{1.5cm}
        >{\raggedright}p{2.5cm}
        >{\raggedleft}p{1.2cm}
        >{\centering}p{2.6cm}
        >{\centering\arraybackslash}p{2.2cm}}
        \toprule
        & & & \multicolumn{2}{c}{Fine-tune Presto embeddings} \\
        \cmidrule(lr){4-5}
        & Dataset type & Data size & \ac{MLP} & \ac{RF} \\
        \midrule
        \multirow{3}{*}{\begin{tabular}{@{}l@{}}Overall\\Accuracy\end{tabular}}
        & SIMB (7 classes) & 1,479
            & \newreplace{\textbf{74.98 ± 1.78}}{78.28 ± 1.05} & \newreplace{74.01 ± 1.83}{\textbf{79.46 ± 0.96}} \\
        & COMB (13 classes) & 1,462
            & \newreplace{\textbf{67.09 ± 0.60}}{69.75 ± 1.45} & \newreplace{65.42 ± 1.92}{\textbf{71.15 ± 2.14}} \\
        & SIBA (7 classes) & 13,790
            & \newreplace{92.23 ± 0.48}{\textbf{97.28 ± 0.17}} & \newreplace{\textbf{92.31 ± 0.18}}{94.64 ± 0.36} \\
        \midrule
        \multirow{3}{*}{F1 Macro}
        & SIMB (7 classes) & 1,479
            & \newreplace{\textbf{65.09 ± 2.97}}{68.43 ± 2.83} & \newreplace{60.00 ± 2.12}{\textbf{70.13 ± 1.85}} \\
        & COMB (13 classes) & 1,462
            & \newreplace{\textbf{50.82 ± 3.87}}{52.23 ± 1.99} & \newreplace{44.67 ± 3.25}{\textbf{55.55 ± 2.34}} \\
        & SIBA (7 classes) & 13,790
            & \newreplace{91.91 ± 0.54}{\textbf{97.28 ± 0.17}} & \newreplace{\textbf{91.95 ± 0.24}}{94.62 ± 0.36} \\
        \midrule
        \multirow{3}{*}{F1 Weighted}
        & SIMB (7 classes) & 1,479
            & \newreplace{\textbf{74.49 ± 1.88}}{77.80 ± 1.23} & \newreplace{72.96 ± 1.87}{\textbf{79.09 ± 1.13}} \\
        & COMB (13 classes) & 1,462
            & \newreplace{\textbf{65.00 ± 1.28}}{67.59 ± 1.58} & \newreplace{61.32 ± 2.10}{\textbf{69.00 ± 2.28}} \\
        & SIBA (7 classes) & 13,790
            & \newreplace{92.20 ± 0.49}{\textbf{97.28 ± 0.17}} & \newreplace{\textbf{92.23 ± 0.20}}{94.62 ± 0.36} \\
        \bottomrule
    \end{tabular}
    \end{threeparttable}
\end{table*}

\paragraph{Comparison between all bands and only S-1 and S-2 bands} \leavevmode\\

Table \ref{tab:RQ1_results3_All_S1S2_comparison} compares fine-tuned Presto embeddings obtained using all available satellite bands with those using only S-1 and S-2 data, evaluated with the \ac{MLP} classifier. Overall, performance differences between the two band configurations \newreplace{are small across datasets, indicating that excluding auxiliary climatic (ERA5) and topographic (SRTM) information does not substantially degrade classification performance for this task.}{are modest across datasets: minimal (within about 1 percentage point) for SIMB and COMB, and likewise minimal for SIBA, where the two configurations are effectively indistinguishable.}

\newreplace{For the SIMB dataset, using only S-1 and S-2 bands yields slightly higher accuracy and F1 scores, whereas for the COMB dataset both configurations perform comparably. In contrast, for the SIBA dataset, which is larger and more balanced, the use of all available bands results in marginally higher performance across all metrics. These results suggest that optical and radar data alone are sufficient to capture most discriminative information, while auxiliary bands provide limited additional benefit that appears to be dataset-dependent.}{For both the SIMB and COMB datasets, using only S-1 and S-2 bands yields slightly higher accuracy and F1 scores than using all available bands. In contrast, for the SIBA dataset, which is larger and more balanced, the use of all available bands results in a more clearly higher performance across all metrics. These results suggest that optical and radar data alone capture most of the discriminative information on the smaller NFI datasets, while the larger, more balanced SIBA dataset benefits more from the auxiliary climatic and topographic bands.}

\begin{table*}[htbp]
    \centering
    \small
    \caption{Classification performance comparison between using all available bands and using only S-1 and S-2 bands for fine-tuned Presto embeddings with an \ac{MLP} classifier (in \%). Each model was run five times and the mean and standard deviation (±) are reported. Bold values indicate the best performance between the two band configurations for each dataset type.}
    \label{tab:RQ1_results3_All_S1S2_comparison}
    \begin{threeparttable}
    \begin{tabular}{
        >{\raggedright}p{1.5cm}
        >{\raggedright}p{2.5cm}
        >{\raggedleft}p{1.2cm}
        >{\centering}p{2.6cm}
        >{\centering\arraybackslash}p{2.6cm}}
        \toprule
        & & & \multicolumn{2}{c}{MLP for fine-tune embeddings} \\
        \cmidrule(lr){4-5}
        & Dataset type & Data size & All bands & S-1 + S-2 only \\
        \midrule
        \multirow{3}{*}{\begin{tabular}{@{}l@{}}Overall\\Accuracy\end{tabular}}
        & SIMB (7 classes) & 1,479
            & \newreplace{73.67 ± 1.29}{78.05 ± 2.20} & \newreplace{\textbf{74.98 ± 1.78}}{\textbf{78.28 ± 1.05}} \\
        & COMB (13 classes) & 1,462
            & \newreplace{\textbf{67.40 ± 1.47}}{69.35 ± 1.08} & \newreplace{67.09 ± 0.60}{\textbf{69.75 ± 1.45}} \\
        & SIBA (7 classes) & 13,790
            & \newreplace{\textbf{93.54 ± 1.27}}{97.26 ± 0.43} & \newreplace{92.23 ± 0.48}{\textbf{97.28 ± 0.17}} \\
        \midrule
        \multirow{3}{*}{F1 Macro}
        & SIMB (7 classes) & 1,479
            & \newreplace{62.29 ± 4.60}{67.94 ± 3.33} & \newreplace{\textbf{65.09 ± 2.97}}{\textbf{68.43 ± 2.83}} \\
        & COMB (13 classes) & 1,462
            & \newreplace{50.76 ± 2.23}{48.59 ± 2.20} & \newreplace{\textbf{50.82 ± 3.87}}{\textbf{52.23 ± 1.99}} \\
        & SIBA (7 classes) & 13,790
            & \newreplace{\textbf{93.54 ± 1.26}}{97.26 ± 0.43} & \newreplace{91.91 ± 0.54}{\textbf{97.28 ± 0.17}} \\
        \midrule
        \multirow{3}{*}{F1 Weighted}
        & SIMB (7 classes) & 1,479
            & \newreplace{73.05 ± 1.52}{77.67 ± 2.32} & \newreplace{\textbf{74.49 ± 1.88}}{\textbf{77.80 ± 1.23}} \\
        & COMB (13 classes) & 1,462
            & \newreplace{\textbf{65.51 ± 1.29}}{66.53 ± 1.08} & \newreplace{65.00 ± 1.28}{\textbf{67.59 ± 1.58}} \\
        & SIBA (7 classes) & 13,790
            & \newreplace{\textbf{93.54 ± 1.26}}{97.26 ± 0.43} & \newreplace{92.20 ± 0.49}{\textbf{97.28 ± 0.17}} \\
        \bottomrule
    \end{tabular}
    \end{threeparttable}
\end{table*}

\new{This ablation result is consistent with the gradient-based band-importance analysis in \cref{sec:interpretability_results} (\cref{fig:gradient_band_importance_grouped}): the auxiliary \ac{ERA5} and \ac{SRTM} bands receive comparatively low attribution magnitude relative to the Sentinel-2 short-wave infrared, red, and red-edge bands that dominate the model's predictions, so excluding them has only a modest effect on classification performance -- most visible on the larger SIBA dataset, where the model has enough training data to exploit whatever additional signal these auxiliary bands provide.}

%% CF
\paragraph{Confusion Matrix Comparison} \leavevmode\\
Sets of confusion matrices for SIBA and COMB datasets show per-class accuracy improvements from deep embeddings, especially with fine-tuned Presto and the MLP classifier. 

\leavevmode\\

\begin{figure*}[ht]
    \centering
    \includegraphics[width=17cm]{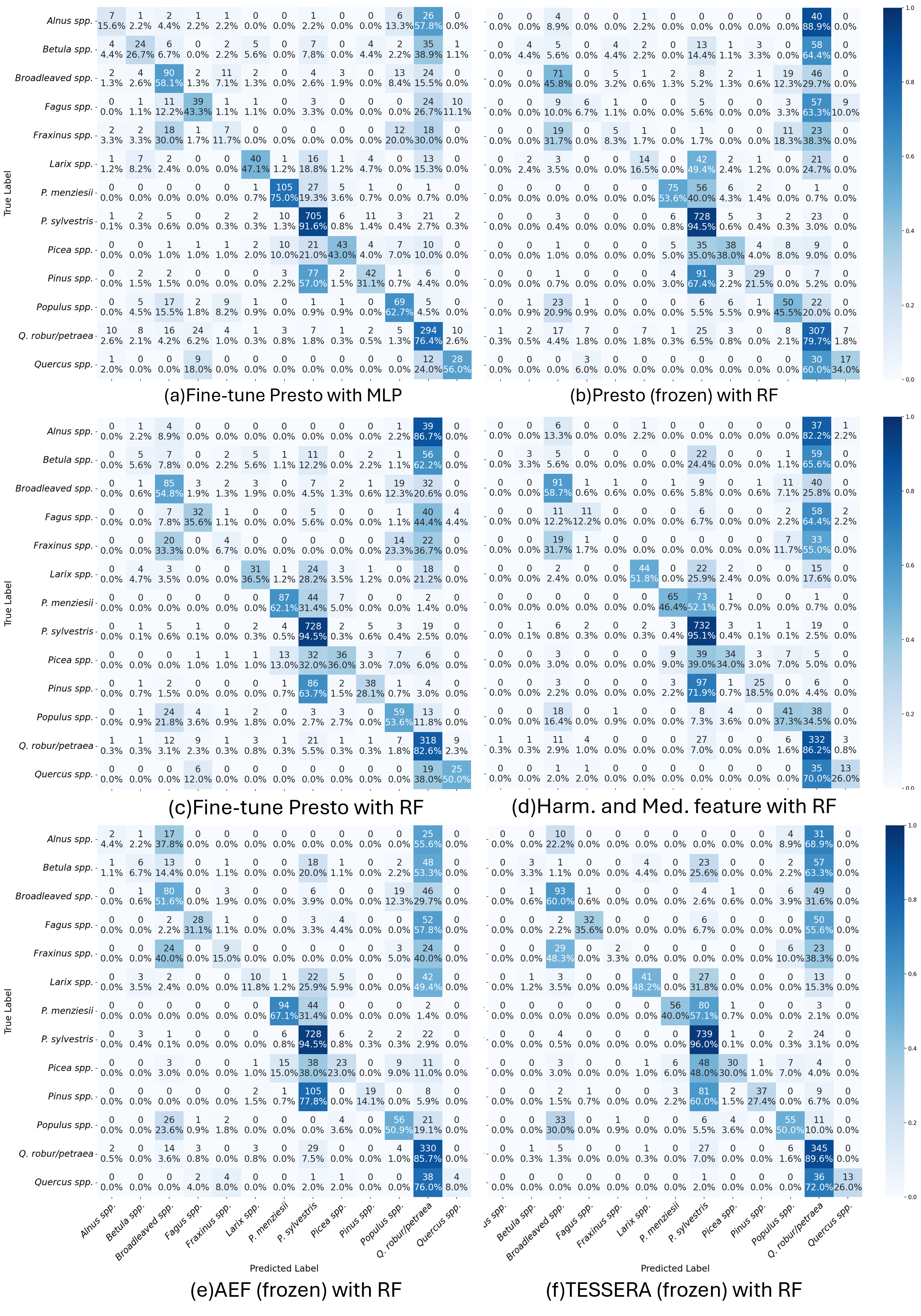}
    \caption{\textit{Confusion matrices for Complex \& Imbalanced (COMB) datasets. Top left: Fine-
tune Presto with MLP Classifier, Top right: Presto (Frozen Presto features), Middle left: Fine-tune
Presto (Presto features), Middle right: Harmonic + medoid feature, Bottom left: AEF, Bottom right: TESSERA. Each cell has average count and recall percentage.}}
    \label{fig:ConfusionMatrix_COMB}
\end{figure*}

Figure \ref{fig:ConfusionMatrix_COMB} presents confusion matrices for the COMB dataset across six model configurations, combining different feature representations (fine-tuned deep embeddings, frozen deep embeddings, harmonic + medoid feature) and classifiers (MLP and Random Forest). This appendix provides a comparative description of class-level behavior across these configurations, complementing the detailed discussion of the fine-tuned MLP results in Section \ref{sec:class_level_performance}.

Across all models, broadleaved species exhibit substantially higher confusion than conifer species. In particular, \textit{Q. robur/petraea} consistently acts as a dominant prediction class, absorbing misclassifications from several less frequent broadleaved species such as \textit{Alnus spp.}, \textit{Fraxinus spp.}, and \textit{Populus spp.}. This pattern is visible irrespective of classifier choice, indicating that class imbalance and overlapping spectral--phenological characteristics dominate model behavior more strongly than the specific learning algorithm.

Comparing classifiers, RF models generally show more fragmented confusion patterns than the fine-tuned MLP. While RF-based models correctly identify dominant classes such as \textit{P. sylvestris} and \textit{Q. robur/petraea}, they exhibit reduced sensitivity to minority classes, with many rare species being almost entirely misclassified into majority classes. This effect is particularly evident for \textit{Alnus spp.} and \textit{Fraxinus spp.}, which achieve very low recall across all RF configurations.

Feature representation also plays a key role. Fine-tuned deep embeddings yield the most diagonally concentrated confusion matrix, especially for conifer species. \textit{P. sylvestris} and \textit{P. menziesii} are consistently well separated under fine-tuned representations, whereas frozen deep features (Presto, AEF, and TESSERA) show increased confusion between conifer taxa, particularly between \textit{Pinus spp.} and \textit{P. sylvestris}. This suggests that task-specific fine-tuning is critical for capturing subtle intra-genus differences.

Models based on harmonic and medoid features display the weakest class separation overall. While they retain some ability to discriminate coarse functional groups (broadleaved vs.\ conifer), fine-grained discrimination among broadleaved species is limited. 

Finally, large-scale pretrained representations (AEF and TESSERA) exhibit mixed behavior. While they perform competitively for dominant classes, they do not consistently improve minority-class recognition in the absence of fine-tuning. In several cases, their confusion patterns resemble those of frozen Presto embeddings, indicating that representation scale alone does not resolve class imbalance or inter-species similarity issues in the COMB dataset.

Overall, the confusion matrices for COMB reinforce three key observations: (i) class imbalance strongly shapes prediction outcomes across all models, (ii) fine-tuning deep embeddings improves intra-group species discrimination, and (iii) classifier choice has a secondary effect compared to data characteristics and feature representation. These findings support the conclusions drawn in the main text regarding the importance of data quality and class balance for operational NFI species mapping.

\leavevmode\\

\begin{figure*}[ht]
    \centering
    \includegraphics[width=17cm]{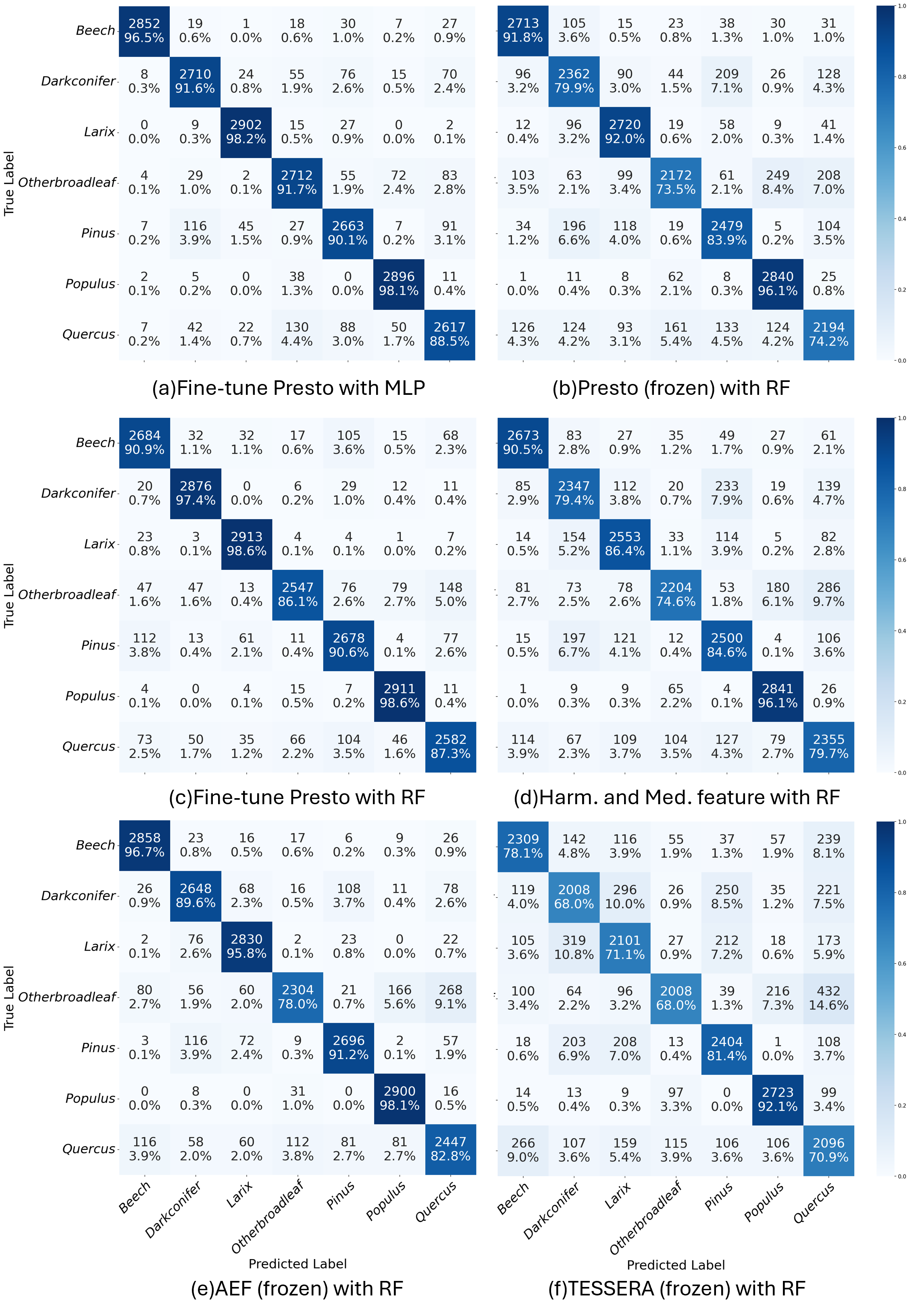}
    \caption{\textit{Confusion matrices for Simplified \& \newreplace{Imbalanced (SIMB)}{Balanced (SIBA)} datasets. Top left: Fine-
tune Presto with MLP Classifier, Top right: Presto (Frozen Presto features), Middle left: Fine-tune
Presto (Presto features), Middle right: Harmonic + medoid feature, Bottom left: AEF, Bottom right: TESSERA. Each cell has average count and recall percentage.}}
    \label{fig:ConfusionMatrix_SIBA}
\end{figure*}

Figure \ref{fig:ConfusionMatrix_SIBA} shows confusion matrices for the SIBA dataset across six model configurations, combining different feature representations and classifiers. In contrast to the COMB dataset, SIBA exhibits markedly stronger class separability across all models, reflecting its larger sample size and substantially improved class balance.

Across configurations, most classes achieve high diagonal dominance, indicating strong recall and precision for both broadleaved and conifer species groups. Fine-tuned Presto with an MLP classifier yields near-optimal performance for all classes, with precision values exceeding 90\,\% for most species and limited off-diagonal confusion. This confirms that, under favorable data conditions, fine-tuned deep embeddings can effectively capture inter-species differences in seasonal and structural signals.

RF classifiers show slightly reduced performance compared to the fine-tuned MLP but retain strong class-level behavior. In particular, dominant classes such as \textit{Beech}, \textit{Larix}, and \textit{Populus} are consistently well identified across all feature representations. Residual confusion is primarily concentrated among broadleaved classes, especially between \textit{Other broadleaf} and \textit{Quercus}, suggesting remaining spectral and phenological overlap even in a balanced dataset.

Frozen deep feature representations (Presto, AEF, and TESSERA) display broadly similar confusion patterns. While they maintain high accuracy for majority classes, they exhibit increased confusion for classes with overlapping ecological traits. For instance, \textit{Dark conifer} shows increased misclassification into \textit{Pinus} and \textit{Larix} under frozen representations, indicating that fine-tuning remains important for resolving subtle intra-group distinctions.

Models based on harmonic and medoid features show the weakest overall class separation, although their performance remains notably stronger than in the COMB dataset. This suggests that the improved balance and density of the SIBA dataset partially compensate for the lower representational capacity of hand-crafted temporal features. Nonetheless, confusion among broadleaved species remains more pronounced than for conifers, consistent with known challenges in deciduous species discrimination.

Overall, the SIBA confusion matrices demonstrate that improved class balance and increased sample size substantially reduce misclassification across all modeling approaches. While fine-tuned deep embeddings provide the most consistent and robust class-level performance, the reduced confusion observed even for simpler models highlights the critical role of data quality in species-level forest classification. These results support the conclusion that targeted data augmentation and balancing strategies can be as influential as model architecture in improving operational NFI mapping performance.

\subsubsection{\new{Full $n\times n$ neighbourhood-patch pooling ablation}}
\label{app:spatial_patch_full}

\new{\cref{tab:spatial_patch_aef_tess_full} gives the full breakdown, across all three pooling strategies (flatten, mean-pool, stats-pool; \cref{sec:spatial_patch_methods}), that underlies the headline mean-pool comparison of \cref{tab:spatial_patch_headline} (\cref{sec:spatial_patch_results}), for both \ac{NFI} (SIMB, COMB) and SIBA (pixel-level and plot-grouped splits). Each dataset has two four-row blocks (one row per pooling strategy): the first pairs AEF with Presto \ac{RF}, the second pairs TESSERA with Presto \ac{MLP}. The center-pixel entries reproduce the corresponding AEF, TESSERA, and fine-tuned Presto values of \cref{tab:RQ1_combined} for \ac{NFI} and \cref{tab:siba_spatial} for SIBA (fine-tuned Presto); the SIBA AEF/TESSERA center-pixel entries are the fixed-year-2020 re-extraction used consistently throughout this spatial-pooling analysis (\cref{sec:spatial_patch_methods}), which differs slightly from \cref{tab:siba_spatial}'s per-plot-year baseline for the same reason already noted for \ac{NFI} (\cref{sec:spatial_patch_results}). Bold values mark, for each dataset, embedding (AEF/TESSERA), or classifier (Presto \ac{RF}/\ac{MLP}) and metric, which of the four pooling strategies (center pixel, flatten, mean-pool, stats-pool) performs best -- i.e.\ up to one bold value per representation per metric per dataset, read down a column within one embedding/classifier's own four rows.}

\new{SIBA's photo-interpreted plots are considerably closer together than \ac{NFI}'s ($\geq$160.8\,m apart) -- some as close as 10--20\,m -- so a $3\times3$ (30\,m) neighbourhood risks incorporating another labelled sample's own pixel. We checked this directly rather than assuming it away: across all 13,790 SIBA plots, none of the eight neighbour positions coincides exactly with another plot's own center-pixel coordinates, so this specific failure mode does not occur despite the closer spacing.
%  For SIBA, the fine-tuned Presto encoder embedding each neighbour pixel is the split-specific encoder already used for the corresponding center-pixel result (pixel-level or plot-grouped, \cref{sec:experiments}), and SIBA's AEF/TESSERA neighbour embeddings use the same fixed-year-2020 re-extraction as the center-pixel comparison, rather than the per-plot-survey-year embeddings used elsewhere in this paper.
 }

% \new{Unlike \ac{NFI}'s well-separated plots ($\geq$160.8\,m apart), some of SIBA's photo-interpreted pixels are only 10--20\,m apart (\cref{app:siba_spatial}), so a $3\times3$ (30\,m) neighbourhood risks incorporating another labelled sample's own pixel; we checked directly for this risk rather than assuming it away, for both the AEF/TESSERA patches and fine-tuned Presto's own re-fine-tuned encoder: across all 13,790 SIBA plots, none of the 8 neighbour positions coincides exactly with another plot's own center-pixel coordinates (0/13,790, verified via nearest-neighbour distance, minimum 9.8\,m; this check applies identically to all three embeddings, since it depends only on the shared neighbour-pixel geometry, not on which embedding is sampled there), so this specific failure mode does not occur even though many plots are spatially close.}

Under SIBA's plot-grouped split, flatten and stats-pool edge out mean-pool for AEF/TESSERA instead, though gains there are small relative to the split's much larger seed-to-seed standard deviation; fine-tuned Presto has no consistent winner on SIBA, since no pooling variant meaningfully beats the center pixel there.

\new{\cref{sec:spatial_patch_results,sec:spatial_patch_discussion} discuss and interpret these results; this table provides the full per-pooling-strategy numbers underlying that discussion.}

\begin{table*}[ht]
    \centering
    \footnotesize
    \setlength{\tabcolsep}{2pt}
    \caption{\new{Full $n\times n$ neighbourhood-patch pooling ablation, AEF/TESSERA (\ac{RF}) and fine-tuned Presto (\ac{RF}/\ac{MLP}) side by side (NFI and SIBA): classification performance (in \%) for the single center pixel and all three pooling strategies. Mean and standard deviation (±) over five seeds. Each dataset has two four-row blocks (one row per pooling strategy): the first pairs AEF with Presto \ac{RF}, the second pairs TESSERA with Presto \ac{MLP}. Bold values indicate, per dataset, per embedding/classifier, and per metric, which of the four pooling strategies (center pixel, flatten, mean-pool, stats-pool) performs best -- i.e.\ up to one bold value per representation per metric within its own four-row block.}}
    \label{tab:spatial_patch_aef_tess_full}
    \begin{newtable}
    \begin{threeparttable}
    \begin{tabular}{
        >{\raggedright}p{1.4cm}
        >{\raggedright}p{2.5cm}|
        >{\centering}p{1.2cm}
        >{\centering}p{1.5cm}
        >{\centering}p{1.5cm}
        >{\centering}p{1.5cm}|
        >{\centering}p{1.2cm}
        >{\centering}p{1.5cm}
        >{\centering}p{1.5cm}
        >{\centering\arraybackslash}p{1.5cm}}
    \toprule
    & & \multicolumn{4}{c}{AEF / TESSERA (RF)} & \multicolumn{4}{c}{Fine-tuned Presto} \\
    \cmidrule(lr){3-6}\cmidrule(lr){7-10}
    Dataset & Pooling & Embed. & OA & F1 Macro & F1 Weighted & Clsf. & OA & F1 Macro & F1 Weighted \\
    \midrule
    \multirow{8}{*}{\makecell[l]{SIMB\\(7 classes)}}
    & Center (single pixel) & \multirow{4}{*}{AEF} & 72.06 ± 1.68 & 52.86 ± 3.08 & 69.83 ± 1.95 & \multirow{4}{*}{RF} & 78.59 ± 1.86 & 68.46 ± 2.13 & 78.17 ± 2.00 \\
    & Flatten & & \textbf{72.20 ± 1.38} & 54.52 ± 2.64 & \textbf{70.32 ± 1.53} & & 81.26 ± 1.17 & 69.81 ± 2.31 & 80.37 ± 1.32 \\
    & Mean-pool & & 72.11 ± 1.33 & \textbf{54.69 ± 2.47} & 70.17 ± 1.56 & & \textbf{81.57 ± 1.29} & \textbf{71.70 ± 1.99} & \textbf{81.01 ± 1.47} \\
    & Stats-pool & & 71.39 ± 1.14 & 52.52 ± 2.69 & 69.14 ± 1.33 & & 80.94 ± 0.68 & 70.10 ± 1.15 & 80.26 ± 0.79 \\
    \cmidrule(lr){2-10}
    & Center (single pixel) & \multirow{4}{*}{TESSERA} & 72.96 ± 1.42 & 57.46 ± 1.72 & 70.40 ± 1.54 & \multirow{4}{*}{MLP} & 78.05 ± 2.20 & 67.94 ± 3.33 & 77.67 ± 2.32 \\
    & Flatten & & 75.61 ± 1.04 & 59.31 ± 2.52 & 72.98 ± 1.18 & & 79.51 ± 0.94 & 70.15 ± 2.93 & 79.26 ± 1.27 \\
    & Mean-pool & & \textbf{78.88 ± 0.96} & \textbf{68.79 ± 1.27} & \textbf{77.76 ± 1.02} & & \textbf{81.03 ± 1.33} & \textbf{72.95 ± 2.22} & \textbf{80.67 ± 1.36} \\
    & Stats-pool & & 77.22 ± 1.21 & 64.72 ± 2.43 & 75.63 ± 1.37 & & 80.18 ± 1.03 & 70.91 ± 2.38 & 79.93 ± 1.07 \\
    \midrule
    \multirow{8}{*}{\makecell[l]{COMB\\(13 classes)}}
    & Center (single pixel) & \multirow{4}{*}{AEF} & \textbf{62.62 ± 0.91} & 36.51 ± 1.71 & 56.81 ± 1.15 & \multirow{4}{*}{RF} & 70.29 ± 0.60 & 53.23 ± 1.34 & 67.96 ± 0.70 \\
    & Flatten & & \textbf{62.62 ± 0.80} & 37.45 ± 1.77 & \textbf{57.10 ± 1.02} & & 71.78 ± 1.17 & 53.58 ± 1.75 & 68.87 ± 1.26 \\
    & Mean-pool & & 62.44 ± 1.08 & \textbf{37.63 ± 2.52} & 56.93 ± 1.24 & & 71.56 ± 1.25 & \textbf{54.47 ± 1.84} & 68.97 ± 1.51 \\
    & Stats-pool & & 62.35 ± 0.71 & 36.00 ± 3.06 & 56.27 ± 1.41 & & \textbf{72.19 ± 1.66} & 53.95 ± 2.47 & \textbf{69.28 ± 1.68} \\
    \cmidrule(lr){2-10}
    & Center (single pixel) & \multirow{4}{*}{TESSERA} & 65.37 ± 1.83 & 43.95 ± 2.89 & 60.59 ± 1.95 & \multirow{4}{*}{MLP} & 69.35 ± 1.08 & 48.59 ± 2.20 & 66.53 ± 1.08 \\
    & Flatten & & 65.96 ± 2.05 & 42.65 ± 3.84 & 60.23 ± 2.48 & & 70.79 ± 1.07 & 53.85 ± 3.59 & 69.02 ± 1.77 \\
    & Mean-pool & & \textbf{71.78 ± 1.87} & \textbf{55.39 ± 2.79} & \textbf{68.64 ± 2.02} & & \textbf{72.10 ± 1.07} & \textbf{57.69 ± 1.61} & \textbf{70.86 ± 1.00} \\
    & Stats-pool & & 69.80 ± 2.17 & 50.59 ± 2.31 & 65.77 ± 2.19 & & 71.15 ± 1.85 & 54.95 ± 5.05 & 69.42 ± 2.75 \\
    \midrule
    \multirow{8}{*}{\makecell[l]{SIBA\\pixel-level}}
    & Center (single pixel) & \multirow{4}{*}{AEF} & 88.86 ± 0.55 & 88.74 ± 0.55 & 88.74 ± 0.55 & \multirow{4}{*}{RF} & 94.50 ± 0.45 & 94.48 ± 0.45 & 94.48 ± 0.45 \\
    & Flatten & & 89.75 ± 0.45 & 89.62 ± 0.46 & 89.62 ± 0.46 & & 94.31 ± 0.39 & 94.29 ± 0.40 & 94.29 ± 0.40 \\
    & Mean-pool & & \textbf{90.67 ± 0.40} & \textbf{90.56 ± 0.41} & \textbf{90.56 ± 0.41} & & 94.48 ± 0.41 & 94.46 ± 0.41 & 94.46 ± 0.41 \\
    & Stats-pool & & 90.00 ± 0.63 & 89.89 ± 0.64 & 89.89 ± 0.64 & & \textbf{94.54 ± 0.52} & \textbf{94.52 ± 0.53} & \textbf{94.52 ± 0.53} \\
    \cmidrule(lr){2-10}
    & Center (single pixel) & \multirow{4}{*}{TESSERA} & 75.33 ± 0.68 & 75.33 ± 0.67 & 75.33 ± 0.67 & \multirow{4}{*}{MLP} & \textbf{97.26 ± 0.43} & \textbf{97.26 ± 0.43} & \textbf{97.26 ± 0.43} \\
    & Flatten & & 78.06 ± 0.77 & 78.08 ± 0.77 & 78.08 ± 0.77 & & 96.84 ± 0.41 & 96.83 ± 0.41 & 96.83 ± 0.41 \\
    & Mean-pool & & \textbf{87.71 ± 0.83} & \textbf{87.67 ± 0.85} & \textbf{87.67 ± 0.85} & & 96.88 ± 0.17 & 96.88 ± 0.17 & 96.88 ± 0.17 \\
    & Stats-pool & & 85.92 ± 0.95 & 85.88 ± 0.97 & 85.88 ± 0.97 & & 96.86 ± 0.17 & 96.85 ± 0.17 & 96.85 ± 0.17 \\
    \midrule
    \multirow{8}{*}{\makecell[l]{SIBA\\plot-grouped}}
    & Center (single pixel) & \multirow{4}{*}{AEF} & 60.78 ± 4.36 & 58.82 ± 4.26 & 58.82 ± 4.40 & \multirow{4}{*}{RF} & \textbf{53.03 ± 3.50} & \textbf{50.85 ± 3.41} & \textbf{50.95 ± 3.44} \\
    & Flatten & & \textbf{61.59 ± 4.39} & \textbf{59.59 ± 4.14} & \textbf{59.58 ± 4.25} & & 52.78 ± 3.67 & 50.65 ± 3.61 & 50.76 ± 3.65 \\
    & Mean-pool & & 60.62 ± 4.52 & 58.42 ± 4.34 & 58.39 ± 4.44 & & 52.93 ± 3.64 & 50.79 ± 3.63 & 50.90 ± 3.68 \\
    & Stats-pool & & 61.12 ± 4.44 & 58.99 ± 4.25 & 58.98 ± 4.37 & & 52.84 ± 3.38 & 50.69 ± 3.36 & 50.79 ± 3.41 \\
    \cmidrule(lr){2-10}
    & Center (single pixel) & \multirow{4}{*}{TESSERA} & 60.76 ± 2.74 & 60.24 ± 2.40 & 60.29 ± 2.57 & \multirow{4}{*}{MLP} & \textbf{48.97 ± 3.14} & \textbf{46.96 ± 3.03} & \textbf{47.04 ± 3.00} \\
    & Flatten & & 62.92 ± 2.67 & \textbf{62.36 ± 2.29} & \textbf{62.37 ± 2.44} & & 47.88 ± 1.78 & 45.22 ± 1.29 & 45.30 ± 1.54 \\
    & Mean-pool & & 62.57 ± 2.76 & 61.44 ± 2.45 & 61.38 ± 2.66 & & 45.89 ± 0.98 & 42.76 ± 1.11 & 42.89 ± 1.56 \\
    & Stats-pool & & \textbf{62.94 ± 2.32} & 61.99 ± 1.96 & 61.95 ± 2.13 & & 45.63 ± 2.80 & 42.95 ± 3.00 & 43.04 ± 2.91 \\
    \bottomrule
    \end{tabular}
    \begin{tablenotes}
    \footnotesize
    \item[] Left block (AEF/TESSERA): the pooled representations are evaluated with the same \ac{RF} protocol used for the center-pixel embeddings (\cref{sec:experiments}); SIBA center-pixel entries are the fixed-year-2020 re-extraction used consistently throughout this analysis (\cref{sec:spatial_patch_methods}), differing slightly from \cref{tab:siba_spatial}'s per-plot-year baseline. Right block (fine-tuned Presto): the pooled representations are evaluated with a freshly trained classifier on the pooled fine-tuned-Presto embeddings, analogous to the fine-tuned Presto \ac{RF} pipeline of \cref{tab:RQ1_combined} rather than the end-to-end \ac{MLP} head (\cref{sec:spatial_patch_methods}).
    \end{tablenotes}
    \end{threeparttable}
    \end{newtable}
\end{table*}

\subsection{\new{1M-pixel SIMB maps}}
\label{app:national_map_1M_validation}

\new{Before the full wall-to-wall maps (\cref{sec:national_map_methods}), all methods were screened on a 1,000,000-point random forest-pixel sample; their overall accuracy and grid-cell agreement with the \ac{NFI} are reported in the main text (\cref{tab:national_map_1M,tab:national_map_1M_gridcell}). The SIMB-trained maps are shown in \cref{fig:national_map_1M_simb} below; the SIBA-trained maps, where the degeneration appears, are in the main text (\cref{fig:national_map_1M_siba}).}

\begin{figure*}[!ht]
    \centering
    \includegraphics[width=\linewidth]{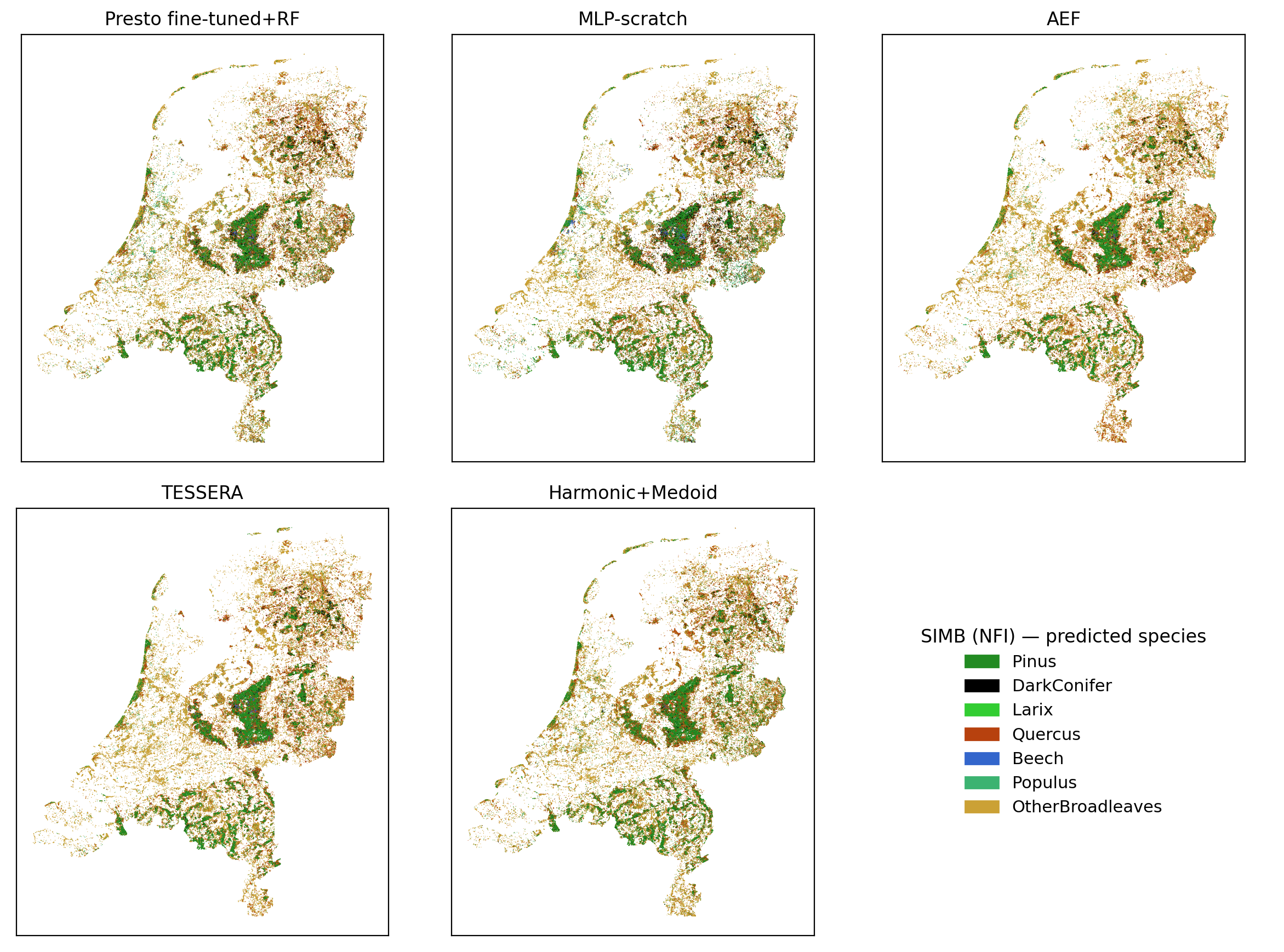}
    \caption{\new{Predicted species at 1,000,000 random forest pixels, SIMB-trained: fine-tuned Presto, from-scratch \ac{MLP}, AEF, TESSERA, and hand-crafted harmonic + medoid. All five produce plausible maps. Corresponding SIBA-trained maps in \cref{fig:national_map_1M_siba} (main text).}}
    \label{fig:national_map_1M_simb}
\end{figure*}

\subsection{\new{National-scale mapping: additional maps, tables, and regional comparison}}
\label{app:national_map_annex}

\begin{figure*}[!ht]
    \begin{subfigure}{.48\textwidth}
        \centering
        \includegraphics[width=\textwidth]{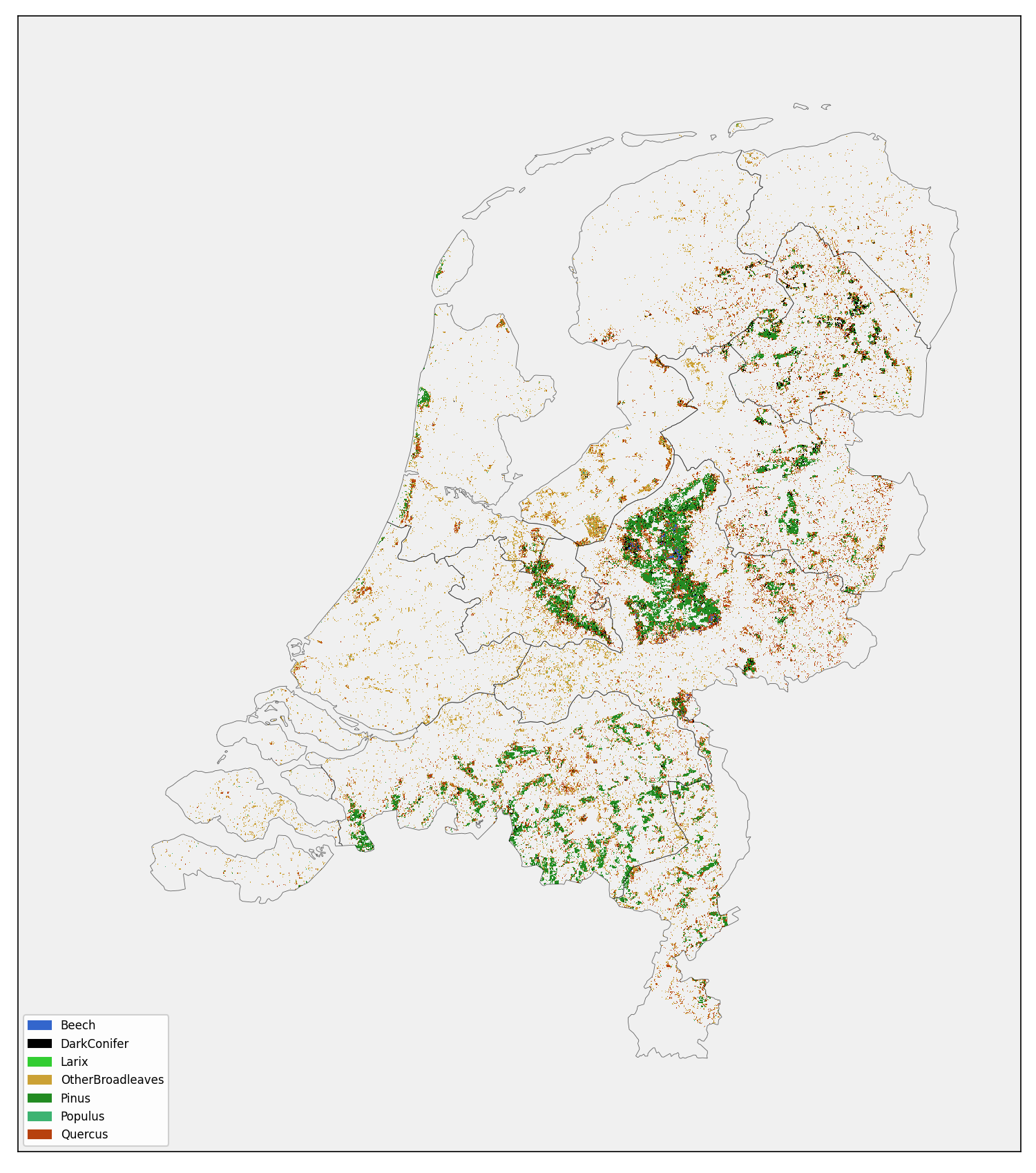}
        \caption{\new{AEF, trained on SIMB}}
        \label{fig:national_map_aef:simb}
    \end{subfigure}
    \hfill
    \begin{subfigure}{.48\textwidth}
        \centering
        \includegraphics[width=\textwidth]{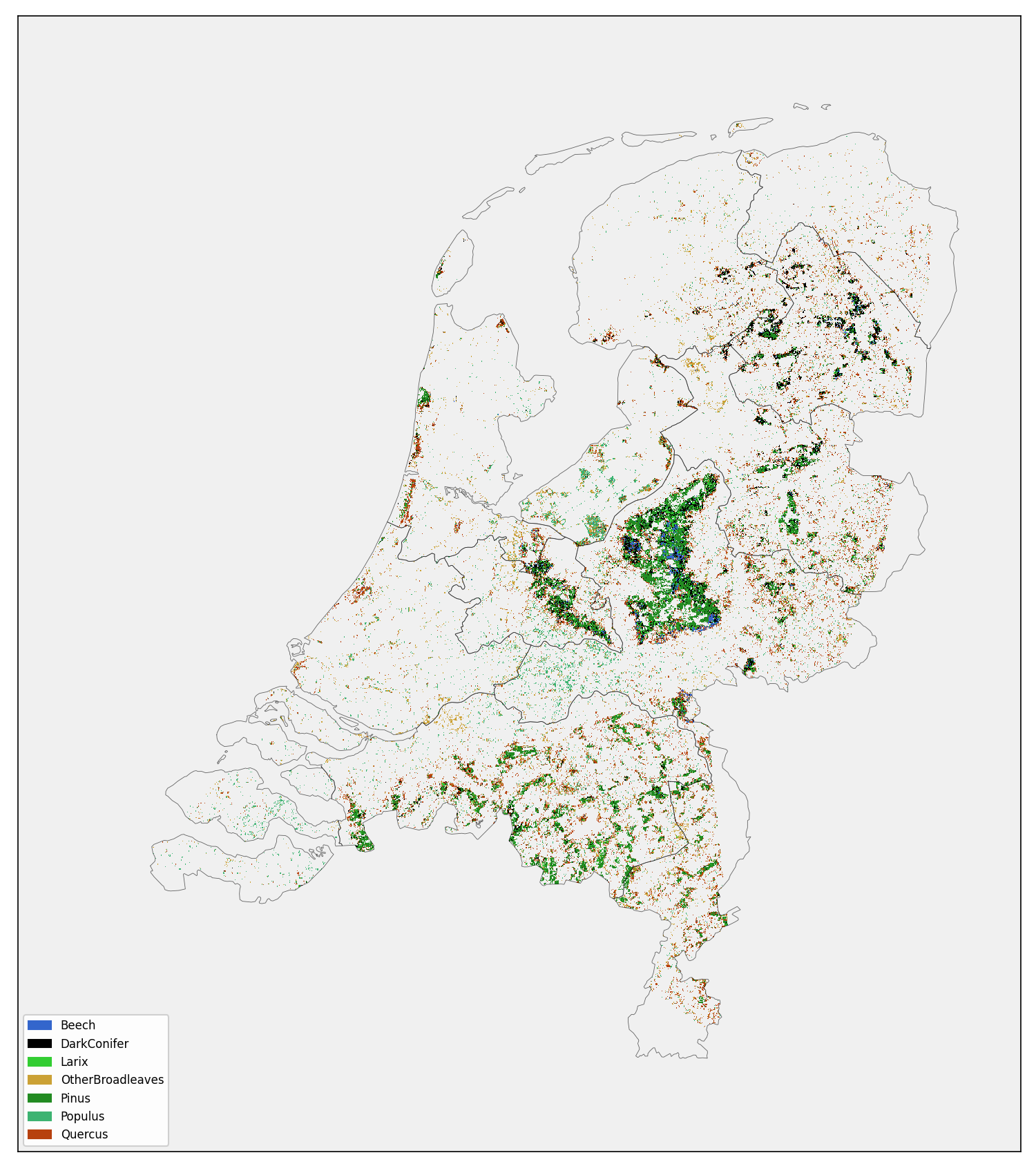}
        \caption{\new{AEF, trained on SIBA}}
        \label{fig:national_map_aef:siba}
    \end{subfigure}
    \caption{\new{National-scale species classification (10\,m resolution, forest pixels only), AEF embeddings trained on SIMB (left, 67.03\% $\pm$ 4.74\% overall accuracy) and SIBA (right, 86.15\% $\pm$ 1.24\%). Corresponding TESSERA maps in \cref{fig:national_map_tessera}.}}
    \label{fig:national_map_aef}
\end{figure*}

\new{This section provides the material supporting \cref{sec:national_map_methods,sec:national_map_results,sec:national_map_discussion}: the AEF national maps (\cref{fig:national_map_aef}), full per-class map accuracy and area estimates (\cref{tab:national_map_aef_simb,tab:national_map_aef_siba,tab:national_map_tessera_simb,tab:national_map_tessera_siba}), the reference confusion matrices underlying those estimates (\cref{tab:national_map_cm_aef_simb,tab:national_map_cm_aef_siba,tab:national_map_cm_tessera_simb}; the TESSERA $\times$ SIBA confusion matrix is given in \cref{sec:national_map_results}, \cref{tab:national_map_cm_tessera_siba}), and the qualitative regional comparison against \citet{francini_DutchForestSpeciesMap_2024}'s own Fig.~7 (\cref{fig:national_map_regions}). The full grid-cell comparison against the \ac{NFI} is given in \cref{sec:national_map_results}, \cref{tab:national_map_gridcell}.}

\subsubsection{\new{Per-class map accuracy and area estimates}}

\new{\cref{tab:national_map_aef_simb,tab:national_map_aef_siba,tab:national_map_tessera_simb,tab:national_map_tessera_siba} report, per class: $W_i$ (proportion of total mapped forest area), user's and producer's accuracy with 95\% confidence intervals, area-adjusted extent with 95\% confidence interval, and $n_i$, the number of reference points landing in that class's predicted stratum (Olofsson et al., 2014; classes with $n_i<20$ have an unstable confidence interval, flagged in the main text discussion).}

\begin{table*}[htbp]
    \centering
    \small
    \caption{\new{AEF, trained on SIMB: per-class map accuracy and area (37,880,448 forest px total; overall accuracy 67.03\% $\pm$ 4.74\%).}}
    \label{tab:national_map_aef_simb}
    \begin{newtable}
    \begin{threeparttable}
    \begin{tabular}{lrrrrr}
    \toprule
    Class & $W_i$ (\%) & User's acc. (\%) & Producer's acc. (\%) & Area (ha) & $n_i$ \\
    \midrule
    Beech & 0.65 & $60.0 \pm 48.0$ & $8.1 \pm 7.0$ & $18{,}316 \pm 8{,}502$ & 5 \\
    DarkConifer & 4.22 & $80.6 \pm 14.1$ & $45.4 \pm 11.3$ & $28{,}414 \pm 6{,}921$ & 31 \\
    Larix & 1.05 & $100.0 \pm 0.0$ & $23.7 \pm 9.7$ & $16{,}796 \pm 6{,}886$ & 2 \\
    OtherBroadleaves & 35.16 & $56.5 \pm 10.2$ & $81.8 \pm 6.8$ & $92{,}046 \pm 15{,}294$ & 92 \\
    Pinus & 31.05 & $82.0 \pm 5.3$ & $85.4 \pm 6.1$ & $112{,}924 \pm 10{,}182$ & 200 \\
    Populus & 0.51 & $75.0 \pm 32.1$ & $6.3 \pm 3.7$ & $23{,}161 \pm 10{,}084$ & 8 \\
    Quercus & 27.35 & $60.2 \pm 9.3$ & $71.6 \pm 9.0$ & $87{,}146 \pm 14{,}077$ & 108 \\
    \bottomrule
    \end{tabular}
    \end{threeparttable}
    \end{newtable}
\end{table*}

\begin{table*}[htbp]
    \centering
    \small
    \caption{\new{AEF, trained on SIBA: per-class map accuracy and area (37,880,448 forest px total; overall accuracy 86.15\% $\pm$ 1.24\%).}}
    \label{tab:national_map_aef_siba}
    \begin{newtable}
    \begin{threeparttable}
    \begin{tabular}{lrrrrr}
    \toprule
    Class & $W_i$ (\%) & User's acc. (\%) & Producer's acc. (\%) & Area (ha) & $n_i$ \\
    \midrule
    Beech & 2.99 & $87.9 \pm 2.6$ & $66.2 \pm 7.2$ & $15{,}037 \pm 1{,}653$ & 622 \\
    DarkConifer & 13.07 & $86.1 \pm 2.8$ & $81.1 \pm 3.5$ & $52{,}556 \pm 2{,}638$ & 583 \\
    Larix & 7.75 & $88.2 \pm 2.5$ & $85.6 \pm 4.1$ & $30{,}239 \pm 1{,}615$ & 626 \\
    OtherBroadleaves & 15.40 & $91.3 \pm 2.5$ & $75.6 \pm 3.3$ & $70{,}493 \pm 3{,}353$ & 496 \\
    Pinus & 18.45 & $89.1 \pm 2.5$ & $90.5 \pm 2.3$ & $68{,}830 \pm 2{,}466$ & 587 \\
    Populus & 13.90 & $93.1 \pm 2.0$ & $97.7 \pm 1.5$ & $50{,}188 \pm 1{,}300$ & 624 \\
    Quercus & 28.44 & $77.3 \pm 3.4$ & $91.1 \pm 1.7$ & $91{,}463 \pm 3{,}985$ & 599 \\
    \bottomrule
    \end{tabular}
    \end{threeparttable}
    \end{newtable}
\end{table*}

\begin{table*}[htbp]
    \centering
    \small
    \caption{\new{TESSERA, trained on SIMB: per-class map accuracy and area (37,880,447 forest px total; overall accuracy 74.69\% $\pm$ 4.47\%).}}
    \label{tab:national_map_tessera_simb}
    \begin{newtable}
    \begin{threeparttable}
    \begin{tabular}{lrrrrr}
    \toprule
    Class & $W_i$ (\%) & User's acc. (\%) & Producer's acc. (\%) & Area (ha) & $n_i$ \\
    \midrule
    Beech & 1.69 & $87.5 \pm 24.5$ & $33.7 \pm 14.8$ & $16{,}597 \pm 6{,}775$ & 8 \\
    DarkConifer & 3.69 & $88.5 \pm 12.5$ & $40.6 \pm 10.4$ & $30{,}390 \pm 7{,}532$ & 26 \\
    Larix & 1.27 & $100.0 \pm 0.0$ & $40.8 \pm 15.3$ & $11{,}783 \pm 4{,}435$ & 7 \\
    OtherBroadleaves & 33.42 & $65.5 \pm 10.0$ & $86.7 \pm 5.9$ & $95{,}720 \pm 14{,}180$ & 87 \\
    Pinus & 31.98 & $82.5 \pm 5.2$ & $88.7 \pm 5.9$ & $112{,}719 \pm 9{,}788$ & 206 \\
    Populus & 0.83 & $100.0 \pm 0.0$ & $13.3 \pm 5.6$ & $23{,}502 \pm 9{,}833$ & 8 \\
    Quercus & 27.12 & $72.1 \pm 8.7$ & $84.1 \pm 7.9$ & $88{,}094 \pm 12{,}009$ & 104 \\
    \bottomrule
    \end{tabular}
    \end{threeparttable}
    \end{newtable}
\end{table*}

\begin{table*}[htbp]
    \centering
    \small
    \caption{\new{TESSERA, trained on SIBA: per-class map accuracy and area (37,880,447 forest px total; overall accuracy 84.68\% $\pm$ 1.36\%).}}
    \label{tab:national_map_tessera_siba}
    \begin{newtable}
    \begin{threeparttable}
    \begin{tabular}{lrrrrr}
    \toprule
    Class & $W_i$ (\%) & User's acc. (\%) & Producer's acc. (\%) & Area (ha) & $n_i$ \\
    \midrule
    Beech & 3.10 & $89.0 \pm 2.5$ & $63.4 \pm 7.3$ & $16{,}473 \pm 1{,}904$ & 611 \\
    DarkConifer & 13.53 & $84.9 \pm 2.9$ & $79.9 \pm 3.6$ & $54{,}452 \pm 2{,}868$ & 576 \\
    Larix & 7.90 & $88.5 \pm 2.5$ & $83.4 \pm 4.1$ & $31{,}775 \pm 1{,}726$ & 610 \\
    OtherBroadleaves & 16.61 & $88.5 \pm 2.7$ & $77.8 \pm 3.4$ & $71{,}526 \pm 3{,}559$ & 522 \\
    Pinus & 18.68 & $85.4 \pm 2.8$ & $88.1 \pm 2.7$ & $68{,}609 \pm 2{,}889$ & 603 \\
    Populus & 7.53 & $93.1 \pm 2.0$ & $92.6 \pm 3.7$ & $28{,}659 \pm 1{,}275$ & 619 \\
    Quercus & 32.66 & $79.0 \pm 3.3$ & $91.0 \pm 1.6$ & $107{,}311 \pm 4{,}480$ & 594 \\
    \bottomrule
    \end{tabular}
    \end{threeparttable}
    \end{newtable}
\end{table*}

\subsubsection{\new{1M vs full pixels map grid-cell deviation comparison}}
\label{app:national_map_1M_vs_census}

\new{For AEF and TESSERA, which have both a 1M-sample and a full 37.9M-pixel census map, the 1M-sample mapping shows a trend of classification at a low cost. User's accuracy and its confidence interval between 1M-sample and full census maps are identical between the two (they do not depend on mapped-area weights), overall accuracy and national \ac{NFI}-area deviation agree to within about one percentage point, and the $10\times10$\,km grid-cell deviation agrees to within 0.5--0.7\,pp (\cref{tab:national_map_1M_vs_census}), a small upward bias in the sample from its sparser per-cell sampling. This implies using a cheap 1M-point map to judge whether a full map will be sound before committing to full-country extraction.}

\begin{table}[htbp]
    \centering
    \small
    \caption{\new{Grid-cell species-proportion deviation (\%) from the \ac{NFI}, full 37.9M-pixel census versus the 1M-pixel sample, for the two embeddings that have both. All / $\geq$10 / $\geq$20 = \ac{NFI} plot-count thresholds per cell.}}
    \label{tab:national_map_1M_vs_census}
    \begin{newtable}
    \begin{threeparttable}
    \begin{tabular}{llrrrrrr}
    \toprule
     & & \multicolumn{3}{c}{Full census} & \multicolumn{3}{c}{1M sample} \\
    \cmidrule(lr){3-5}\cmidrule(lr){6-8}
    Embedding & Source & All & $\geq$10 & $\geq$20 & All & $\geq$10 & $\geq$20 \\
    \midrule
    AEF     & SIMB &  8.70 & 6.10 & 6.03 &  9.38 & 6.74 & 6.48 \\
    AEF     & SIBA & 11.00 & 8.12 & 6.78 & 11.01 & 7.92 & 6.61 \\
    TESSERA & SIMB &  8.50 & 5.82 & 6.07 &  9.07 & 6.31 & 6.34 \\
    TESSERA & SIBA & 10.43 & 7.95 & 6.05 &  9.92 & 7.38 & 5.77 \\
    \bottomrule
    \end{tabular}
    \end{threeparttable}
    \end{newtable}
\end{table}

\subsubsection{\new{Reference confusion matrices}}

\new{\cref{tab:national_map_cm_aef_simb,tab:national_map_cm_aef_siba,tab:national_map_cm_tessera_simb} give the reference confusion matrix underlying each map's accuracy estimate (cell values: raw counts of held-out reference points), in the same format as \citeauthor{francini_DutchForestSpeciesMap_2024}'s own Table 2. The corresponding TESSERA $\times$ SIBA confusion matrix is given in the main text (\cref{sec:national_map_results}, \cref{tab:national_map_cm_tessera_siba}).}

\begin{table*}[htbp]
    \centering
    \footnotesize
    \caption{\new{AEF $\times$ SIMB reference confusion matrix ($n=446$).}}
    \label{tab:national_map_cm_aef_simb}
    \begin{newtable}
    \begin{threeparttable}
    \begin{tabular}{lrrrrrrrrr}
    \toprule
     & \multicolumn{7}{c}{Reference (true) class} & \\
    \cmidrule(lr){2-8}
    Map (predicted) class & Be & DC & La & OB & Pi & Po & Qu & User's acc. (\%) \\
    \midrule
    Beech & 3 & 0 & 0 & 0 & 0 & 0 & 2 & $60.0 \pm 48.0$ \\
    DarkConifer & 0 & 25 & 1 & 1 & 4 & 0 & 0 & $80.6 \pm 14.1$ \\
    Larix & 0 & 0 & 2 & 0 & 0 & 0 & 0 & $100.0 \pm 0.0$ \\
    OtherBroadleaves & 5 & 1 & 3 & 52 & 4 & 13 & 14 & $56.5 \pm 10.2$ \\
    Pinus & 0 & 17 & 7 & 6 & 164 & 0 & 6 & $82.0 \pm 5.3$ \\
    Populus & 0 & 1 & 0 & 1 & 0 & 6 & 0 & $75.0 \pm 32.1$ \\
    Quercus & 10 & 4 & 4 & 13 & 9 & 3 & 65 & $60.2 \pm 9.3$ \\
    \bottomrule
    \multicolumn{9}{r}{Overall accuracy: $67.03\% \pm 4.74\%$}
    \end{tabular}
    \begin{tablenotes}
    \footnotesize
    \item[] Rows: map (predicted) class; columns: reference (true) class. Be = Beech, DC = DarkConifer, La = Larix, OB = OtherBroadleaves, Pi = Pinus, Po = Populus, Qu = Quercus.
    \end{tablenotes}
    \end{threeparttable}
    \end{newtable}
\end{table*}

\begin{table*}[htbp]
    \centering
    \footnotesize
    \caption{\new{AEF $\times$ SIBA reference confusion matrix ($n=4137$).}}
    \label{tab:national_map_cm_aef_siba}
    \begin{newtable}
    \begin{threeparttable}
    \begin{tabular}{lrrrrrrrrr}
    \toprule
     & \multicolumn{7}{c}{Reference (true) class} & \\
    \cmidrule(lr){2-8}
    Map (predicted) class & Be & DC & La & OB & Pi & Po & Qu & User's acc. (\%) \\
    \midrule
    Beech & 547 & 7 & 1 & 16 & 0 & 1 & 50 & $87.9 \pm 2.6$ \\
    DarkConifer & 11 & 502 & 20 & 6 & 31 & 1 & 12 & $86.1 \pm 2.8$ \\
    Larix & 9 & 19 & 552 & 11 & 20 & 0 & 15 & $88.2 \pm 2.5$ \\
    OtherBroadleaves & 6 & 5 & 0 & 453 & 1 & 6 & 25 & $91.3 \pm 2.5$ \\
    Pinus & 2 & 33 & 10 & 7 & 523 & 0 & 12 & $89.1 \pm 2.5$ \\
    Populus & 1 & 1 & 0 & 27 & 0 & 581 & 14 & $93.1 \pm 2.0$ \\
    Quercus & 15 & 24 & 8 & 71 & 16 & 2 & 463 & $77.3 \pm 3.4$ \\
    \bottomrule
    \multicolumn{9}{r}{Overall accuracy: $86.15\% \pm 1.24\%$}
    \end{tabular}
    \begin{tablenotes}
    \footnotesize
    \item[] Rows: map (predicted) class; columns: reference (true) class. Be = Beech, DC = DarkConifer, La = Larix, OB = OtherBroadleaves, Pi = Pinus, Po = Populus, Qu = Quercus.
    \end{tablenotes}
    \end{threeparttable}
    \end{newtable}
\end{table*}

\begin{table*}[htbp]
    \centering
    \footnotesize
    \caption{\new{TESSERA $\times$ SIMB reference confusion matrix ($n=446$).}}
    \label{tab:national_map_cm_tessera_simb}
    \begin{newtable}
    \begin{threeparttable}
    \begin{tabular}{lrrrrrrrrr}
    \toprule
     & \multicolumn{7}{c}{Reference (true) class} & \\
    \cmidrule(lr){2-8}
    Map (predicted) class & Be & DC & La & OB & Pi & Po & Qu & User's acc. (\%) \\
    \midrule
    Beech & 7 & 0 & 0 & 1 & 0 & 0 & 0 & $87.5 \pm 24.5$ \\
    DarkConifer & 0 & 23 & 2 & 1 & 0 & 0 & 0 & $88.5 \pm 12.5$ \\
    Larix & 0 & 0 & 7 & 0 & 0 & 0 & 0 & $100.0 \pm 0.0$ \\
    OtherBroadleaves & 2 & 2 & 0 & 57 & 4 & 14 & 8 & $65.5 \pm 10.0$ \\
    Pinus & 2 & 19 & 5 & 6 & 170 & 0 & 4 & $82.5 \pm 5.2$ \\
    Populus & 0 & 0 & 0 & 0 & 0 & 8 & 0 & $100.0 \pm 0.0$ \\
    Quercus & 7 & 4 & 3 & 8 & 7 & 0 & 75 & $72.1 \pm 8.7$ \\
    \bottomrule
    \multicolumn{9}{r}{Overall accuracy: $74.69\% \pm 4.47\%$}
    \end{tabular}
    \begin{tablenotes}
    \footnotesize
    \item[] Rows: map (predicted) class; columns: reference (true) class. Be = Beech, DC = DarkConifer, La = Larix, OB = OtherBroadleaves, Pi = Pinus, Po = Populus, Qu = Quercus.
    \end{tablenotes}
    \end{threeparttable}
    \end{newtable}
\end{table*}

\subsubsection{\new{Qualitative regional comparison}}

\new{Since \citeauthor{francini_DutchForestSpeciesMap_2024}'s own Fig.~7 shows three zoom-ins, \cref{fig:national_map_regions} shows the corresponding regions from all four national maps (AEF and TESSERA, SIMB- and SIBA-trained) as our own approximation of the same areas, hand-tuned against the classified rasters rather than lifted from the source figure. Veluwe is described by \citeauthor{francini_DutchForestSpeciesMap_2024} as remnants of oak/beech complexes, dark conifers, and Scots pine afforestation; Noord-Brabant as a smaller-scale pattern of Scots pine, Larix, and oak coppice. The third region, Flevopolder, is discussed in the main text (\cref{sec:national_map_results}, \cref{fig:national_map_regions_flevo}).}

\begin{figure*}[!ht]
    \centering
    \includegraphics[width=0.9\linewidth]{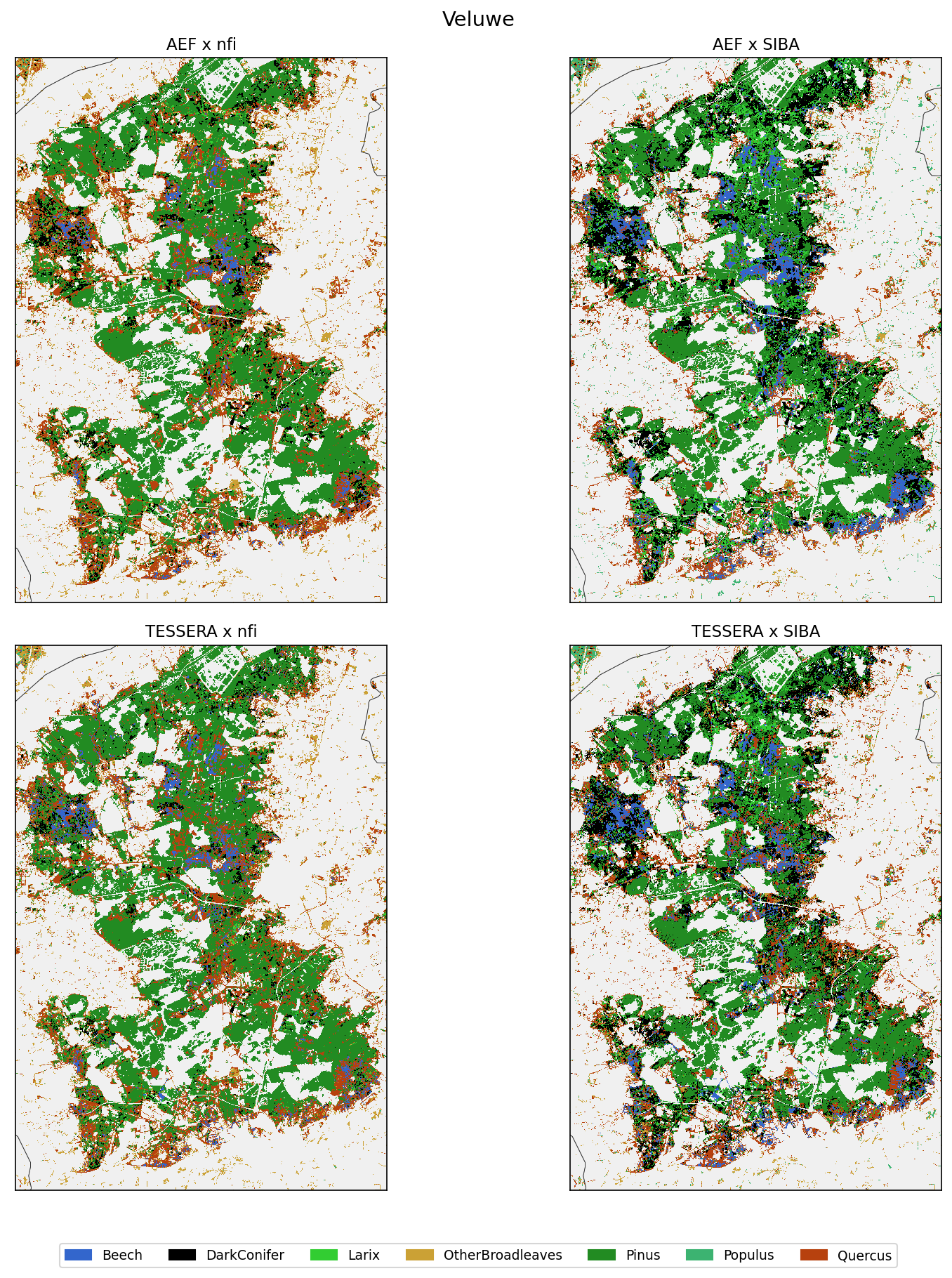}
    \caption{\new{Regional comparison, Veluwe.}}
    \label{fig:national_map_regions_veluwe}
\end{figure*}

\begin{figure*}[!ht]
    \centering
    \includegraphics[width=0.9\linewidth]{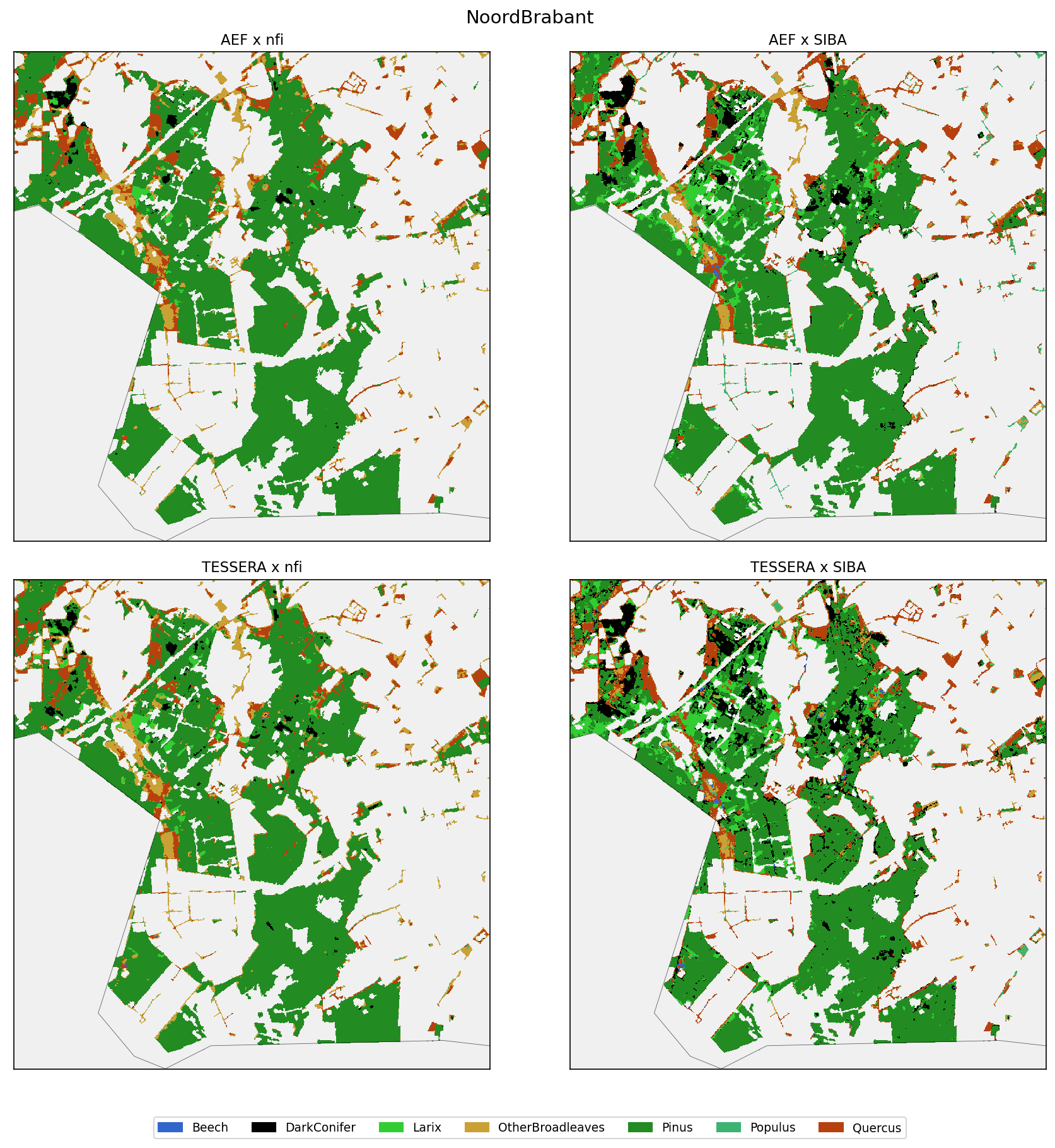}
    \caption{\new{Regional comparison, Noord-Brabant.}}
    \label{fig:national_map_regions}
\end{figure*}

\end{document}